\documentclass{article}
\usepackage[margin=1.0in]{geometry}

\usepackage[dvipsnames,svgnames, table,xcdraw]{xcolor}
\usepackage{relsize}
\usepackage{tcolorbox}
\usepackage{float}
\usepackage{listings}
\usepackage{subcaption}
\usepackage{algorithm}
\usepackage{algorithmic}
\usepackage{comment}
\usepackage{graphicx}
\usepackage{booktabs}
\usepackage{multirow}
\usepackage{todonotes}
\usepackage{enumitem}
\usepackage{url}
\usepackage{mathpazo}
\usepackage{amsmath,amssymb,amsthm,amsfonts,bm}
\usepackage[toc,page,header]{appendix}
\usepackage{fancyhdr}
\usepackage[backend=biber,style=nature,natbib=true,maxbibnames=99,minalphanames=3]{biblatex}
\usepackage[colorlinks=true]{hyperref}
\usepackage[T1]{fontenc}
\usepackage{makecell}
\usepackage[misc]{ifsym}

\usepackage{microtype}
\usepackage{xspace}
\usepackage{tgheros}
\usepackage{minitoc}
\usepackage{titletoc}
\usepackage{etoc}

\usepackage{auth_detailed}

\counterwithout{figure}{section}
\counterwithout{table}{section}
\definecolor{darkgoldenrod}{rgb}{0.72, 0.53, 0.04}
\definecolor{backgroundcolor}{RGB}{250, 250, 252}   
\definecolor{keywordcolor}{RGB}{30, 0, 178}       
\definecolor{stringcolor}{RGB}{204, 0, 102}        
\definecolor{numbercolor}{RGB}{0, 128, 128}        
\definecolor{emphcolor}{RGB}{30, 0, 178}            
\definecolor{commentcolor}{RGB}{0, 128, 0}       
\definecolor{basiccodecolor}{RGB}{61, 61, 61}       

\lstdefinestyle{customstyle}{
    backgroundcolor=\color{backgroundcolor},   
    commentstyle=\color{commentcolor},
    keywordstyle=\color{keywordcolor},
    numberstyle=\color{numbercolor},
    stringstyle=\color{stringcolor},
    basicstyle=\color{basiccodecolor}\ttfamily\footnotesize,
    breakatwhitespace=false,         
    breaklines=true,                 
    captionpos=b,                    
    keepspaces=true,                 
    numbers=left,     
    basicstyle=\color{basiccodecolor}\ttfamily\footnotesize,
    numbersep=5pt,             
    xleftmargin=2em,
    showspaces=false,                
    showstringspaces=false,
    showtabs=false,                  
    tabsize=1,
    frame=single,
    framesep=5pt,
    framexleftmargin=1.5em,
    framexrightmargin=1.5em,
    framextopmargin=1pt,
    framexbottommargin=1pt,
    aboveskip=10pt,
    belowskip=10pt,
    breaklines=true,
    breakautoindent=true,
    emph={textgrad, tg, Variable, MultipleChoiceTestTime,
    TextualGradientDescent, BlackboxLLM},             
    emphstyle={\color{emphcolor}},
    extendedchars=true,
}

\usepackage[normalem]{ulem}
\usepackage{xcolor}

\newcommand{\scrcmd}{\textsuperscript}
\newcommand{\scr}[1]{\scrcmd{#1}}

\definecolor{logocolor}{RGB}{30, 0, 178}                

\newtcolorbox{ttcolorbox}[1][]{colframe=logocolor, colback=logocolor!5!white, title=#1}

\newtcolorbox{apxtcolorbox}[1][]{colframe=black, colback=black!3!white, title=#1}

\definecolor{cerebrablue}{RGB}{66,113,155}

\newcommand{\CEREBRA}{\textcolor{cerebrablue}{\textbf{CEREBRA}}\xspace}

\newcounter{codesnippet}

\hypersetup{
  colorlinks=true,
  urlcolor=BrickRed,
  citecolor=ForestGreen,
  linkcolor=BrickRed
} 
\newcommand{\arialtitle}[1]{{\fontfamily{phv}\selectfont #1}}

\title{\raisebox{-0.\height}{\includegraphics[height=1.8em]{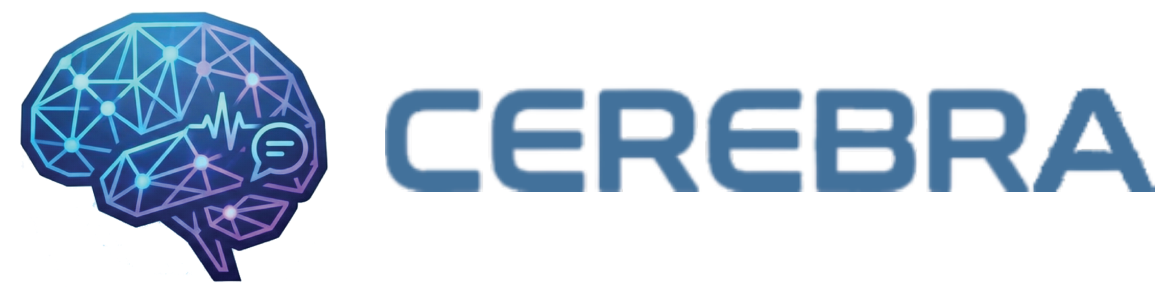}}{\fontsize{13pt}{12pt}\selectfont \textbf{\arialtitle{\\A Multidisciplinary AI Board for Multimodal Dementia Characterization and Risk Assessment}}}}

\begin{document}

\fancyhead[L]{\raisebox{-0.3\height}{\includegraphics[height=1.5em]
{figures/cerebra_logo_v1_cropped.pdf}}}

\fancyhead[R]{\small A Multidisciplinary AI Board for Multimodal Dementia Characterization and Risk Assessment}
\pagestyle{fancy}

\newcounter{suppfigure}
\newcounter{supptable}
\makeatletter
\newcommand\suppfigurename{Supplementary Figure}
\newcommand\supptablename{Supplementary Table}
\newcommand\suppfigureautorefname{\suppfigurename}
\newcommand\supptableautorefname{\supptablename}
\let\oldappendix\appendix
\renewcommand\appendix{%
    \oldappendix
    \setcounter{figure}{0}%
    \setcounter{table}{0}%
    \renewcommand\figurename{\suppfigurename}%
    \renewcommand\tablename{\supptablename}%
}
\makeatother
\maketitle
\vspace{-2.75em}

{Sheng Liu$^{1,2,*,~\textrm{\Letter}}$, Long Chen$^{3,*}$, Zeyun Zhao$^{4,\dagger}$, Qinglin Gou$^{5,\dagger}$, Qingyue Wei$^{2}$, Arjun Masurkar$^{6,7}$, Kevin M Spiegler$^{6}$, Philip Kuball$^{6}$, Stefania C Bray$^{8}$, Megan Bernath$^{9}$, Deanna R Willis$^{9}$, Jiang Bian$^{10,13}$, Lei Xing$^{2}$, Eric Topol$^{11}$, Kyunghyun Cho$^{3,12}$, Yu Huang$^{10,13}$, Ruogu Fang$^{4,14}$, Narges Razavian$^{3,15,16,~\textrm{\Letter}}$, James Zou$^{1,~\textrm{\Letter}}$
\par}

\vspace{0.75em}

{\normalsize\small
$^{1}$Department of Biomedical Data Science, Stanford University \par
$^{2}$Department of Radiation Oncology, Stanford University \par
$^{3}$Center for Data Science, New York University \par
$^{4}$J. Crayton Pruitt Family Department of Biomedical Engineering, University of Florida \par
$^{5}$Department of Biomedical Engineering and Informatics, Indiana University Indianapolis \par
$^{6}$Department of Neurology, NYU Grossman School of Medicine \par
$^{7}$Department of Neuroscience, NYU Grossman School of Medicine \par
$^{8}$UF Health Family Medicine -- Haile Plantation \par
$^{9}$Department of Family Medicine, Indiana University School of Medicine \par
$^{10}$Department of Biostatistics and Health Data Science, Indiana University School of Medicine \par
$^{11}$Scripps Research Translational Institute \par
$^{12}$Courant Institute, New York University \par
$^{13}$Center for Biomedical Informatics, Regenstrief Institute \par
$^{14}$Center for Cognitive Aging and Memory, McKnight Brain Institute, University of Florida \par
$^{15}$Population Health Department, NYU Langone Health \par
$^{16}$Radiology Department, NYU Langone Health \par
\par}

\begin{center} \href{https://github.com/shengliu66/Cerebra}{\raisebox{-0.1\height}{\includegraphics[height=1em]
{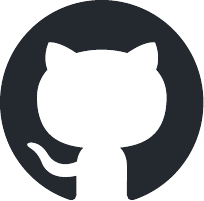}}
\textcolor{cerebrablue}{Code Repository}} \quad\quad\quad\quad\quad \href{https://Cerebra-Health.com}{\raisebox{-0.1\height}{\includegraphics[height=1em]
{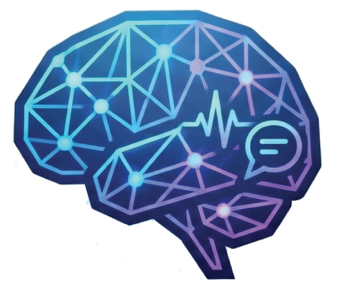}}
\textcolor{cerebrablue}{Website \& Demo}}

\end{center}
\begingroup
\renewcommand{\thefootnote}{}
\makeatletter
\renewcommand{\@makefnmark}{}
\makeatother
\footnotetext{%
\noindent\parbox[t]{0.95\linewidth}{%
$^{*}$Co-first authors, $^{\dagger}$Equal contribution, $^{\textrm{\Letter}}$Corresponding authors.\\
Emails: \{shengl, jamesz\}@stanford.edu, narges.razavian@nyulangone.org
}%
}
\addtocounter{footnote}{0}
\endgroup

\begin{abstract}
    Modern clinical practice increasingly depends on reasoning over heterogeneous, evolving, and incomplete patient data. Although recent advances in multimodal foundation models have improved performance on various clinical tasks, most existing models remain static, opaque, and poorly aligned with real-world clinical workflows. We present Cerebra, an interactive multi-agent AI team that coordinates specialized agents for EHR, clinical notes, and medical imaging analysis. These outputs are synthesized into a clinician-facing dashboard that combines visual analytics with a conversational interface, enabling clinicians to interrogate predictions and contextualize risk at the point of care. Cerebra supports privacy-preserving deployment by operating on structured representations and remains robust when modalities are incomplete. We evaluated Cerebra using a massive multi-institutional dataset spanning 3 million patients from four independent healthcare systems. Cerebra consistently outperformed both state-of-the-art single-modality models and large multimodal language model baselines. In dementia risk prediction, it achieved AUROCs up to 0.80, compared with 0.74 for the strongest single-modality model and 0.68 for language model baselines. For dementia diagnosis, it achieved an AUROC of 0.86, and for survival prediction, a C-index of 0.81. In a reader study with experienced physicians, Cerebra significantly improved expert performance, increasing accuracy by 17.5 percentage points in prospective dementia risk estimation. These results demonstrate Cerebra’s potential for interpretable, robust decision support in clinical care.

\end{abstract}

\section{Introduction}
Modern clinical decision-making is inherently multimodal. Physicians routinely integrate information from electronic health records, laboratory measurements, medical imaging, and unstructured clinical notes to assess disease risk, guide diagnostic evaluation, and plan treatment~\cite{Jensen2012MiningEH, Bates2014BigDI, Miotto2016DeepPA, Esteva2019AGT, Wang2018ClinicalIE, tu2024towards}. In practice, however, these data sources are often incomplete, unevenly sampled, and heterogeneous across patients~\cite{rajkomar2018scalable, Weiskopf2013MethodsAD}. Clinical presentations vary widely, with data availability differs by care setting, and relevant evidence may be distributed across disparate systems~\cite{, Johnson2016MIMICIIIAF, Johnson2023MIMICIVAF}. Despite these challenges, clinicians must make timely, high-stakes decisions, often under considerable uncertainty.

This complexity is particularly evident in neurology, where symptoms such as memory complaints are common, nonspecific, subtle, and arise from a wide range of etiologies~\cite{jack2018nia, dubois2016preclinical, Reid2006SubjectiveMC, garand2009diagnostic, Petersen2018PracticeGU, mitchell2014risk, schneider2007mixed}. Despite consistent and explicit guidelines~\cite{gauthier2006mild, albert2011diagnosis, sperling2011toward} on neurodegenerative diseases treatments, the clinical practice is not often done reliably. For cognitive impairment diagnosis, only a subset of patients presenting with cognitive concerns progress to neurodegenerative dementia (Alzheimer's disease and Alzheimer disease related dementias, AD/ADRD)~\cite{paulose2021national, national2017health, sonnega2014cohort}, yet neurology clinics face substantial referral burden and diagnostic uncertainty~\cite{Stone2010WhoIR, Bradford2009MissedAD}. Determining which patients in real-world populations warrant specialist evaluation, advanced diagnostic imaging, longitudinal monitoring, or prioritized diagnostic testing remains a challenge~\cite{Wippold2015ACRAC, Soderlund2025ACRAC, Johnson2013AppropriateUC, rabinovici2025updated}. Addressing these problems requires not only accurate risk estimation, but also transparent integration of heterogeneous clinical evidence that clinicians can inspect, interrogate, and contextualize.

Recent advances in medical foundation models and multimodal machine learning have improved performance on individual prediction tasks, such as disease classification or outcome forecasting~\cite{wen2020convolutional, xue2024ai, huang2019diagnosis, ding2019deep, huang2020multimodal, moor2023foundation, liu2020design, liu2021development, liu2022generalizable}. However, most existing approaches rely on static, task-specific pipelines that are brittle in the face of missing data and poorly aligned with real-world clinical workflows. These systems often collapse diverse inputs into a single opaque prediction, providing limited insight into how different data modalities contribute to risk~\cite{ghassemi2021false, kelly2019key, amann2020explainability, chen2023automatic}. As a result, clinicians are left with black-box outputs that are difficult to interpret, limiting trust and adoption in practice~\cite{tonekaboni2019clinicians, rudin2019stop}.

Agentic AI systems driven by large language models have emerged as a promising paradigm for coordinating tools, models, and data sources in flexible ways~\cite{yao2022react, xu2023gentopia, xu2023rewoo, liang2024taskmatrix, lu2024multimodal, luoctotools}. In principle, such systems could adapt to variable data availability, orchestrate multimodal analyses, and support interactive clinical workflows~\cite{zhao2025agentic}. In medical settings, however, existing agentic and multimodal AI approaches, while enabling tool-augmented reasoning and prediction~\cite{luoctotools, schick2024toolformer, shen2023hugginggpt}, remain constrained by limited interpretability and lack mechanisms for effective human–AI collaboration in clinical settings~\cite{zhao2026ai, topol2019high,  amann2022explain}.

To address these challenges, we introduce \CEREBRA, a multimodal agentic AI system designed to support transparent, clinician-guided dementia risk assessment in real-world clinical settings. Rather than functioning as a single monolithic predictor, \CEREBRA decomposes risk assessment across multiple data modalities, including structured health records, medical imaging, and unstructured clinical notes, and aggregates their outputs into an interpretable, patient-level risk profile. This modular design allows \CEREBRA to adapt naturally to heterogeneous and partially missing data in real clinical practices, while preserving modality-specific reasoning and fusion across modalities that can be examined by clinicians.

\CEREBRA presents its analyses through an interactive dashboard that combines visual summaries with a conversational interface, enabling clinicians to explore results, query supporting evidence, and navigate modality-specific findings. Through this interface, clinicians can ``interrogate'' why a patient is assessed as high or low risk, examine the contributions of individual data sources, and request targeted summaries of relevant clinical evidence. Importantly, this interaction guides how results are explored and interpreted, without altering the underlying predictive models or compromising analytical integrity.

\CEREBRA is also designed to be continuously evolving with physician feedback. Rather than enhancing the agent's reasoning on a case-by-case basis, \CEREBRA collects physician feedback and suggestions into a constantly updating medical knowledge notebook to allow improved performance for future cases. This feature can not only fill in the gap on clinically-limited knowledge in AI systems, but also allow tailored medical recommendations for flexibility across medical systems with heterogeneous clinical resources.

We demonstrate \CEREBRA's utility through comprehensive evaluations focused on dementia risk stratification and triage-relevant assessment. Using large-scale, multimodal clinical data, we show that \CEREBRA accurately identifies patients at elevated risk of dementia along with their related outcomes such as speed of progression and subtype, while providing transparent explanations grounded in heterogeneous clinical evidence. These results illustrate how agentic, multimodal AI systems, when coupled with clinician-centered interaction, show promise as a clinician-facing decision-support framework for dementia risk assessment.

\begin{figure}
    \centering
    \includegraphics[width=0.97\linewidth]{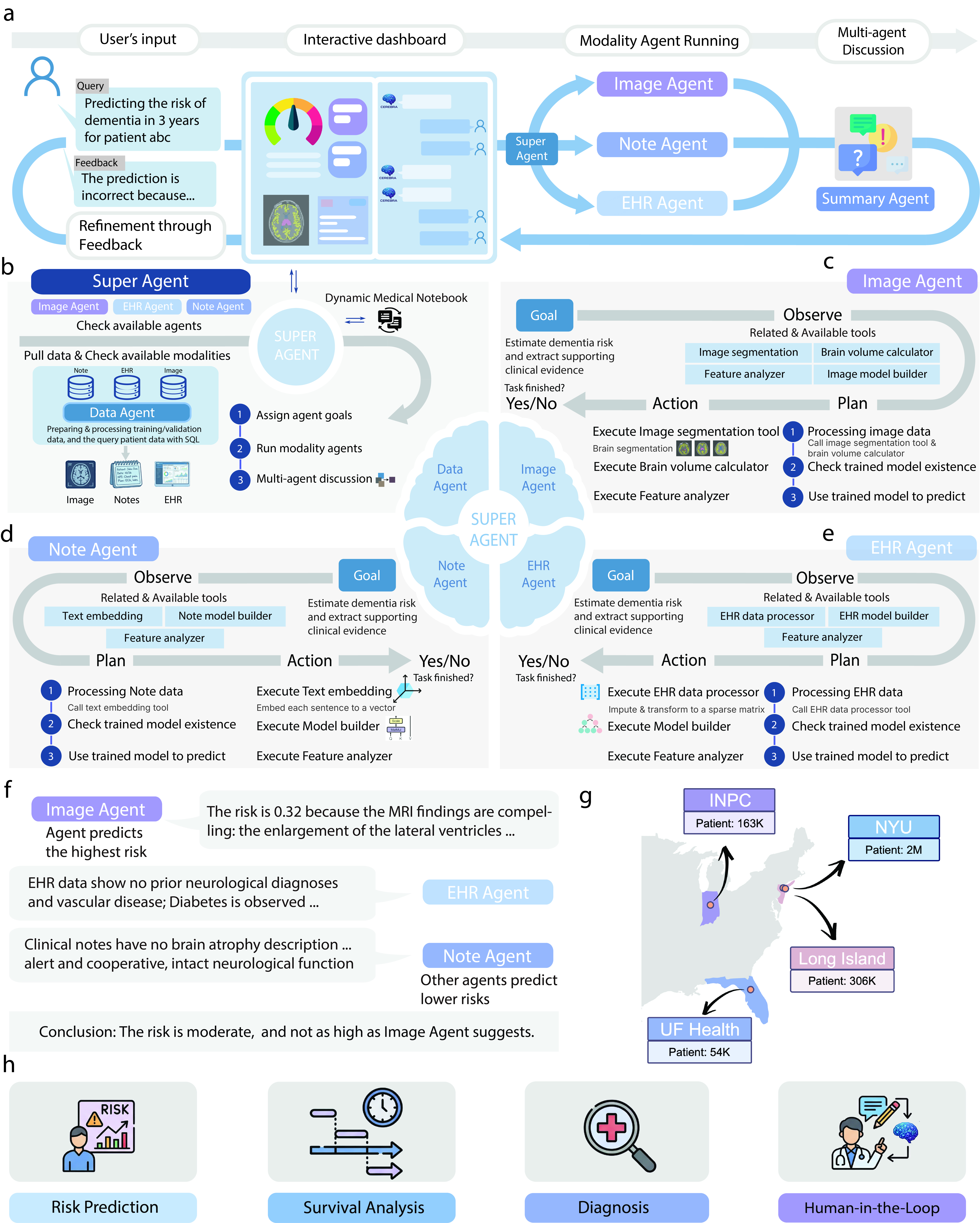}
    \caption{\textbf{Overview of the \CEREBRA agentic AI system for clinical decision support.} 
\textbf{a.} Clinicians interact with \CEREBRA through an interactive dashboard that accepts queries and feedback. Modality-specific agents analyze patient data and participate in a structured multi-agent discussion coordinated by a summary agent to produce a final conclusion. 
\textbf{b.} A super agent moderates the data agent and modality agents.
\textbf{c–e.} Modality agents operating under an independent workflow, training models, and producing modality-specific evidence and results.
\textbf{f.} The summary agent performs a multi-agent discussion in a propose-and-critique style.
\textbf{g.} Multi-institutional data integration and cohort construction.
\textbf{h.} \CEREBRA supports various clinical tasks, including risk prediction, survival analysis, diagnostic assistance with human-in-the-loop.} 
    \label{fig:overview}
\end{figure}
\section{Results}
\subsection{Building a multimodal agentic AI system for dementia risk assessment}
We developed \CEREBRA, a multimodal agentic AI system designed to support interactive, evidence-based dementia characterization, prognosis and risk assessment by integrating heterogeneous clinical data into a clinician-facing analytical workflow. The system is motivated by a central challenge in cognitive care: healthcare systems, particularly those serving aging populations, face growing demand while specialist resources remain limited. As a result, many patients with cognitive concerns are initially evaluated by non-specialists who must rapidly determine whether symptoms warrant specialist referral, additional testing or longitudinal monitoring based on complex and fragmented patient records.

To address this challenge, \CEREBRA organizes clinical reasoning through coordinated agents that retrieve, analyze and synthesize information across data modalities. Given a clinician's query specifying a clinical task through the interactive dashboard, for example predicting the risk of dementia within a defined time horizon, assessing current diagnostic status or estimating disease progression. \CEREBRA first activates a data agent responsible for identifying and retrieving the relevant patient information sources (Fig.~\ref{fig:overview}b). This agent determines the appropriate temporal window, available data modalities, and prediction horizon, enabling the system to flexibly support different clinical decision-making scenarios without manual reconfiguration.

\CEREBRA then decomposes the task into modality-specific analyses coordinated by a central super agent (Fig.~\ref{fig:overview}b). Dedicated modality agents (Fig.~\ref{fig:overview}c-e) independently process different patient modalities, internally train and validate models when needed following machine learning best practices in the field, treating these learned models as reusable analytical tools within each modality agent, and produce an intermediate probability estimate together with structured evidence supporting their decision. By utilizing mainstream and state-of-the-art machine learning approaches for each modality in real-life clinical settings, we demonstrate \CEREBRA's ability in fusing complementary information from each modality agent evidenced by the superior performances across tasks and medical sites. Rather than collapsing all inputs into a single opaque model, the super agent aggregates these modality-level outputs to generate a final task-specific risk analysis while preserving modality-specific contributions and supporting evidence. This design allows \CEREBRA to maintain transparency in its decision-making process and to operate robustly under heterogeneous or partially missing data conditions.

To test the flexibility of \CEREBRA, we benchmarked the system in an elderly population on the diagnoses of different levels of cognitive impairments. Specifically, we define patients as dementia when they are presented with AD/ADRD defining diagnostic or medication medical records from electronic health records (EHR), those as mild cognitive impairment (MCI) when they have corresponding MCI-related records, and those as normal cognition (NC) if they are presented neither dementia nor MCI conditions (Detailed definition in~Supplementary Table.~\ref{tab:ad_criteria}). We then experimented on a variety of clinical tasks, including prediction of risks of conversion from normal cognition\footnote{Normal cognition was defined as the absence of both MCI and ADRD diagnosis codes in the patient’s historical medical record.} to dementia, specifically Alzheimer’s disease and Alzheimer’s disease related dementias, or AD/ADRD, and risk of conversion from MCI to dementia. \CEREBRA can also perform dementia diagnosis, and survival analysis. These tasks were developed and benchmarked on data from four institutions and affiliations, namely NYU Langone Hospital (NYU), NYU Langone Hospital - Long Island (LI), University of Florida Health (UF), and Indiana University/Regenstrief Institute (INPC), with exposure to 3 million patients and \textgreater{}100K selected cohort population as reported in Fig.~\ref{fig:overview}g. These four sites possess diverse population characteristics, demonstrating \CEREBRA's adaptability and generality in practical use. Data were split at the patient level into training, validation, and test sets (80/10/10), with additional out-of-distribution evaluation performed across institutions.  Details of demographics and distribution of ADRD disease subtypes across the four sites are reported in Table.~\ref{tab:study_population}.

A useful feature of \CEREBRA is its interactive dashboard, which enables clinicians to explore multimodal risk assessments and other tasks in a unified interface. The dashboard presents overall dementia risk trajectories alongside modality-specific risk factors. For each modality, \CEREBRA exposes representative features and clinical snippets that support the various medical tasks, enabling direct inspection of the evidence underlying the model’s conclusions.

Beyond static visualization, \CEREBRA supports clinician-guided interaction through a conversational interface embedded within the dashboard. Through this interface, clinicians can query the system to retrieve supporting evidence, request targeted summaries of specific clinical factors, and navigate across modalities to contextualize probability estimations for various tasks. This interaction allows clinicians to interrogate and interpret model outputs in a manner aligned with clinical reasoning, without modifying the underlying analytical pipeline or predictive models.

Importantly, this interactive framework supports clinical triage and diagnostic prioritization. By distinguishing patients with elevated multimodal risk for ADRD from those without, \CEREBRA helps clinicians focus attention and resources on patients most likely to benefit from further ADRD-specialized care. In addition, modality-specific risk patterns provide information that may support the choice of diagnostic tests, such as imaging, laboratory evaluation, or longitudinal monitoring.

Together, these results demonstrate that \CEREBRA extends dementia characterization and risk modeling beyond static predictions toward an interactive decision-support paradigm. By combining agentic multimodal analysis with clinician-centered exploration of risk and evidence, \CEREBRA enables more informed, efficient, and transparent clinical assessment in settings characterized by uncertainty, heterogeneous data, and high referral burden.

\begin{figure}
    \centering
    \includegraphics[width=1\linewidth]{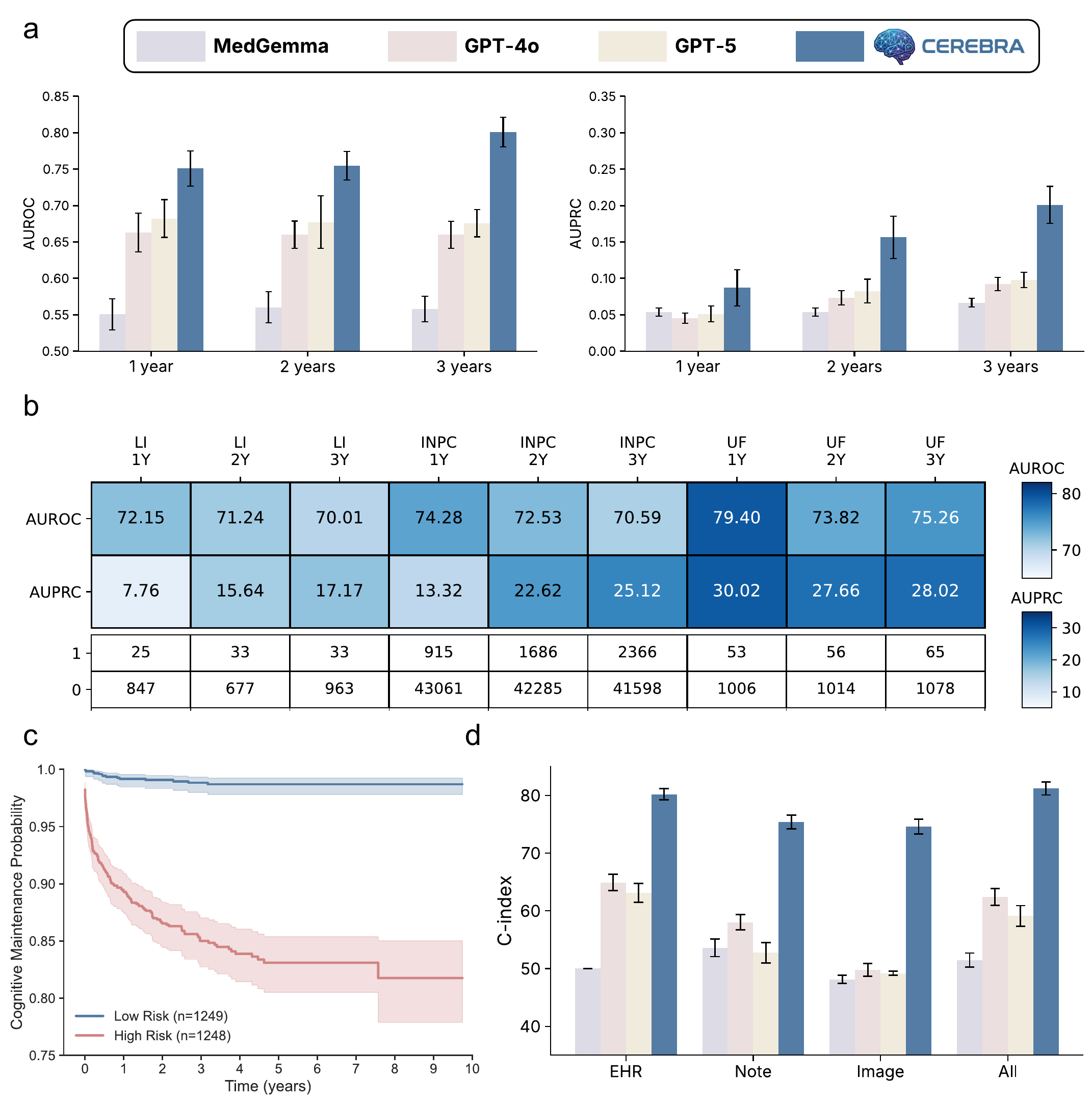}
    \caption{
    \textbf{Dementia risk prediction performance of \CEREBRA.} 
    \textbf{a.} Dementia risk prediction on the NYU cohort for prediction horizons of 1, 2, and 3 years. Performance is measured using AUROC (left) and AUPRC (right), comparing \CEREBRA with MedGemma, GPT-4o, and GPT-5 baselines.
    \textbf{b.} External validation across independent healthcare systems (NYU Langone Hospital - Long Island, Indiana University, and University of Florida). Heatmaps show AUROC and AUPRC for 1-, 2-, and 3-year prediction tasks, with cohort statistics shown below; rows labeled 1 and 0 denote the number of positive (dementia diagnosis onset) and negative (no dementia diagnosis) samples, respectively.
    \textbf{c.} Kaplan–Meier survival analysis stratified by \CEREBRA-predicted risk groups (top and bottom 25\%).
    \textbf{d.} C-index from Cox proportional hazards modeling performed by \CEREBRA across modalities (EHR, clinical notes, imaging) and their multimodal combination. Error bars indicate uncertainty estimated via bootstrap resampling of the test set (1,000 replicates); error bars indicate $\pm$ 1 standard deviation for AUROC and AUPRC.
    }
    \label{fig:prediction_results}
\end{figure}

\begin{table}[t]
\centering
\caption{Glossary related to cognitive conditions}
\label{tab:glossary}
\begin{tabular}{c|c|l}
\toprule
\makecell[c]{\textbf{Category}\\\textbf{Label}} & \textbf{Acronym} & \textbf{Description} \\
\cline{1-3}
NC  & NC  & Normal cognition \\
\cline{1-3}
MCI & MCI & Mild cognitive impairment \\
\cline{1-3}
\multirow{8}{*}{AD/ADRD}
& AD & Alzheimer’s disease \\
\cline{2-3}
& LBD & \makecell[l]{Lewy body dementia, including dementia with Lewy bodies and Parkinson's\\ disease dementia} \\
\cline{2-3}
& VD & \makecell[l]{Vascular dementia, vascular brain injury and vascular dementia,\\ including stroke} \\
\cline{2-3}
& FTD & \makecell[l]{Frontotemporal lobar degeneration and its variants, including primary\\ progressive, aphasia, corticobasal degeneration and progressive supranuclear \\palsy, and with or without amyotrophic lateral sclerosis} \\
\cline{2-3}
& Others & \makecell[l]{Other dementia conditions, including amnestic dementia, degenerative\\ disease of nervous system, and general unspecified dementia} \\
\bottomrule
\end{tabular}
\end{table}

\subsection{Prediction of future dementia diagnosis}
\label{sec:prediction}
We evaluated \CEREBRA for predicting future dementia diagnosis among individuals without pre-existing cognitive impairment diagnoses, across three forecasting horizons (1/2/3 years) and three modalities (EHR, clinical notes, and imaging where available). As shown in Fig.~\ref{fig:prediction_results}a, \CEREBRA performs better (AUROCs of 0.751, 0.755 and 0.801, AUPRCs of 0.087, 0.156 and 0.201 for 1/2/3 years prediction horizon respectively) in most experiments than the the performance of the best performing modality agent, which employs mainstream or state-of-the-art single modality machine learning approaches (AUROCs of 0.719, 0.719 and 0.735, AUPRCs of 0.095, 0.122 and 0.174 for 1/2/3 years prediction horizon respectively) in the NYU cohort.

In addition, we conducted two complementary evaluations. First, we benchmarked \CEREBRA against Large Language Model (LLM) baselines on the NYU Langone cohort, with the exact same experimental setup except the data is verbalized and used in the prompt (detail in Appendix.~\ref{appendix:llm_baseline}), with \CEREBRA consistently outperforming all LLMs. Second, to evaluate robustness and generalization, we trained and evaluated \CEREBRA independently with modality data (EHR, note, and image, if available) from different sites (details in Sec.~\ref{sec:modality_agents_method}), without cross-site mixing or site-specific tailoring. Across sites, AUROC and AUPRC indicate stable performance across horizons and modalities. Notably, we evaluated \CEREBRA on real-world, imbalanced clinical data from four independent sites with a low positive rate of $5\%$, showing the efficacy of the agent in actual medical practices.

\paragraph{Comparison of predictive performance with multimodal LLM baselines.}
As shown in Fig.~\ref{fig:prediction_results}a, we report AUROC for dementia risk prediction at 1/2/3-year forcasting windows for \CEREBRA comparing with the three multimodal LLM baselines, namely, OpenAI GPT-4o~\cite{openai_gpt4, openai_chatgpt4o} (GPT-4o, model version: 2024-11-20), OpenAI GPT-5~\cite{openai_gpt5_system_card_2025} (GPT-5, model version: 2025-08-07) and MedGemma-4B~\cite{sellergren2025medgemma} (MedGemma). Details of our prompting approach are included in Appendix~\ref{appendix:llm_baseline}. The baseline models perform similarly across all forcasting horizons, indicating limited differentiability in this pre-diagnostic prediction setting (details in Appendix.~\ref{appendix:llm_baseline}. In contrast, \CEREBRA consistently achieves significantly higher AUROC (1-yr: 0.751, 2-yr: 0.755, 3-yr: 0.801) at every horizon, demonstrating a substantial and robust improvement over the strongest baselines. To objectively evaluate performance on highly imbalanced real-world data, we also evaluated performance on AUPRC (Fig.~\ref{fig:prediction_results}a). Consistent with the AUROC results, \CEREBRA achieves substantially higher AUPRC (1-yr: 0.087, 2-yr: 0.156, 3-yr: 0.201) across all forecasting windows. While absolute AUPRC values are modest due to the low prevalence of positive cases, these results represent substantial improvements over baseline methods under severe class imbalance.

\paragraph{Comparison of prediction performance among four external study sites.}
We assessed \CEREBRA’s robustness and cross-site generalizability by evaluating it independently at four health systems. As shown in Fig.~\ref{fig:prediction_results}b, \CEREBRA demonstrates consistent differentiation power on predictive dementia risks across heterogeneous real-world settings, achieving AUROC from $0.700$ - $0.794$ across LI, INPC and UF similar to the performance on NYU data, despite large differences in cohort size and dementia prevalence (positive rate $2.87\%$ – $8.53\%$, which matches the prevalence statistics reported in previous studies ranging from 0.81\% to 10\%~\cite{prince2015world, manly2022estimating, kramarow2024diagnosed, spargo2023estimating}) as well as available modalities (comparing with NYU and LI, UF has optical coherent tomography (OCT) imaging as imaging data, while INPC dataset does not have image data available). To further assess the generality in practice, we compute AUPRC from $0.078$ to $0.300$. Given the low disease prevalence, the random baseline AUPRC is 2.87–8.53\%, making absolute AUPRC values appear low even for useful models. Our AUPRC of $0.076$ to $0.300$ is around $2.6 \times$ to $6\times$ above the random baseline, indicating strong enrichment of true cases among top-ranked patients, exactly what matters for real-world screening under class imbalance. To understand the performance attribution across modalities, we report metrics on modality agents for predictive tasks in Supplementary Table.~\ref{tab:prediction_results}, where different sites showing diversified modality-specific performances, demonstrating \CEREBRA's ability in fusing heterogeneous information from modalities where insightful evidences are presented.

\subsection{Longitudinal disease prognosis and survival modeling}
Beyond fixed-term risk prediction, we evaluated \CEREBRA’s capacity for longitudinal risk assessment using survival analysis and time-to-event modeling. Similarly as observed in the previous setting, \CEREBRA achieves superior concordance index (C-index) performance of 0.812 comparing to single modality approaches (best performance achieved by EHR using the SoTA survival model~\cite{spooner2020comparison} with C-index of 0.782 $\pm$ 0.096). In addition, \CEREBRA achieved robust risk stratification and outperformed existing large language models in time-to-event predictive accuracy.

\paragraph{Risk stratification via Kaplan--Meier analysis.} 
\CEREBRA successfully stratified patients into distinct prognosis groups based on predicted risk scores. As shown in Fig.~\ref{fig:prediction_results}c, the Kaplan--Meier survival curves reveal a significant divergence in survival probabilities between the ``Low Risk'' and ``High Risk'' cohorts in the holdout set. This separation remains consistent and statistically significant over an extended follow-up period, confirming the model's ability to identify high-risk cohorts long before clinical deterioration occurs.

\paragraph{Comparative benchmarking of C-Index.} 
We further compared \CEREBRA's survival prediction performance against state-of-the-art baselines, including MedGemma, GPT-4o, and GPT-5, across unimodal (EHR, Clinical Notes, Imaging) and multimodal inputs (Fig.~\ref{fig:prediction_results}d). \CEREBRA consistently outperformed all comparator models in every category. \CEREBRA achieved a C-index  $0.812$, significantly surpassing the next best performing model (GPT-4o, C-index $0.649$). The modality agents also outperform the baseline LLMs with the corresponding single modality as inputs. 

\begin{figure}[t]
    \centering
    \includegraphics[width=.9\linewidth]{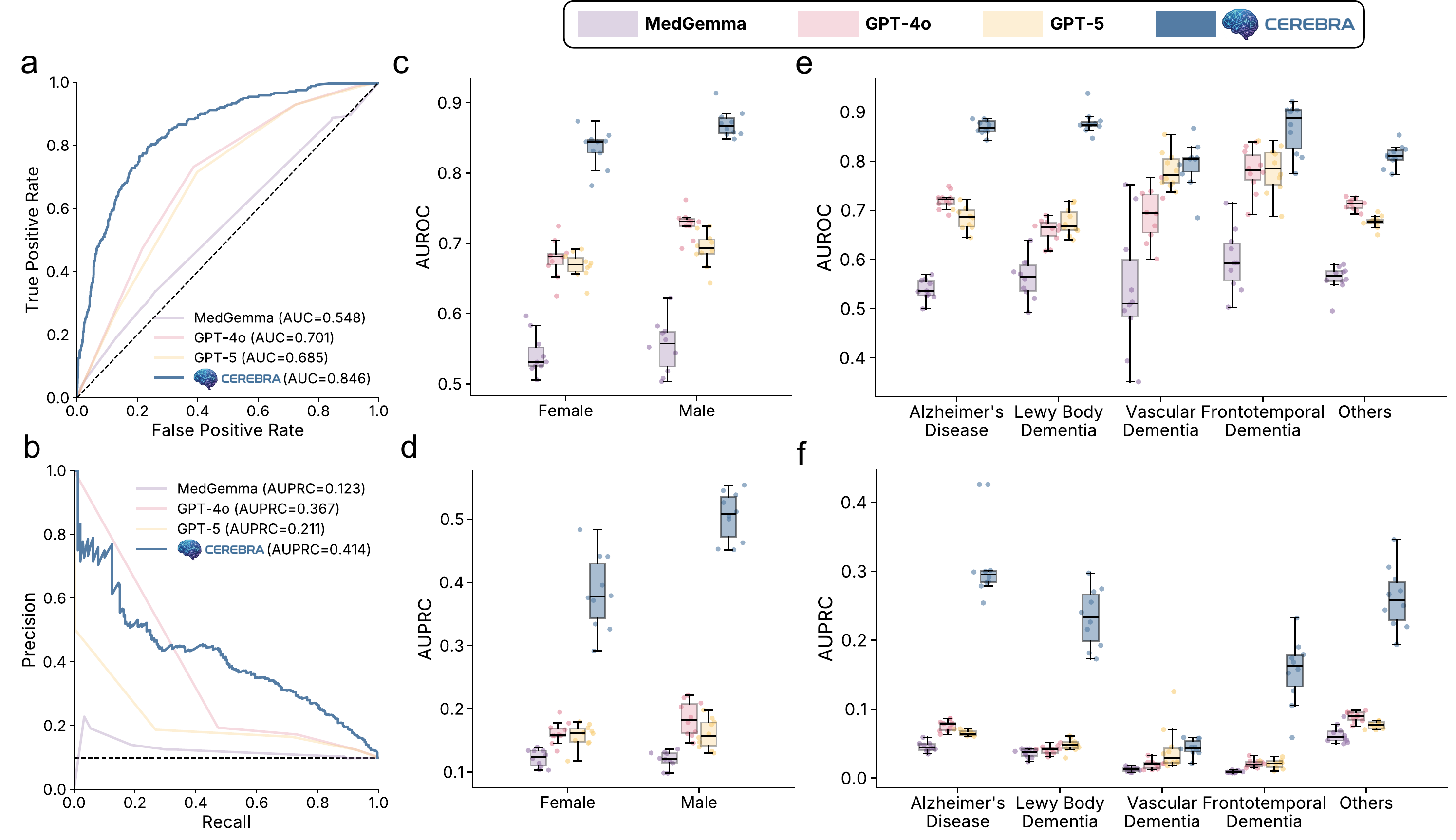}
    \caption{\textbf{Dementia diagnosis performance of \CEREBRA in NYU cohort}. \textbf{a-b.} Receiver operating characteristic (ROC) curves and Precision–recall curves for dementia diagnosis on the NYU cohort, comparing \CEREBRA with MedGemma, GPT-4o, and GPT-5. \textbf{c-d.}  Performance stratified by sex, shown using the area under the ROC curve (AUROC) and the area under the precision–recall curve (AUPRC). \textbf{e-f.} Performance stratified by dementia subtype (Alzheimer’s disease, Lewy body dementia, vascular dementia, frontotemporal dementia, and others), shown using AUROC and AUPRC. Each dot denotes one bootstrap estimate of model performance ($n = 100$ resamples); boxplots summarize the distribution of these bootstrap estimates.}
    \label{fig:diagnosis}
\end{figure}

\subsection{Dementia diagnosis across subtypes}
We evaluate \CEREBRA's capability to perform dementia diagnosis as another clinically relevant task. Similarly to the predictive tasks, we assess diagnostic performance by comparing \CEREBRA with single-modality models across multiple medical institutions (Supplementary Table~\ref{tab:dagnosis_results}). \CEREBRA achieved the best AUROC across all sites (NYU: +4.1, LI: +2.5, INPC: +1.18, UF: +1.1 versus the best unimodal agent baseline) and the best AUPRC at INPC (+0.48) and UF (+1.5), while remaining close to the top unimodal model at NYU and LI.

\paragraph{Comparison of diagnosis performance with advanced LLM baselines.}
As shown in Fig.~\ref{fig:diagnosis}a, \CEREBRA achieves the highest discriminative performance for dementia diagnosis, substantially exceeding all three LLM baselines. The LLM models show only modest discrimination, with AUROC values clustered around $0.548$ – $0.701$, whereas \CEREBRA reaches an AUROC of $0.846$. This clear separation, with low variability across runs, indicates that the agent more reliably distinguishes dementia cases from non-cases than state-of-the-art LLM baselines under the same protocol. We also reported AUPRC (Fig.~\ref{fig:diagnosis}b) to assess performance on highly imbalanced real-world data. The LLM baselines achieve similarly low AUPRC ($0.123$ – $0.367$), whereas \CEREBRA attains a markedly higher AUPRC of $0.414$, indicating substantially better enrichment of true dementia cases among individuals prioritized as high risk under the same evaluation scheme.

To further analyze \CEREBRA's effectiveness for clinical use, we assessed performance across gender demographics and dementia subtypes (Fig.~\ref{fig:diagnosis}c–f). Dementia subtype definitions are provided in Supplementary Table~\ref{tab:subtype_definition}. Compared with baseline large language models, \CEREBRA demonstrates superior AUROC and AUPRC for both female and male populations, as well as for cohorts corresponding to different dementia subtypes, including Alzheimer’s disease, Lewy body dementia, vascular dementia, frontotemporal dementia, and other unspecified dementia.

\begin{figure*}[!t]
    \centering
    \includegraphics[width=1.\linewidth]{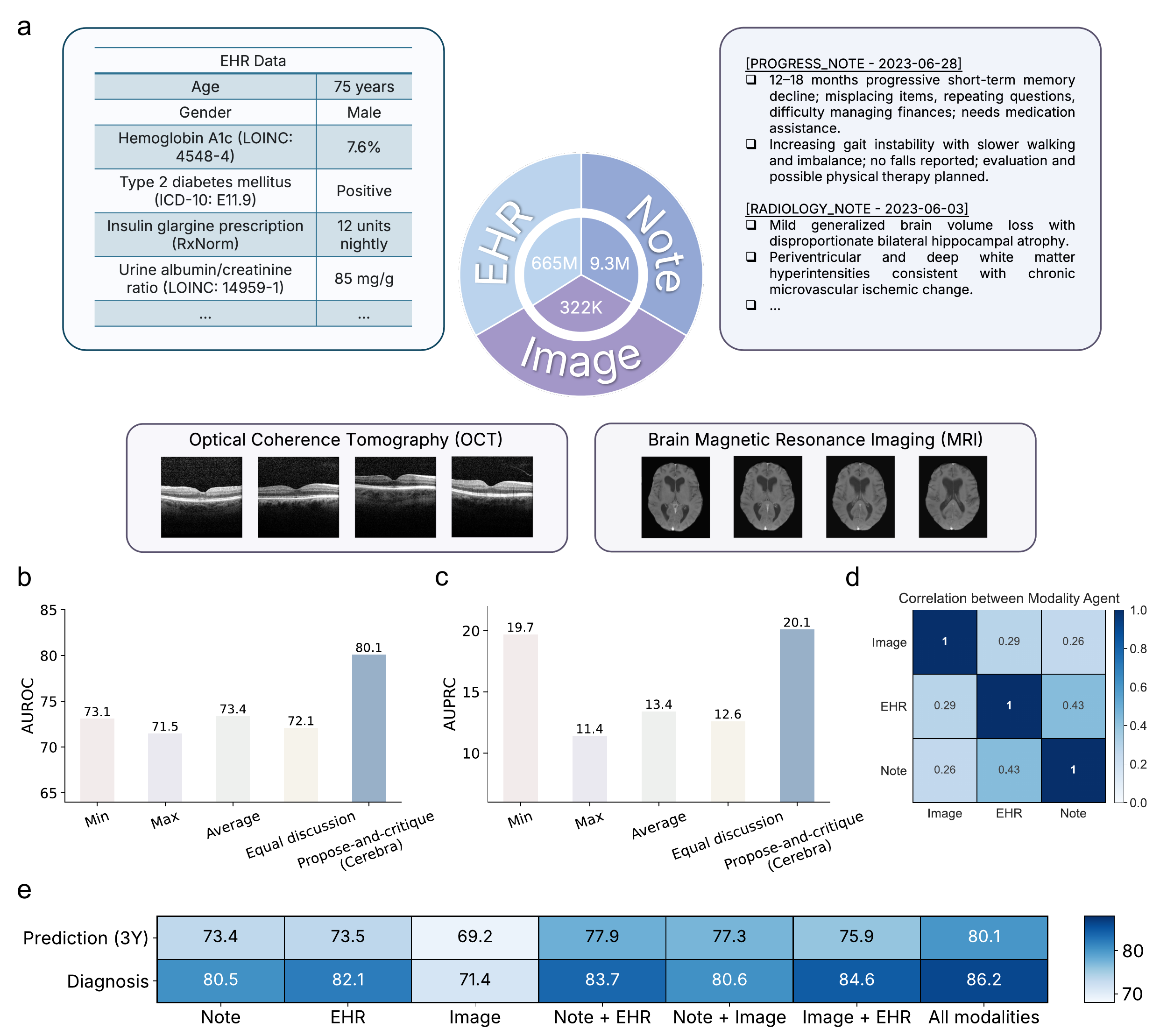}
    \caption{\textbf{Evaluation of multimodal fusion and missing-modality robustness in \CEREBRA.} \textbf{a.} Multimodal inputs used by \CEREBRA, including structured EHR data, clinical notes, and medical images (OCT and brain MRI), with the number of samples available for each modality. \textbf{b–c.} Risk prediction (3 Year) performance comparing heuristic fusion methods based on minimum (lowest-risk), maximum (highest-risk), and average risk scores with agentic equal discussion and propose-and-critique style discussion.
\textbf{d.} Correlation between modality agents' risk scores, showing complementary (non-redundant) risk assessments across image, EHR, and note agents.
\textbf{e.} Performance of \CEREBRA on the NYU cohort with missing modalities. 
    }
    \label{fig:modality}
\end{figure*}

\subsection{MCI to ADRD conversion risk prediction.}
\label{sec:conversion_prediction}
In addition to forecasting dementia risk in non-diagnostic cohorts (i.e. general older-adult populations), we also evaluate \CEREBRA on MCI-to-ADRD conversion, a clinically common ``next-step'' question that arises once impairment is already suspected. Here, conversion refers to progression from MCI, often a prodromal stage with heterogeneous causes, to ADRD, indicating transition to overt dementia. This task complements normal-to-dementia prediction by testing whether the model can capture stage-to-stage progression and detect subtler, near-term signals of imminent decline, which is typically more challenging.

As summarized in Supplementary Table.~\ref{tab:conversion_prediction_results}, predictive performance improved with longer follow-up windows. \CEREBRA achieves AUROC scores of $0.641$, $0.665$, $0.699$, and AUPRC scores of $0.237$, $0.323$, $0.533$ across 1-, 2- and 3-year horizons, surpassing the best performing single modality result (EHR agent, with AUROCs of $0.623$, $0.645$, $0.668$, and AUPRCs of $0.207$, $0.312$, $0.504$). These findings demonstrate stable generalization of the proposed \CEREBRA framework for the conversion prediction task.

\subsection{Robust multimodal integration under real-world data variability}
In routine clinical settings, patient records are rarely complete or uniform: modality availability varies (EHR vs. notes vs. imaging), longitudinal depth differs across patients, and each modality can be unevenly informative for a given individual. Moreover, modalities may provide partially overlapping signals, or even appear to disagree, because they reflect distinct aspects of cognitive health (e.g., vascular comorbidities in EHR, functional concerns in notes, structural changes in imaging). Because \CEREBRA is explicitly designed to reason over heterogeneous, partially observed modalities, by producing modality-specific hypotheses through modality agents and integrating them through a critique-based multi-agent discussion, we expect it to remain robust when inputs are unevenly informative or missing. To test this in practice, we performed a targeted analysis of (i) how multimodal evidence is fused, (ii) whether modalities contribute complementary information, and (iii) how performance degrades when modalities are missing, summarized in Fig.~\ref{fig:modality}.

\paragraph{Representative examples of the inputs.} We first illustrate representative examples of the inputs available to \CEREBRA across modalities (Fig.~\ref{fig:modality}a). Structured EHR captures longitudinal diagnoses, medication exposures and symptom codes; clinical notes provide narrative descriptions of cognitive and functional status; and imaging modalities (for example, OCT and brain MRI) reflect complementary anatomical changes. These examples highlight the heterogeneity of real-world clinical evidence and motivate the need for modality-aware reasoning when forming patient-level risk assessments. Top dementia-contributing risk factors from respective modality agents are reported in Supplementary Fig.~\ref{fig:modality_agent_evidences}.

\paragraph{Fusion strategy matters beyond score-level heuristics.}
We next compared \CEREBRA’s agentic fusion with simple heuristic aggregation of modality-specific predictions (minimum, maximum, and average), and a standard multi-agent discussion baseline in which each agent contributes in a fixed sequential order with equal participation. (Fig.~\ref{fig:modality}b,c). Across metrics, naive score fusion as well as simple agent equal discussion underperformed, whereas the proposed propose-and-critique style agentic discussion-based aggregation used in 
\CEREBRA (see details in Sec.~\ref{method:summary_agent}) yielded the strongest performance, indicating that multimodal integration requires more than collapsing modality outputs into a single summary statistic.

\paragraph{Complementary modality signals.}
To characterize cross-modality relationships, we computed correlations between risk scores produced by the image, EHR and note agents (Fig.~\ref{fig:modality}d). Correlations were moderate, suggesting that modalities provide overlapping but non-redundant views of dementia risk, consistent with the gains observed from principled fusion.

\paragraph{Graceful degradation with missing modalities.}
Finally, we evaluated \CEREBRA under missing-modality settings by withholding one or more modalities at inference (Fig.~\ref{fig:modality}e). Performance remained competitive when only a single modality was available and improved as additional modalities were added, with the best results achieved when all modalities were present. These results suggest robustness to variable modality availability in retrospective evaluation.

\begin{figure*}[!t]
    \centering
    \includegraphics[width=1\linewidth]{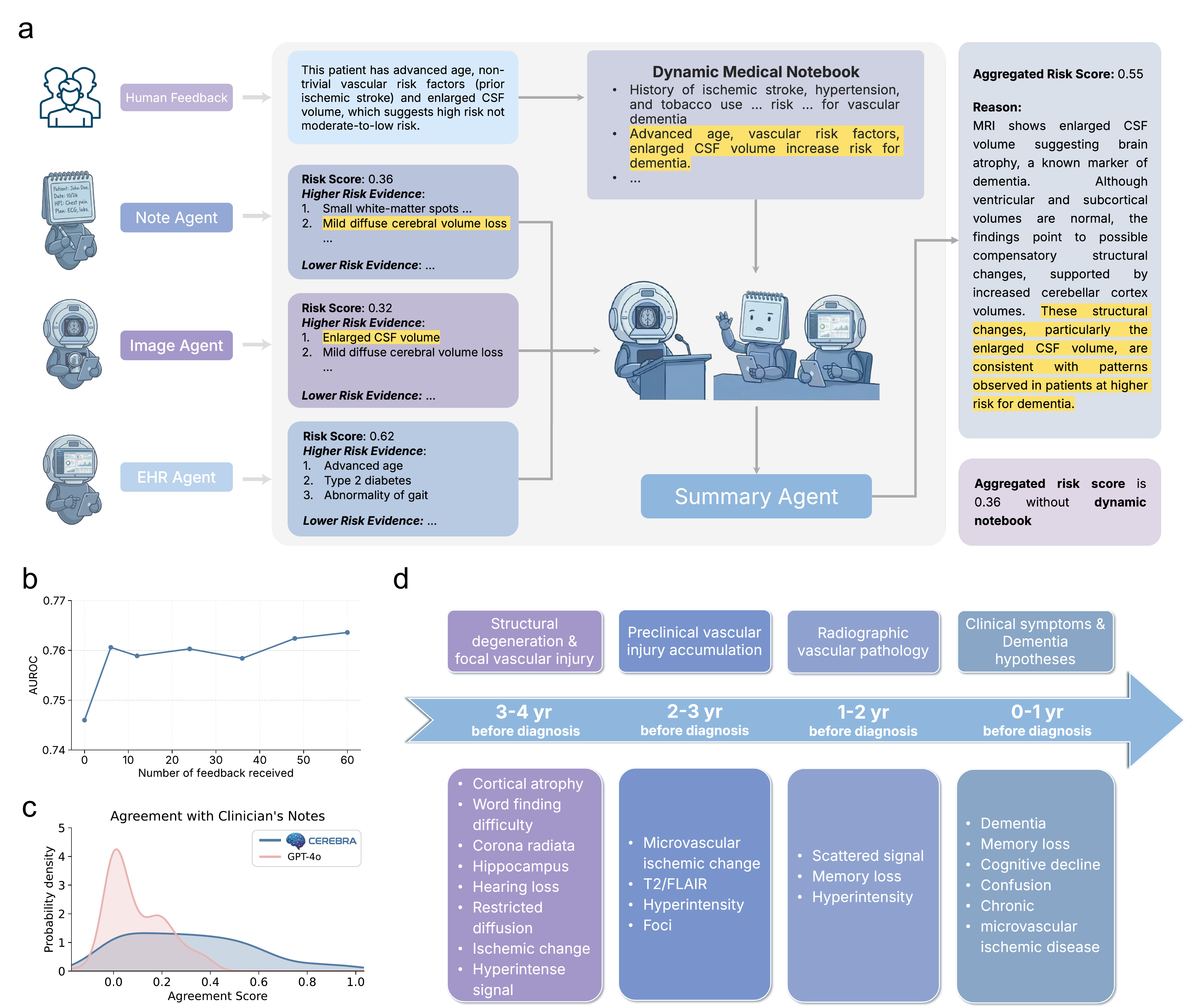}
    \caption{
    \textbf{Feedback loop and reasoning evidence analysis.} 
    \textbf{a.} Demonstration of summary agent pipeline with the dynamic medical notebook. Summary agent collects outputs from modality agents for a multi-agent discussion, while the aggregated risk score is properly calibrated with the prior feedback provided by clinicians in the dynamic medical notebook.
    \textbf{b.} Performance in AUROC on dementia prediction task improves with increasing number of automatically generated feedback from erroneous predictions. 
    \textbf{c.} \CEREBRA's reasoning achieves significantly better agreement with reference clinical notes than GPT-4o.
    \textbf{d.} Qualitative analysis on the most frequent positive keywords across time in \CEREBRA reasoning. Different time horizons manifest differential clinical evidence characteristics, as well as diversified medical keywords.}
    \label{fig:analysis}
\end{figure*}

\subsection{Analysis on extracted evidence and reasoning capability}

Clinical decision-making requires clinicians to identify clinically relevant signals from heterogeneous electronic health records and integrate this evidence through sound clinical reasoning to produce accurate risk assessments. Errors can arise both from failing to identify critical signals and from incorrect reasoning that links evidence to conclusions. Therefore, beyond evaluating end-point predictive accuracy, it is important to assess both the ability to extract clinically relevant evidence and the correctness of the reasoning process used to synthesize this information.

\CEREBRA is designed to support reliable clinical decision-making by jointly optimizing evidence extraction and evidence-based reasoning. In each modality agent, underlying machine learning models first identify candidate clinical signals, while domain knowledge–guided large language models and a final summary agent integrate these signals through structured reasoning to generate risk assessments.

\paragraph{Evaluation of alignment with ground-truth reports.}
We evaluated the alignment between model-generated reports and clinician-authored reports, reflecting consistency in evidence extraction by \CEREBRA and reasoning performed in the clinical workflow (see Methods). In Fig.~\ref{fig:analysis}d, we report the distributions of agreement scores for \CEREBRA with GPT-4o as a baseline. 

\CEREBRA exhibits a right-shifted distribution of percentage agreement compared with GPT-4o, indicating stronger alignment with clinician-authored reports. Notably, perfect agreement is not expected because clinician reports are often written with access to longitudinal follow-up information and disease progression that may not be available at the time of prediction. Despite this constraint, \CEREBRA consistently achieves agreement scores extending into moderate-to-high ranges, demonstrating robust and reproducible alignment with clinician-authored clinical reasoning.

\paragraph{Characteristics of extracted evidence across time horizons.}
We analyzed the evidence extracted by \CEREBRA to examine how diagnostic signals evolve as dementia onset approaches. Samples were grouped according to the time from index date to dementia diagnosis. Distinct stage-dependent patterns emerged as the prediction horizon shortened.

At longer horizons (3--4 years), early neurodegenerative and vascular abnormalities were more prominent, including cortical atrophy, white-matter hyperintensities, and other markers of small-vessel injury, consistent with prior studies showing that structural MRI changes and vascular white-matter burden can be detected years before dementia diagnosis \cite{tondelli2012structural,debette2010clinical}. During the intermediate period (2--3 years), preclinical vascular injury became more evident, characterized by ischemic microvascular foci and persistent T2/FLAIR abnormalities, in agreement with evidence that white-matter hyperintensities are radiological markers of cerebral small-vessel disease and predict later cognitive decline and dementia \cite{debette2010clinical,alber2019white}. Closer to diagnosis (1--2 years), vascular pathology appeared more pronounced, with hyperintensities and scattered signal changes consistent with evolving vascular cognitive impairment or vascular dementia \cite{lambert2018identifying,alber2019white}. Within the final year (0--1 year), overt clinical manifestations dominated, including memory loss, cognitive decline, confusion, and established dementia, often accompanied by chronic microvascular ischemic disease.\cite{wilson2011cognitive,biessels2018cognitive,debette2010clinical}.

These extracted keywords are consistent with prior studies describing neuroimaging biomarkers of Alzheimer’s disease and related dementias, vascular cognitive impairment and white-matter hyperintensities, and clinical manifestations of cognitive decline \cite{johnson2012brain,jack2018nia,gorno2011classification,chui2007subcortical,gorelick2016vascular,wong2022vascular,alber2019white,jahn2013memory,livingston2024dementia}. Together, these results illustrate a progression from early structural and vascular abnormalities to overt clinical symptoms as disease onset approaches, highlighting \CEREBRA’s ability to extract clinically meaningful evidence across disease stages.

\subsection{Refining \CEREBRA through a feedback-compatible experience module}
A fundamental capability of the \CEREBRA architecture is its support for a feedback-compatible framework designed for iterative refinement through the accumulation of validated reasoning patterns. This is achieved through the \textit{Dynamic Medical Notebook}, an externalized experience module that conditions the system’s reasoning on prior evidence and established clinical pathways. The notebook is designed to ingest and store corrective logic, allowing the system to refine its reasoning by referring to a continuously updated repository of clinical patterns.

To evaluate the computational scalability and knowledge integration capacity of this design, we performed an ablation study using automated knowledge distillation. This approach allowed us to simulate high-volume knowledge-transfer scenarios, representing the potential integration of large volumes of corrective clinical reasoning patterns, to observe how the system matures as its memory expands. We extracted corrective strategies from a cohort of mispredicted samples with an LLM and iteratively incorporated these insights into the notebook.

As shown in Fig.~\ref{fig:analysis}b, the integration of this distilled knowledge provided an immediate performance lift, increasing the AUROC from $0.746$ to $0.760$ with only six distilled examples. Performance continued to scale with memory size, reaching a maximum AUROC of $0.763$. These results demonstrate that the \CEREBRA architecture can successfully ingest and scale complex textual evidence to enhance the reliability of its multimodal synthesis. This mechanism serves as a potential ``clinical guardrail'' and channel for clinicians providing guidance to steer the system, standardizing reasoning, and preventing the repetition of past errors.

\begin{figure*}
    \centering
    \includegraphics[width=1\linewidth]{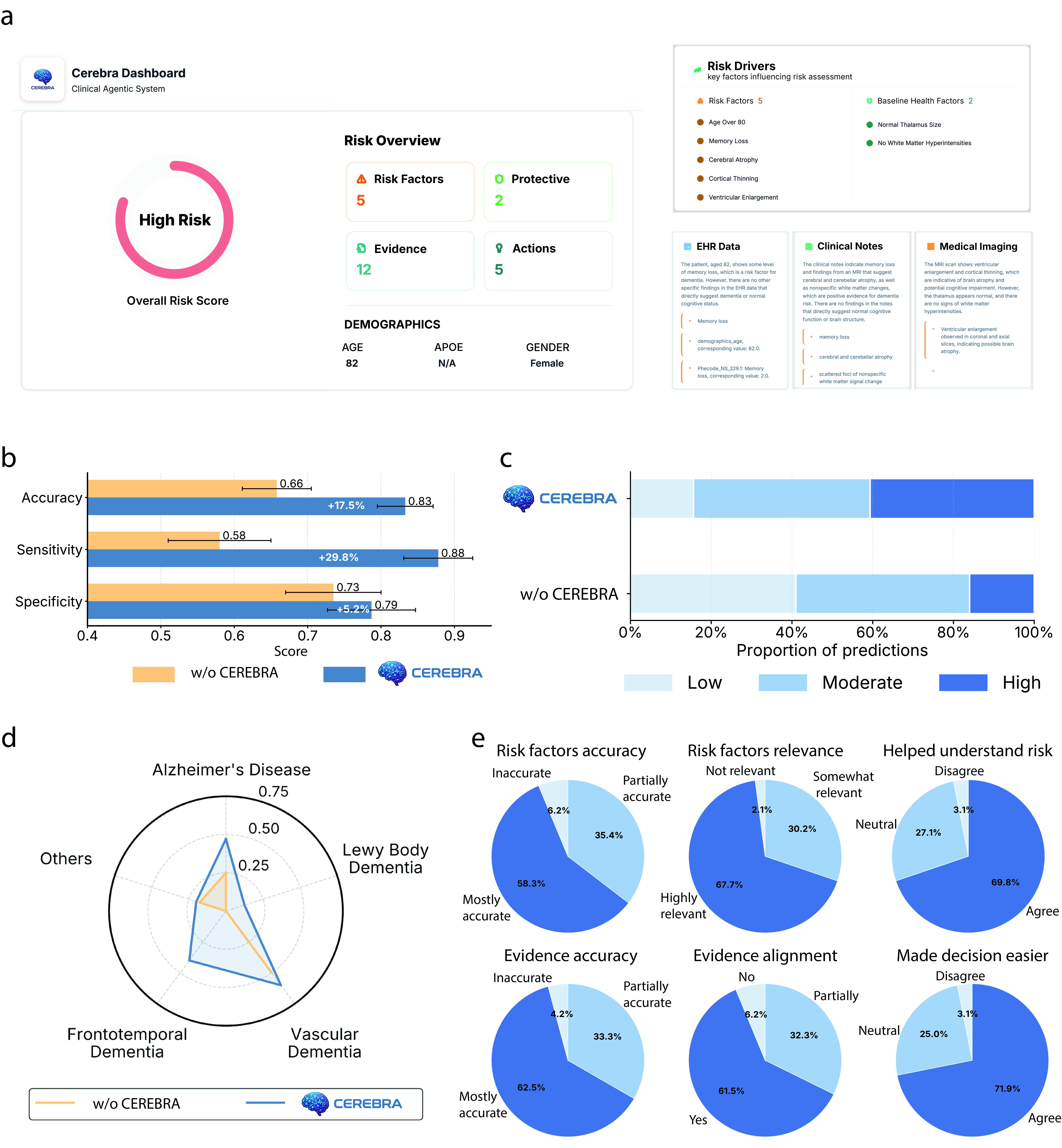}
    \caption{
    \textbf{Clinician reader study evaluating \CEREBRA.}
    \textbf{a.} Example dashboard interface summarizing predicted 3-year dementia risk, extracted risk factors, supporting evidence and suggested actions.
\textbf{b.} Aggregate performance of clinician risk assessments with versus without dashboard assistance (accuracy, sensitivity and specificity; six clinicians; 40 cases per clinician; 20 per condition; $n=120$ case reads per condition). Percent-point changes are annotated.
\textbf{c.} Self-reported confidence distribution (low, moderate, high) with versus without dashboard assistance ($n=120$ per condition).
\textbf{d.} Clinician's performance (sensitivity) stratified by predicted diagnostic category across major dementia etiologies and an \textit{Others} category. \textbf{e.} Clinician evaluation of dashboard content and utility for dashboard-assisted reads ($n=120$): perceived accuracy and relevance of risk factors, accuracy and alignment of supporting evidence, and perceived helpfulness for understanding risk and easing decision-making.}
\label{fig:reader_study}
\end{figure*}

\subsection{\CEREBRA-augmented clinician assessment}
We next evaluated whether \CEREBRA and its clinician-facing dashboard (demonstrated in Fig.~\ref{fig:reader_study}a) improved expert performance on prospective 3-year risk of being diagnosed with dementia. The reader study was conducted with six clinicians, each reviewing 40 held-out cases (20 without \CEREBRA assistance and 20 with assistance; 20 high-risk and 20 low-risk) in a randomized cross-over design. Detailed survey instruments used to record clinician feedback on risk factors, evidence accuracy, and dashboard utility are provided in Section~\ref{sec:reader_study_appendix}.

\paragraph{\CEREBRA assistance improves clinician risk prediction and confidence.} Across readers and cases, access to the dashboard improved clinicians' performance on 3-year risk of being diagnosed with dementia (Fig.~\ref{fig:reader_study}b). Overall accuracy increased from $0.658 \pm 0.047$ without the dashboard to $0.833 \pm 0.038$ with the dashboard (+17.5\% points). Sensitivity saw the most substantial gain, increasing from $0.580 \pm 0.070$ to $0.878 \pm 0.047$ (+29.8\% points), while specificity improved from $0.735 \pm 0.065$ to $0.787 \pm 0.060$ (+5.2\% points). In addition to improved correctness, clinicians exhibited higher self-reported confidence when using the dashboard, with a significant shift toward moderate-to-high confidence assessments compared with reviewing raw data alone (Fig.~\ref{fig:reader_study}c).

\paragraph{Differential Diagnosis Performance} Quantitative analysis of clinician recall reveals a substantial improvement across all dementia subtypes when using the \CEREBRA dashboard (Fig.~\ref{fig:reader_study}d). The most notable gain was observed in Vascular Dementia, where recall increased from approximately 0.25 to nearly 0.75. Significant improvements were also noted in Frontotemporal Dementia and Alzheimer's Disease, suggesting that the dashboard effectively assists clinicians in identifying specific pathological markers that are otherwise easily missed in standard assessments.

\paragraph{Clinician-reported fidelity and clinical utility of \CEREBRA dashboard} Beyond performance comparisons, we qualitatively assessed the fidelity and clinical utility of the \CEREBRA dashboard explanations (Fig.~\ref{fig:reader_study}e). Most clinicians rated the displayed risk factors as mostly accurate (58.3\%) and highly relevant (67.7\%). Supporting evidence was also well-received, with 62.5\% of responses judging it as accurate and 61.5\% finding it fully aligned with neurological interpretations (with only 6.2\% non-alignment). Recommendations were deemed appropriate in all cases, aligning partially or fully with expert care. Crucially, \CEREBRA demonstrated strong clinical utility: 69.8\% of clinicians agreed it improved their understanding of patient risk, and 71.9\% reported that it facilitated easier decision-making. These findings suggest that \CEREBRA was associated with improved performance in our reader study.

\section{Discussion}
In this study, we present \CEREBRA, an LLM-enabled agentic system designed to support dementia risk analysis in real-world clinical environments. The framework integrates heterogeneous clinical data modalities routinely used in patient care, including electronic health records, clinical notes, and medical imaging. Modality-specific agents employ established machine learning approaches, selected based on prior literature and adapted or trained as needed, to generate risk analysis together with interpretable insights grounded in the patient’s medical history. These outputs are subsequently synthesized by a summary agent into an integrated risk assessment. By combining complementary signals across modalities, \CEREBRA enables a more comprehensive characterization of patient health and produces synergistic improvements when information from multiple data sources is integrated. While prior studies~\cite{xue2024ai, huang2020multimodal} have explored multimodal dementia prediction using combinations of imaging and structured clinical variables, most rely on curated research cohorts and do not incorporate both clinical notes and routine EHR data within a unified framework.

A key feature of \CEREBRA is its alignment with real-world clinical workflows operating on patients’ longitudinal medical records. Unlike many existing multimodal prediction systems that assume complete data availability~\cite{huang2019diagnosis, chen2026emad, leming2025differential, prabhu2022multi}, \CEREBRA dynamically activates modality-specific agents conditioned on available patient data, thereby accommodating the heterogeneous and frequently incomplete nature of clinical records. Furthermore, the framework generates structured, modality-level evidence that explicitly characterizes each source’s contribution to the final assessment, enhancing interpretability relative to conventional black-box models that provide only a single prediction without actionable justification.

Another property of \CEREBRA is its design for clinician-feedback integration. Predictions or evidence that clinicians disagree with can be reviewed and stored in a knowledge notebook, enabling the system to iteratively refine its reasoning and adapt to local clinical practice patterns. This design allows the system to evolve alongside clinical workflows rather than remaining static after deployment.

Compared with recent agentic approaches in medicine~\cite{zhao2025agentic, lu2024multimodal}, \CEREBRA adopts a hybrid architecture that combines modality-specific machine learning models with controlled LLM reasoning. Rather than relying solely on LLM prior knowledge to analyze complete patient records, \CEREBRA first extracts structured clinical signals using validated machine learning models and then provides summarized and de-identified information to the LLM for reasoning and synthesis. This design improves robustness while limiting the amount of sensitive data exposed to LLM components, supporting more privacy-conscious deployment in healthcare settings and enabling the potential use of external or LLM services.

Across experiments spanning multiple clinical institutions, \CEREBRA demonstrates three key properties. First, it consistently outperforms strong or SoTA single-modality baselines across diverse tasks. Second, the framework supports multiple clinically relevant objectives, including dementia risk prediction, diagnosis, conversation prediction, and survival analysis, suggesting broad applicability across cognitive health management tasks. Third, the system generalizes across healthcare institutions with minimal architectural changes, highlighting its adaptability to heterogeneous real-world clinical environments.

\subsection*{Limitations}
Despite these promising results, several limitations should be noted.
First, although the current framework constrains LLM reasoning to structured outputs, the approach still relies on general-purpose LLMs for summarization and reasoning. Future work may benefit from incorporating medically specialized LLMs or additional knowledge-grounding mechanisms to further improve reliability. Second, the performance of each modality agent depends on the underlying machine learning model it independently trains for its respective data type. While the models trained are strong for structured data, text, or imaging, advances in domain-specific foundation models may further improve performance. Third, \CEREBRA relies on retrospective dementia modeling using EHR-derived AD/ADRD/MCI diagnoses as prediction targets. Such labels, based on diagnosis codes, may be noisy, incomplete, and temporally imprecise, introducing potential outcome misclassification, despite incorporating diverse coding systems (e.g., LOINC and ATC). Prior work has shown that computable phenotypes integrating structured and unstructured EHR data can improve case identification and provide a more robust foundation for downstream prediction models~\cite{li2023early}. Although \CEREBRA outperforms baseline approaches, future work can adopt more clinically grounded labeling strategies to further strengthen evaluation.
Fourth, although \CEREBRA provides clinically grounded insights alongside predictions, its recommendation capabilities currently remain high-level. Future work could integrate more detailed treatment guidance or decision-support tools tailored to specific clinical pathways. 
Fifth, despite a wide range of modalities being experimented in our study, a more diverse set of biologically relevant dementia risk predictors can be integrated, such as APOE4~\cite{fortea2024apoe4} (a feature that we have already utilized in EHR if available, but yet to be highlighted), polygenic risk scores~\cite{jung2022transferability}, and fluid biomarkers (e.g. plasma phosphorylated Tau 217~\cite{brum2023two}). The addition of these genomics, proteomics, and related molecular biomarkers into the existing architecture could provide a biologically richer extension of \CEREBRA that contributes to more holistic clinical evaluation of patients. Sixth, although \CEREBRA demonstrates strong performance across 1–3 year prediction horizons, longer-term forecasting (e.g., 10 years) would be more clinically meaningful for identifying early- and young-onset dementias. These conditions are clinically distinct, with greater heterogeneity, more atypical presentations, and in some cases stronger genetic contributions, and are often subject to substantial diagnostic delay~\cite{sirkis2022dissecting, loi2023young}. However, such long-horizon prediction is currently limited by insufficient longitudinal data, as available EHR in this study span only 8–13 years across sites, constraining reliable labeling in younger populations.
Finally, our evaluation focused on dementia-related conditions. Although the framework is designed to be generalizable, future studies are needed to assess its applicability across other disease domains and clinical decision-making tasks.

\section*{Data availability}
The datasets used in this study were obtained from multiple healthcare institutions and contain protected health information (PHI). Due to institutional review board (IRB) restrictions and data use agreements with participating institutions, the raw patient data cannot be made publicly available. Access to the data may be granted to qualified researchers subject to approval by the corresponding institutions and appropriate data use agreements. Requests for data access should be directed to the corresponding authors.

\section*{Code availability}
Source code for Cerebra is available at \url{https://github.com/shengliu66/Cerebra}.

\section*{Acknowledgment}
We thank Pan Lu, Bowen Chen, Fan Nie, Aneesh Pappu, Rahul Thapa, Batu El, and other members of the Zou Group for helpful discussions and feedback on this work. We are also grateful to Teresa Phuongtram Nguyen for her clinical advice. We thank Ruining Zhao for her assistance with early system prototyping and initial project discussions. We also thank Liu Tan for her assistance with figure design. L.C., A.M. and N.R. are supported by National Institute of Health, National Institute on Aging awards R01AG085617. AM and NR are also supported by National Institute of Health, National Institute on Aging award R01AG079175.
For the work at UF, we acknowledge the University of Florida Integrated Data Repository (IDR) and the UF Health Office of the Chief Data Officer for providing the analytic data set for this project. Additionally, the Research reported in this publication was supported by the National Center for Advancing Translational Sciences of the National Institutes of Health under University of Florida Clinical and Translational Science Awards UL1TR000064 and UL1TR001427.
For the work at INPC, we would like to thank the Regenstrief Institute Data Core for the provision of data from the Indiana Network for Patient Care (INPC). The work from this site is partially supported by the National Academy of Medicine under Award No. SCON-10001638. The content is solely the responsibility of the authors and does not necessarily represent the official views of the National Academy of Medicine. 
R.F. is supported by the National Institute on Aging of the National Institutes of Health (NIA RF1/R01AG971469) and the National Science Foundation (2123809).
J.Z. is supported by the Chan Zuckerberg Biohub.

\clearpage
\printbibliography

@article{invernizzi2018normative,
  title={Normative data for retinal-layer thickness maps generated by spectral-domain OCT in a white population},
  author={Invernizzi, Alessandro and Pellegrini, Marco and Acquistapace, Alessandra and Benatti, Eleonora and Erba, Stefano and Cozzi, Mariano and Cigada, Mario and Viola, Francesco and Gillies, Mark and Staurenghi, Giovanni},
  journal={Ophthalmology Retina},
  volume={2},
  number={8},
  pages={808--815},
  year={2018},
  publisher={Elsevier}
}

@article{schick2024toolformer,
  title={Toolformer: Language models can teach themselves to use tools},
  author={Schick, Timo and Dwivedi-Yu, Jane and Dessì, Roberto and Raileanu, Roberta and Lomeli, Maria and Hambro, Eric and Zettlemoyer, Luke and Cancedda, Nicola and Scialom, Thomas},
  journal={Advances in Neural Information Processing Systems},
  volume={36},
  year={2024}
}

@inproceedings{luoctotools,
  title={OctoTools: An Agentic Framework with Extensible Tools for Complex Reasoning},
  author={Lu, Pan and Chen, Bowen and Liu, Sheng and Thapa, Rahul and Boen, Joseph and Zou, James},
  booktitle={Workshop on Reasoning and Planning for Large Language Models},
  year={2025}
}

@article{shen2023hugginggpt,
  title={Hugginggpt: Solving ai tasks with chatgpt and its friends in hugging face},
  author={Shen, Yongliang and Song, Kaitao and Tan, Xu and Li, Dongsheng and Lu, Weiming and Zhuang, Yueting},
  journal={Advances in Neural Information Processing Systems},
  volume={36},
  pages={38154--38180},
  year={2023}
}

@article{topol2019high,
  title={High-performance medicine: the convergence of human and artificial intelligence},
  author={Topol, Eric J},
  journal={Nature medicine},
  volume={25},
  number={1},
  pages={44--56},
  year={2019},
  publisher={Nature Publishing Group US New York}
}

@article{zhao2026ai,
  title={AI agent in healthcare: applications, evaluations, and future directions},
  author={Zhao, Lina and Liu, Shengrui and Xin, Tangsiwei and Tan, Jiawen and Wang, Xiaoran and Li, Yafang and Bian, Zihao and Chen, Yiyang and Kong, Fanyi and Bian, Jinwei and others},
  journal={npj Artificial Intelligence},
  volume={2},
  number={1},
  pages={31},
  year={2026},
  publisher={Nature Publishing Group UK London}
}

@article{amann2022explain,
  title={To explain or not to explain?—Artificial intelligence explainability in clinical decision support systems},
  author={Amann, Julia and Vetter, Dennis and Blomberg, Stig Nikolaj and Christensen, Helle Collatz and Coffee, Megan and Gerke, Sara and Gilbert, Thomas K and Hagendorff, Thilo and Holm, Sune and Livne, Michelle and others},
  journal={PLOS Digital Health},
  volume={1},
  number={2},
  pages={e0000016},
  year={2022},
  publisher={Public Library of Science}
}

@article{spooner2020comparison,
  title={A comparison of machine learning methods for survival analysis of high-dimensional clinical data for dementia prediction},
  author={Spooner, Annette and Chen, Emily and Sowmya, Arcot and Sachdev, Perminder and Kochan, Nicole A and Trollor, Julian and Brodaty, Henry},
  journal={Scientific reports},
  volume={10},
  number={1},
  pages={20410},
  year={2020},
  publisher={Nature Publishing Group UK London}
}

@article{garvin2009automated,
  title={Automated 3-D intraretinal layer segmentation of macular spectral-domain optical coherence tomography images},
  author={Garvin, Mona Kathryn and Abramoff, Michael David and Wu, Xiaodong and Russell, Stephen R and Burns, Trudy L and Sonka, Milan},
  journal={IEEE transactions on medical imaging},
  volume={28},
  number={9},
  pages={1436--1447},
  year={2009},
  publisher={IEEE}
}

@inproceedings{kwon2023efficient,
  title={Efficient Memory Management for Large Language Model Serving with PagedAttention},
  author={Woosuk Kwon and Zhuohan Li and Siyuan Zhuang and Ying Sheng and Lianmin Zheng and Cody Hao Yu and Joseph E. Gonzalez and Hao Zhang and Ion Stoica},
  booktitle={Proceedings of the ACM SIGOPS 29th Symposium on Operating Systems Principles},
  year={2023}
}

@misc{litellm_github,
  author       = {BerriAI},
  title        = {LiteLLM},
  year         = {2024},
  howpublished = {\url{https://github.com/BerriAI/litellm}},
}

@misc{ollama_github,
  author       = {Ollama},
  title        = {Ollama},
  year         = {2024},
  howpublished = {\url{https://github.com/ollama/ollama}},
}

@article{mcdonald2003loinc,
  title   = {LOINC, a universal standard for identifying laboratory observations: a 5-year update},
  author  = {McDonald, Clement J. and Huff, Stanley M. and Suico, John G. and et al.},
  journal = {Clinical Chemistry},
  year    = {2003},
  volume  = {49},
  number  = {4},
  pages   = {624--633}
}

@article{steindel2010international,
  title={International classification of diseases, clinical modification and procedure coding system: descriptive overview of the next generation HIPAA code sets},
  author={Steindel, Steven J},
  journal={Journal of the American Medical Informatics Association},
  volume={17},
  number={3},
  pages={274--282},
  year={2010},
  publisher={BMJ Group BMA House, Tavistock Square, London, WC1H 9JR}
}

@misc{who_atc,
  author       = {{World Health Organization Collaborating Centre for Drug Statistics Methodology}},
  title        = {ATC/DDD Index},
  year         = {2024},
  howpublished = {\url{https://www.whocc.no/atc_ddd_index/}},
}

@misc{fda_ndc,
  author       = {{U.S. Food and Drug Administration}},
  title        = {National Drug Code Directory},
  year         = {2024},
  howpublished = {\url{https://www.fda.gov/drugs/drug-approvals-and-databases/national-drug-code-directory}},
}

@misc{ama_cpt4,
  author       = {{American Medical Association}},
  title        = {Current Procedural Terminology (CPT) Professional Edition},
  year         = {2024},
  howpublished = {\url{https://www.ama-assn.org/practice-management/cpt}},
}

@book{ama_cpt_book,
  author    = {{American Medical Association}},
  title     = {CPT 2024 Professional Edition},
  publisher = {American Medical Association},
  year      = {2023},
  address   = {Chicago, IL}
}

@article{nelson2011rxnorm,
  title   = {Normalized names for clinical drugs: RxNorm at 6 years},
  author  = {Nelson, Stuart J. and Zeng, Kelly and Kilbourne, Jim and Powell, Tammy and Moore, Robert},
  journal = {Journal of the American Medical Informatics Association},
  year    = {2011},
  volume  = {18},
  number  = {4},
  pages   = {441--448}
}

@misc{cms_hcpcs,
  author       = {{Centers for Medicare \& Medicaid Services}},
  title        = {Healthcare Common Procedure Coding System (HCPCS)},
  year         = {2024},
  howpublished = {\url{https://www.cms.gov/medicare/coding-billing/healthcare-common-procedure-system}},
}

@misc{cdc_icd9cm,
  author       = {{Centers for Disease Control and Prevention}},
  title        = {International Classification of Diseases, Ninth Revision, Clinical Modification (ICD-9-CM)},
  year         = {2015},
  howpublished = {\url{https://archive.cdc.gov/www_cdc_gov/nchs/icd/icd9cm.htm}},
}

@book{who_icd9,
  author    = {{World Health Organization}},
  title     = {International Classification of Diseases, 9th Revision},
  publisher = {World Health Organization},
  year      = {1977},
  address   = {Geneva, Switzerland}
}

@article{openai_chatgpt4o,
  title={Gpt-4o system card},
  author={Hurst, Aaron and Lerer, Adam and Goucher, Adam P and Perelman, Adam and Ramesh, Aditya and Clark, Aidan and Ostrow, AJ and Welihinda, Akila and Hayes, Alan and Radford, Alec and others},
  journal={arXiv preprint arXiv:2410.21276},
  year={2024}
}

@article{openai_gpt4,
  title   = {GPT-4 Technical Report},
  author  = {{OpenAI}},
  journal = {arXiv preprint arXiv:2303.08774},
  year    = {2023}
}

@article{hong2025next,
  title={Next-generation database interfaces: A survey of llm-based text-to-sql},
  author={Hong, Zijin and Yuan, Zheng and Zhang, Qinggang and Chen, Hao and Dong, Junnan and Huang, Feiran and Huang, Xiao},
  journal={IEEE Transactions on Knowledge and Data Engineering},
  year={2025},
  publisher={IEEE}
}

@article{BILLOT2023102789,
title = {SynthSeg: Segmentation of brain MRI scans of any contrast and resolution without retraining},
journal = {Medical Image Analysis},
volume = {86},
pages = {102789},
year = {2023},
issn = {1361-8415},
doi = {https://doi.org/10.1016/j.media.2023.102789},
author = {Benjamin Billot and Douglas N. Greve and Oula Puonti and Axel Thielscher and Koen {Van Leemput} and Bruce Fischl and Adrian V. Dalca and Juan Eugenio Iglesias}
}

@article{Li2016TheFS,
  title={The first step for neuroimaging data analysis: DICOM to NIfTI conversion},
  author={Xiangrui Li and Paul Simon Morgan and John Ashburner and Jolinda C. Smith and Chris Rorden},
  journal={Journal of Neuroscience Methods},
  year={2016},
  volume={264},
  pages={47-56},
}

@book{talairach1988co,
  title={Co-planar Stereotaxic Atlas of the Human Brain: 3-dimensional Proportional System : an Approach to Cerebral Imaging},
  author={Talairach, J. and Tournoux, P.},
  isbn={9783137117018},
  lccn={lc88029462},
  series={Thieme Publishers Series},
  year={1988},
  publisher={G. Thieme}
}

@article{hoopes2022synthstrip,
    title={{SynthStrip}: skull-stripping for any brain image},
    author={Hoopes, Andrew and Mora, Jocelyn S and Dalca, Adrian V and Fischl, Bruce and Hoffmann, Malte},
    journal={NeuroImage},
    volume={260},
    pages={119474},
    year={2022},
    publisher={Elsevier}
}

@article{mni_152_template,
author = {Fonov, Vladimir and Evans, Alan and Botteron, Kelly and Almli, C and Mckinstry, Robert and Collins, Louis},
year = {2011},
month = {01},
pages = {313-27},
title = {Unbiased Average Age-Appropriate Atlases for Pediatric Studies},
volume = {54},
journal = {NeuroImage},
doi = {10.1016/j.neuroimage.2010.07.033}
}

@article{Fonov2009UnbiasedNA,
  title={Unbiased nonlinear average age-appropriate brain templates from birth to adulthood},
  author={V. Fonov and A. Evans and RC McKinstry and C. Robert Almli and DL Collins},
  journal={NeuroImage},
  year={2009},
  volume={47},
}

@misc{obi_deid_roberta_i2b2,
  title        = {deid\_roberta\_i2b2: RoBERTa model fine-tuned for medical note de-identification},
  author       = {{OBI}},
  year         = {2023},
  howpublished = {\url{https://huggingface.co/obi/deid_roberta_i2b2}},
}

@article{liu2019roberta,
  title={Roberta: A robustly optimized bert pretraining approach},
  author={Liu, Yinhan and Ott, Myle and Goyal, Naman and Du, Jingfei and Joshi, Mandar and Chen, Danqi and Levy, Omer and Lewis, Mike and Zettlemoyer, Luke and Stoyanov, Veselin},
  journal={arXiv preprint arXiv:1907.11692},
  year={2019}
}

@article{dua2025clinically,
  title={Clinically Grounded Agent-based Report Evaluation: An Interpretable Metric for Radiology Report Generation},
  author={Dua, Radhika and Joon, Young and Dogra, Siddhant and Freedman, Daniel and Ruan, Diana and Nashawaty, Motaz and Rigau, Danielle and Alber, Daniel Alexander and Zhang, Kang and Cho, Kyunghyun and others},
  journal={arXiv preprint arXiv:2508.02808},
  year={2025}
}

@article{suzgun2025dynamic,
  title={Dynamic cheatsheet: Test-time learning with adaptive memory},
  author={Suzgun, Mirac and Yuksekgonul, Mert and Bianchi, Federico and Jurafsky, Dan and Zou, James},
  journal={arXiv preprint arXiv:2504.07952},
  year={2025}
}

@article{yuksekgonul2024textgrad,
  title={Textgrad: Automatic" differentiation" via text},
  author={Yuksekgonul, Mert and Bianchi, Federico and Boen, Joseph and Liu, Sheng and Huang, Zhi and Guestrin, Carlos and Zou, James},
  journal={arXiv preprint arXiv:2406.07496},
  year={2024}
}

@article{Jensen2012MiningEH,
  title={Mining electronic health records: towards better research applications and clinical care},
  author={Peter Bj{\o}dstrup Jensen and Lars Juhl Jensen and S{\o}ren Brunak},
  journal={Nature Reviews Genetics},
  year={2012},
  volume={13},
  pages={395-405},
}

@article{Bates2014BigDI,
  title={Big data in health care: using analytics to identify and manage high-risk and high-cost patients.},
  author={D. Bates and Suchi Saria and Lucila Ohno-Machado and Anand Shah and Gabriel J. Escobar},
  journal={Health affairs},
  year={2014},
  volume={33 7},
  pages={
          1123-31
        },
}

@article{Miotto2016DeepPA,
  title={Deep Patient: An Unsupervised Representation to Predict the Future of Patients from the Electronic Health Records},
  author={Riccardo Miotto and Li Li and Brian A. Kidd and Joel T. Dudley},
  journal={Scientific Reports},
  year={2016},
  volume={6},
}

@article{Esteva2019AGT,
  title={A guide to deep learning in healthcare},
  author={Andre Esteva and Alexandre Robicquet and Bharath Ramsundar and Volodymyr Kuleshov and Mark A. DePristo and Katherine Chou and Claire Cui and Greg S. Corrado and Sebastian Thrun and Jeff Dean},
  journal={Nature Medicine},
  year={2019},
  volume={25},
  pages={24 - 29},
}

@article{Wang2018ClinicalIE,
  title={Clinical information extraction applications: A literature review},
  author={Yanshan Wang and Liwei Wang and Majid Rastegar-Mojarad and Sungrim Moon and Feichen Shen and Naveed Afzal and Sijia Liu and Yuqun Zeng and Saeed Mehrabi and Sunghwan Sohn and Hongfang Liu},
  journal={Journal of biomedical informatics},
  year={2018},
  volume={77},
  pages={
          34-49
        },
}

@article{rajkomar2018scalable,
  title={Scalable and accurate deep learning with electronic health records},
  author={Rajkomar, Alvin and Oren, Eyal and Chen, Kai and Dai, Andrew M and Hajaj, Nissan and Hardt, Michaela and Liu, Peter J and Liu, Xiaobing and Marcus, Jake and Sun, Mimi and others},
  journal={NPJ digital medicine},
  volume={1},
  number={1},
  pages={18},
  year={2018},
  publisher={Nature Publishing Group UK London}
}

@article{Johnson2016MIMICIIIAF,
  title={MIMIC-III, a freely accessible critical care database},
  author={Alistair E. W. Johnson and Tom J. Pollard and Lu Shen and Li-wei H. Lehman and Mengling Feng and Mohammad Mahdi Ghassemi and Benjamin Moody and Peter Szolovits and Leo Anthony Celi and Roger G. Mark},
  journal={Scientific Data},
  year={2016},
  volume={3},
}

@article{Johnson2023MIMICIVAF,
  title={MIMIC-IV, a freely accessible electronic health record dataset},
  author={Alistair E. W. Johnson and Lucas Bulgarelli and Lu Shen and Alvin Gayles and Ayad Shammout and Steven Horng and Tom J. Pollard and Benjamin Moody and Brian Gow and Li-wei H. Lehman and Leo Anthony Celi and Roger G. Mark},
  journal={Scientific Data},
  year={2023},
  volume={10},
}

@article{Weiskopf2013MethodsAD,
  title={Methods and dimensions of electronic health record data quality assessment: enabling reuse for clinical research},
  author={Nicole Gray Weiskopf and Chunhua Weng},
  journal={Journal of the American Medical Informatics Association : JAMIA},
  year={2013},
  volume={20},
  pages={144 - 151},
}

@article{dubois2016preclinical,
  title={Preclinical Alzheimer's disease: definition, natural history, and diagnostic criteria},
  author={Dubois, Bruno and Hampel, Harald and Feldman, Howard H and Scheltens, Philip and Aisen, Paul and Andrieu, Sandrine and Bakardjian, Hovagim and Benali, Habib and Bertram, Lars and Blennow, Kaj and others},
  journal={Alzheimer's \& Dementia},
  volume={12},
  number={3},
  pages={292--323},
  year={2016},
  publisher={Elsevier}
}

@article{Reid2006SubjectiveMC,
  title={Subjective Memory Complaints and Cognitive Impairment in Older People},
  author={Louise M. Reid and Alasdair M. J. MacLullich},
  journal={Dementia and Geriatric Cognitive Disorders},
  year={2006},
  volume={22},
  pages={471 - 485},
}

@article{garand2009diagnostic,
  title={Diagnostic labels, stigma, and participation in research related to dementia and mild cognitive impairment},
  author={Garand, Linda and Lingler, Jennifer H and Conner, Kyaien O and Dew, Mary Amanda},
  journal={Research in gerontological nursing},
  volume={2},
  number={2},
  pages={112--121},
  year={2009},
  publisher={SLACK Incorporated Thorofare, NJ}
}

@article{Petersen2018PracticeGU,
  title={Practice guideline update summary: Mild cognitive impairment},
  author={Ronald C. Petersen and Oscar Lopez and Melissa J. Armstrong and Thomas S. D. Getchius and Mary Ganguli and David S. Gloss and Gary S. Gronseth and Daniel Marson and Tamara M Pringsheim and Gregory S. Day and Mark A. Sager and James C. Stevens and Alexander D. Rae-Grant},
  journal={Neurology},
  year={2018},
  volume={90},
  pages={126 - 135},
}

@article{mitchell2014risk,
  title={Risk of dementia and mild cognitive impairment in older people with subjective memory complaints: meta-analysis},
  author={Mitchell, Alex J and Beaumont, Helen and Ferguson, David and Yadegarfar, Motahare and Stubbs, Brendon},
  journal={Acta Psychiatrica Scandinavica},
  volume={130},
  number={6},
  pages={439--451},
  year={2014},
  publisher={Wiley Online Library}
}

@article{schneider2007mixed,
  title={Mixed brain pathologies account for most dementia cases in community-dwelling older persons},
  author={Schneider, Julie A and Arvanitakis, Zoe and Bang, Woojeong and Bennett, David A},
  journal={Neurology},
  volume={69},
  number={24},
  pages={2197--2204},
  year={2007},
  publisher={Lippincott Williams \& Wilkins}
}

@article{jack2018nia,
  title={NIA-AA research framework: toward a biological definition of Alzheimer's disease},
  author={Jack Jr, Clifford R and Bennett, David A and Blennow, Kaj and Carrillo, Maria C and Dunn, Billy and Haeberlein, Samantha Budd and Holtzman, David M and Jagust, William and Jessen, Frank and Karlawish, Jason and others},
  journal={Alzheimer's \& dementia},
  volume={14},
  number={4},
  pages={535--562},
  year={2018},
}

@article{Stone2010WhoIR,
  title={Who is referred to neurology clinics?—The diagnoses made in 3781 new patients},
  author={Jon Stone and Alan J. Carson and Roderick Duncan and Richard C. Roberts and Charles P. Warlow and Carina Hibberd and Richard J Coleman and Roger E. Cull and Gordon D. Murray and Anthony J. Pelosi and Jonathan Cavanagh and Keith Matthews and Rainer Goldbeck and Roger Smyth and Jane Walker and Michael Sharpe},
  journal={Clinical Neurology and Neurosurgery},
  year={2010},
  volume={112},
  pages={747-751},
}

@article{Bradford2009MissedAD,
  title={Missed and Delayed Diagnosis of Dementia in Primary Care: Prevalence and Contributing Factors},
  author={Andrea Bradford and Mark E. Kunik and Paul E. Schulz and Susan P. Williams and Hardeep Singh},
  journal={Alzheimer Disease \& Associated Disorders},
  year={2009},
  volume={23},
  pages={306-314},
}

@article{Wippold2015ACRAC,
  title={ACR Appropriateness Criteria Dementia and Movement Disorders.},
  author={Franz J. Wippold and Douglas C. Brown and Daniel F. Broderick and Judah Burns and A S Corey and Tejaswini K Deshmukh and Annette C Douglas and Kathryn L. Holloway and Bharathi Dasan Jagadeesan and Jennifer S Jurgens and T A Kennedy and Nandini D. Patel and Joel S. Perlmutter and Joshua M. Rosenow and Konstantin Slavin and Ratham M Subramaniam},
  journal={Journal of the American College of Radiology : JACR},
  year={2015},
  volume={12 1},
  pages={
          19-28
        },
}

@article{Soderlund2025ACRAC,
  title={ACR Appropriateness Criteria{\textregistered} Dementia: 2024 Update.},
  author={Karl A. Soderlund and Matthew J Austin and Sharona Ben-Haim and Sammy Chu and Jana Ivanidze and Pallavi Joshi and Aleks Kalnins and Maura Kennedy and Ambar Kulshreshtha and Phillip H. Kuo and Joseph C. Masdeu and Tejas Nikumbh and Bruno P. Soares and Ashesh Thaker and Lily L Wang and Sevil Yasar and Robert Shih},
  journal={Journal of the American College of Radiology : JACR},
  year={2025},
  volume={22 5S},
  pages={
          S202-S233
        },
}

@article{Johnson2013AppropriateUC,
  title={Appropriate Use Criteria for Amyloid PET: A Report of the Amyloid Imaging Task Force, the Society of Nuclear Medicine and Molecular Imaging, and the Alzheimer’s Association},
  author={Keith A. Johnson and Satoshi Minoshima and Nicolaas I. Bohnen and Kevin J. Donohoe and Norman L. Foster and Peter Herscovitch and Jason Karlawish and Christopher C. Rowe and Maria C. Carrillo and Dean Mitchell Hartley and Saima Hedrick and Virginia Pappas and William H. Thies},
  journal={The Journal of Nuclear Medicine},
  year={2013},
  volume={54},
  pages={476 - 490},
}

@article{rabinovici2025updated,
  title={Updated appropriate use criteria for amyloid and tau PET: a report from the Alzheimer’s Association and Society for Nuclear Medicine and Molecular Imaging Workgroup},
  author={Rabinovici, Gil D and Knopman, David S and Arbizu, Javier and Benzinger, Tammie LS and Donohoe, Kevin J and Hansson, Oskar and Herscovitch, Peter and Kuo, Phillip H and Lingler, Jennifer H and Minoshima, Satoshi and others},
  journal={Journal of Nuclear Medicine},
  volume={66},
  number={Supplement 2},
  pages={S5--S31},
  year={2025},
  publisher={Society of Nuclear Medicine}
}

@article{xue2024ai,
  title={AI-based differential diagnosis of dementia etiologies on multimodal data},
  author={Xue, Chonghua and Kowshik, Sahana S and Lteif, Diala and Puducheri, Shreyas and Jasodanand, Varuna H and Zhou, Olivia T and Walia, Anika S and Guney, Osman B and Zhang, J Diana and Po{\'e}sy, Serena and others},
  journal={Nature Medicine},
  volume={30},
  number={10},
  pages={2977--2989},
  year={2024},
  publisher={Nature Publishing Group US New York}
}

@article{wen2020convolutional,
  title={Convolutional neural networks for classification of Alzheimer's disease: Overview and reproducible evaluation},
  author={Wen, Junhao and Thibeau-Sutre, Elina and Diaz-Melo, Mauricio and Samper-Gonz{\'a}lez, Jorge and Routier, Alexandre and Bottani, Simona and Dormont, Didier and Durrleman, Stanley and Burgos, Ninon and Colliot, Olivier and others},
  journal={Medical image analysis},
  volume={63},
  pages={101694},
  year={2020},
  publisher={Elsevier}
}

@article{huang2019diagnosis,
  title={Diagnosis of Alzheimer’s disease via multi-modality 3D convolutional neural network},
  author={Huang, Yechong and Xu, Jiahang and Zhou, Yuncheng and Tong, Tong and Zhuang, Xiahai and Alzheimer’s Disease Neuroimaging Initiative (ADNI)},
  journal={Frontiers in neuroscience},
  volume={13},
  pages={509},
  year={2019},
  publisher={Frontiers Media SA}
}

@article{ding2019deep,
  title={A deep learning model to predict a diagnosis of Alzheimer disease by using 18F-FDG PET of the brain},
  author={Ding, Yiming and Sohn, Jae Ho and Kawczynski, Michael G and Trivedi, Hari and Harnish, Roy and Jenkins, Nathaniel W and Lituiev, Dmytro and Copeland, Timothy P and Aboian, Mariam S and Mari Aparici, Carina and others},
  journal={Radiology},
  volume={290},
  number={2},
  pages={456--464},
  year={2019},
  publisher={Radiological Society of North America}
}

@article{huang2020multimodal,
  title={Multimodal fusion with deep neural networks for leveraging CT imaging and electronic health record: a case-study in pulmonary embolism detection},
  author={Huang, Shih-Cheng and Pareek, Anuj and Zamanian, Roham and Banerjee, Imon and Lungren, Matthew P},
  journal={Scientific reports},
  volume={10},
  number={1},
  pages={22147},
  year={2020},
  publisher={Nature Publishing Group UK London}
}

@article{moor2023foundation,
  title={Foundation models for generalist medical artificial intelligence},
  author={Moor, Michael and Banerjee, Oishi and Abad, Zahra Shakeri Hossein and Krumholz, Harlan M and Leskovec, Jure and Topol, Eric J and Rajpurkar, Pranav},
  journal={Nature},
  volume={616},
  number={7956},
  pages={259--265},
  year={2023},
  publisher={Nature Publishing Group UK London}
}

@article{ghassemi2021false,
  title={The false hope of current approaches to explainable artificial intelligence in health care},
  author={Ghassemi, Marzyeh and Oakden-Rayner, Luke and Beam, Andrew L},
  journal={The lancet digital health},
  volume={3},
  number={11},
  pages={e745--e750},
  year={2021},
  publisher={Elsevier}
}

@inproceedings{tonekaboni2019clinicians,
  title={What clinicians want: contextualizing explainable machine learning for clinical end use},
  author={Tonekaboni, Sana and Joshi, Shalmali and McCradden, Melissa D and Goldenberg, Anna},
  booktitle={Machine learning for healthcare conference},
  pages={359--380},
  year={2019},
  organization={PMLR}
}

@article{kelly2019key,
  title={Key challenges for delivering clinical impact with artificial intelligence},
  author={Kelly, Christopher J and Karthikesalingam, Alan and Suleyman, Mustafa and Corrado, Greg and King, Dominic},
  journal={BMC medicine},
  volume={17},
  number={1},
  pages={195},
  year={2019},
  publisher={Springer}
}

@article{rudin2019stop,
  title={Stop explaining black box machine learning models for high stakes decisions and use interpretable models instead},
  author={Rudin, Cynthia},
  journal={Nature machine intelligence},
  volume={1},
  number={5},
  pages={206--215},
  year={2019},
  publisher={Nature Publishing Group UK London}
}

@article{amann2020explainability,
  title={Explainability for artificial intelligence in healthcare: a multidisciplinary perspective},
  author={Amann, Julia and Blasimme, Alessandro and Vayena, Effy and Frey, Dietmar and Madai, Vince I and Precise4Q Consortium},
  journal={BMC medical informatics and decision making},
  volume={20},
  number={1},
  pages={310},
  year={2020},
  publisher={Springer}
}

@inproceedings{yao2022react,
  title={React: Synergizing reasoning and acting in language models},
  author={Yao, Shunyu and Zhao, Jeffrey and Yu, Dian and Du, Nan and Shafran, Izhak and Narasimhan, Karthik R and Cao, Yuan},
  booktitle={The eleventh international conference on learning representations},
  year={2022}
}

@inproceedings{xu2023gentopia,
  title={Gentopia. ai: A collaborative platform for tool-augmented llms},
  author={Xu, Binfeng and Liu, Xukun and Shen, Hua and Han, Zeyu and Li, Yuhan and Yue, Murong and Peng, Zhiyuan and Liu, Yuchen and Yao, Ziyu and Xu, Dongkuan},
  booktitle={Proceedings of the 2023 Conference on Empirical Methods in Natural Language Processing: System Demonstrations},
  pages={237--245},
  year={2023}
}

@article{xu2023rewoo,
  title={Rewoo: Decoupling reasoning from observations for efficient augmented language models},
  author={Xu, Binfeng and Peng, Zhiyuan and Lei, Bowen and Mukherjee, Subhabrata and Liu, Yuchen and Xu, Dongkuan},
  journal={arXiv preprint arXiv:2305.18323},
  year={2023}
}

@article{liang2024taskmatrix,
  title={Taskmatrix. ai: Completing tasks by connecting foundation models with millions of apis},
  author={Liang, Yaobo and Wu, Chenfei and Song, Ting and Wu, Wenshan and Xia, Yan and Liu, Yu and Ou, Yang and Lu, Shuai and Ji, Lei and Mao, Shaoguang and others},
  journal={Intelligent Computing},
  volume={3},
  pages={0063},
  year={2024},
  publisher={AAAS}
}

@article{lu2024multimodal,
  title={A multimodal generative AI copilot for human pathology},
  author={Lu, Ming Y and Chen, Bowen and Williamson, Drew FK and Chen, Richard J and Zhao, Melissa and Chow, Aaron K and Ikemura, Kenji and Kim, Ahrong and Pouli, Dimitra and Patel, Ankush and others},
  journal={Nature},
  volume={634},
  number={8033},
  pages={466--473},
  year={2024},
  publisher={Nature Publishing Group UK London}
}

@article{tu2024towards,
  title={Towards generalist biomedical AI},
  author={Tu, Tao and Azizi, Shekoofeh and Driess, Danny and Schaekermann, Mike and Amin, Mohamed and Chang, Pi-Chuan and Carroll, Andrew and Lau, Charles and Tanno, Ryutaro and Ktena, Ira and others},
  journal={Nejm Ai},
  volume={1},
  number={3},
  pages={AIoa2300138},
  year={2024},
  publisher={Massachusetts Medical Society}
}

@article{zhao2025agentic,
  title={An agentic system for rare disease diagnosis with traceable reasoning},
  author={Zhao, Weike and Wu, Chaoyi and Fan, Yanjie and Zhang, Xiaoman and Qiu, Pengcheng and Sun, Yuze and Zhou, Xiao and Wang, Yanfeng and Sun, Xin and Zhang, Ya and others},
  journal={arXiv preprint arXiv:2506.20430},
  year={2025}
}

@inproceedings{fleming2024medalign,
  title={Medalign: A clinician-generated dataset for instruction following with electronic medical records},
  author={Fleming, Scott L and Lozano, Alejandro and Haberkorn, William J and Jindal, Jenelle A and Reis, Eduardo and Thapa, Rahul and Blankemeier, Louis and Genkins, Julian Z and Steinberg, Ethan and Nayak, Ashwin and others},
  booktitle={Proceedings of the AAAI Conference on Artificial Intelligence},
  volume={38},
  number={20},
  pages={22021--22030},
  year={2024}
}

@inproceedings{chen2016xgboost,
  title={Xgboost: A scalable tree boosting system},
  author={Chen, Tianqi and Guestrin, Carlos},
  booktitle={Proceedings of the 22nd acm sigkdd international conference on knowledge discovery and data mining},
  pages={785--794},
  year={2016}
}

@book{hosmer2013applied,
  title={Applied logistic regression},
  author={Hosmer Jr, David W and Lemeshow, Stanley and Sturdivant, Rodney X},
  year={2013},
  publisher={John Wiley \& Sons}
}

@article{breiman2001random,
  title={Random forests},
  author={Breiman, Leo},
  journal={Machine learning},
  volume={45},
  number={1},
  pages={5--32},
  year={2001},
  publisher={Springer}
}

@article{kvamme2019time,
  title={Time-to-event prediction with neural networks and Cox regression},
  author={Kvamme, H{\aa}vard and Borgan, {\O}rnulf and Scheel, Ida},
  journal={Journal of machine learning research},
  volume={20},
  number={129},
  pages={1--30},
  year={2019}
}

@article{katzman2018deepsurv,
  title={DeepSurv: personalized treatment recommender system using a Cox proportional hazards deep neural network},
  author={Katzman, Jared L and Shaham, Uri and Cloninger, Alexander and Bates, Jonathan and Jiang, Tingting and Kluger, Yuval},
  journal={BMC medical research methodology},
  volume={18},
  number={1},
  pages={24},
  year={2018},
  publisher={Springer}
}

@article{paszke2019pytorch,
  title={Pytorch: An imperative style, high-performance deep learning library},
  author={Paszke, Adam and Gross, Sam and Massa, Francisco and Lerer, Adam and Bradbury, James and Chanan, Gregory and Killeen, Trevor and Lin, Zeming and Gimelshein, Natalia and Antiga, Luca and others},
  journal={Advances in neural information processing systems},
  volume={32},
  year={2019}
}

@inproceedings{wolf2020transformers,
  title={Transformers: State-of-the-art natural language processing},
  author={Wolf, Thomas and Debut, Lysandre and Sanh, Victor and Chaumond, Julien and Delangue, Clement and Moi, Anthony and Cistac, Pierric and Rault, Tim and Louf, R{\'e}mi and Funtowicz, Morgan and others},
  booktitle={Proceedings of the 2020 conference on empirical methods in natural language processing: system demonstrations},
  pages={38--45},
  year={2020}
}

@article{paulose2021national,
  title={The National Health and Nutrition Examination Survey (NHANES), 2021--2022: adapting data collection in a COVID-19 environment},
  author={Paulose-Ram, Ryne and Graber, Jessica E and Woodwell, David and Ahluwalia, Namanjeet},
  journal={American journal of public health},
  volume={111},
  number={12},
  pages={2149--2156},
  year={2021},
  publisher={American Public Health Association}
}

@article{national2017health,
  title={Health, United States, 2016, with chartbook on long-term trends in health},
  author={National Center for Health Statistics},
  year={2017},
  publisher={US Deptartment of Health and Human Services}
}

@article{sonnega2014cohort,
  title={Cohort profile: the health and retirement study (HRS)},
  author={Sonnega, Amanda and Faul, Jessica D and Ofstedal, Mary Beth and Langa, Kenneth M and Phillips, John WR and Weir, David R},
  journal={International journal of epidemiology},
  volume={43},
  number={2},
  pages={576--585},
  year={2014},
  publisher={Oxford University Press}
}

@article{johnson2012brain,
  title={Brain imaging in Alzheimer disease},
  author={Johnson, Keith A and Fox, Nick C and Sperling, Reisa A and Klunk, William E},
  journal={Cold Spring Harbor perspectives in medicine},
  volume={2},
  number={4},
  pages={a006213},
  year={2012},
  publisher={Cold Spring Harbor Laboratory Press}
}

@article{gorno2011classification,
  title={Classification of primary progressive aphasia and its variants},
  author={Gorno-Tempini, Maria Luisa and Hillis, Argye E and Weintraub, Sandra and Kertesz, Andrew and Mendez, Mario and Cappa, Stefano F and Ogar, Jennifer M and Rohrer, Jonathan D and Black, Steven and Boeve, Bradley F and others},
  journal={Neurology},
  volume={76},
  number={11},
  pages={1006--1014},
  year={2011},
  publisher={Lippincott Williams \& Wilkins Hagerstown, MD}
}

@article{chui2007subcortical,
  title={Subcortical ischemic vascular dementia},
  author={Chui, Helena C},
  journal={Neurologic clinics},
  volume={25},
  number={3},
  pages={717--740},
  year={2007},
  publisher={Elsevier}
}

@article{livingston2024dementia,
  title={Dementia prevention, intervention, and care: 2024 report of the Lancet standing Commission},
  author={Livingston, Gill and Huntley, Jonathan and Liu, Kathy Y and Costafreda, Sergi G and Selb{\ae}k, Geir and Alladi, Suvarna and Ames, David and Banerjee, Sube and Burns, Alistair and Brayne, Carol and others},
  journal={The lancet},
  volume={404},
  number={10452},
  pages={572--628},
  year={2024},
  publisher={Elsevier}
}

@article{gorelick2016vascular,
  title={Vascular cognitive impairment and dementia},
  author={Gorelick, Philip B and Counts, Scott E and Nyenhuis, David},
  journal={Biochimica et Biophysica Acta (BBA)-Molecular Basis of Disease},
  volume={1862},
  number={5},
  pages={860--868},
  year={2016},
  publisher={Elsevier}
}

@article{wong2022vascular,
  title={Vascular cognitive impairment and dementia},
  author={Wong, Ellen Chang and Chui, Helena Chang},
  journal={Continuum: lifelong learning in neurology},
  volume={28},
  number={3},
  pages={750--780},
  year={2022},
  publisher={LWW}
}

@article{alber2019white,
  title={White matter hyperintensities in vascular contributions to cognitive impairment and dementia (VCID): knowledge gaps and opportunities},
  author={Alber, Jessica and Alladi, Suvarna and Bae, Hee-Joon and Barton, David A and Beckett, Laurel A and Bell, Joanne M and Berman, Sara E and Biessels, Geert Jan and Black, Sandra E and Bos, Isabelle and others},
  journal={Alzheimer's \& Dementia: Translational Research \& Clinical Interventions},
  volume={5},
  pages={107--117},
  year={2019},
  publisher={Elsevier}
}

@article{jahn2013memory,
  title={Memory loss in Alzheimer's disease},
  author={Jahn, Holger},
  journal={Dialogues in clinical neuroscience},
  volume={15},
  number={4},
  pages={445--454},
  year={2013},
  publisher={Taylor \& Francis}
}

@phdthesis{prince2015world,
  title={World Alzheimer report 2015. The global impact of dementia: an analysis of prevalence, incidence, cost and trends.},
  author={Prince, Martin and Wimo, Anders and Guerchet, Ma{\"e}lenn and Ali, Gemma-Claire and Wu, Yu-Tzu and Prina, Matthew},
  year={2015},
  school={Alzheimer's Disease International}
}

@article{manly2022estimating,
  title={Estimating the prevalence of dementia and mild cognitive impairment in the US: the 2016 health and retirement study harmonized cognitive assessment protocol project},
  author={Manly, Jennifer J and Jones, Richard N and Langa, Kenneth M and Ryan, Lindsay H and Levine, Deborah A and McCammon, Ryan and Heeringa, Steven G and Weir, David},
  journal={JAMA neurology},
  volume={79},
  number={12},
  pages={1242--1249},
  year={2022}
}

@article{kramarow2024diagnosed,
  title={Diagnosed dementia in adults age 65 and older: United States, 2022},
  author={Kramarow, Ellen A},
  year={2024}
}

@article{spargo2023estimating,
  title={Estimating prevalence of early symptomatic Alzheimer's disease in the United States},
  author={Spargo, Drew and Zur, Richard and Lin, Pei-Jung and Synnott, Patricia and Klein, Eric and Hartry, Ann},
  journal={Alzheimer's \& Dementia: Diagnosis, Assessment \& Disease Monitoring},
  volume={15},
  number={4},
  pages={e12497},
  year={2023},
  publisher={Wiley Online Library}
}

@article{popuri2020using,
  title={Using machine learning to quantify structural MRI neurodegeneration patterns of Alzheimer's disease into dementia score: Independent validation on 8,834 images from ADNI, AIBL, OASIS, and MIRIAD databases},
  author={Popuri, Karteek and Ma, Da and Wang, Lei and Beg, Mirza Faisal},
  journal={Human Brain Mapping},
  volume={41},
  number={14},
  pages={4127--4147},
  year={2020},
  publisher={Wiley Online Library}
}

@article{hwang2022disentangling,
  title={Disentangling Alzheimer’s disease neurodegeneration from typical brain ageing using machine learning},
  author={Hwang, Gyujoon and Abdulkadir, Ahmed and Erus, Guray and Habes, Mohamad and Pomponio, Raymond and Shou, Haochang and Doshi, Jimit and Mamourian, Elizabeth and Rashid, Tanweer and Bilgel, Murat and others},
  journal={Brain Communications},
  volume={4},
  number={3},
  pages={fcac117},
  year={2022},
  publisher={Oxford University Press}
}

@article{wang2025retinal,
  title={Retinal biomarkers for the risk of Alzheimer’s disease and frontotemporal dementia},
  author={Wang, Ruihan and Cai, Jiajie and Gao, Yuzhu and Tang, Yingying and Gao, Hui and Qin, Linyuan and Cai, Hanlin and Yang, Feng and Ren, Yimeng and Luo, Caimei and others},
  journal={Frontiers in Aging Neuroscience},
  volume={16},
  pages={1513302},
  year={2025},
  publisher={Frontiers}
}

@article{ibrahim2023systematic,
  title={A systematic review on retinal biomarkers to diagnose dementia from OCT/OCTA images},
  author={Ibrahim, Yehia and Xie, Jianyang and Macerollo, Antonella and Sardone, Rodolfo and Shen, Yaochun and Romano, Vito and Zheng, Yalin},
  journal={Journal of Alzheimer's disease reports},
  volume={7},
  number={1},
  pages={1201--1235},
  year={2023},
  publisher={SAGE Publications Sage UK: London, England}
}

@article{perfalk2024predicting,
  title={Predicting involuntary admission following inpatient psychiatric treatment using machine learning trained on electronic health record data},
  author={Perfalk, Erik and Damgaard, Jakob Gr{\o}hn and Bernstorff, Martin and Hansen, Lasse and Danielsen, Andreas Aalkj{\ae}r and {\O}stergaard, S{\o}ren Dinesen},
  journal={Psychological Medicine},
  volume={54},
  number={15},
  pages={4348--4361},
  year={2024},
  publisher={Cambridge University Press}
}

@article{mao2023ad,
  title={AD-BERT: Using pre-trained language model to predict the progression from mild cognitive impairment to Alzheimer's disease},
  author={Mao, Chengsheng and Xu, Jie and Rasmussen, Luke and Li, Yikuan and Adekkanattu, Prakash and Pacheco, Jennifer and Bonakdarpour, Borna and Vassar, Robert and Shen, Li and Jiang, Guoqian and others},
  journal={Journal of Biomedical Informatics},
  volume={144},
  pages={104442},
  year={2023},
  publisher={Elsevier}
}

@article{li2023early,
  title={Early prediction of Alzheimer's disease and related dementias using real-world electronic health records},
  author={Li, Qian and Yang, Xi and Xu, Jie and Guo, Yi and He, Xing and Hu, Hui and Lyu, Tianchen and Marra, David and Miller, Amber and Smith, Glenn and others},
  journal={Alzheimer's \& Dementia},
  volume={19},
  number={8},
  pages={3506--3518},
  year={2023},
  publisher={Wiley Online Library}
}

@article{tang2024leveraging,
  title={Leveraging electronic health records and knowledge networks for Alzheimer’s disease prediction and sex-specific biological insights},
  author={Tang, Alice S and Rankin, Katherine P and Cerono, Gabriel and Miramontes, Silvia and Mills, Hunter and Roger, Jacquelyn and Zeng, Billy and Nelson, Charlotte and Soman, Karthik and Woldemariam, Sarah and others},
  journal={Nature Aging},
  volume={4},
  number={3},
  pages={379--395},
  year={2024},
  publisher={Nature Publishing Group US New York}
}

@article{openai_gpt5_system_card_2025,
  title={Openai gpt-5 system card},
  author={Singh, Aaditya and Fry, Adam and Perelman, Adam and Tart, Adam and Ganesh, Adi and El-Kishky, Ahmed and McLaughlin, Aidan and Low, Aiden and Ostrow, AJ and Ananthram, Akhila and others},
  journal={arXiv preprint arXiv:2601.03267},
  year={2025}
}

@article{sellergren2025medgemma,
  author       = {Sellergren, Andrew and others},
  title        = {MedGemma Technical Report},
  journal      = {arXiv preprint arXiv:2507.05201},
  year         = {2025}
}

@article{agarwal2025gpt,
  title={gpt-oss-120b \& gpt-oss-20b model card},
  author={Agarwal, Sandhini and Ahmad, Lama and Ai, Jason and Altman, Sam and Applebaum, Andy and Arbus, Edwin and Arora, Rahul K and Bai, Yu and Baker, Bowen and Bao, Haiming and others},
  journal={arXiv preprint arXiv:2508.10925},
  year={2025}
}

@article{tondelli2012structural,
  title={Structural MRI changes detectable up to ten years before clinical Alzheimer's disease},
  author={Tondelli, Manuela and Wilcock, Gordon K and Nichelli, Paolo and De Jager, Celeste A and Jenkinson, Mark and Zamboni, Giovanna},
  journal={Neurobiology of aging},
  volume={33},
  number={4},
  pages={825--e25},
  year={2012},
  publisher={Elsevier}
}

@article{debette2010clinical,
  title={The clinical importance of white matter hyperintensities on brain magnetic resonance imaging: systematic review and meta-analysis},
  author={Debette, St{\'e}phanie and Markus, HS20660506},
  journal={Bmj},
  volume={341},
  year={2010},
  publisher={British Medical Journal Publishing Group}
}

@article{lambert2018identifying,
  title={Identifying preclinical vascular dementia in symptomatic small vessel disease using MRI},
  author={Lambert, Christian and Zeestraten, Eva and Williams, Owen and Benjamin, Philip and Lawrence, Andrew J and Morris, Robin G and Mackinnon, Andrew D and Barrick, Thomas R and Markus, Hugh S},
  journal={NeuroImage: Clinical},
  volume={19},
  pages={925--938},
  year={2018},
  publisher={Elsevier}
}

@article{wilson2011cognitive,
  title={Cognitive decline in prodromal Alzheimer disease and mild cognitive impairment},
  author={Wilson, Robert S and Leurgans, Sue E and Boyle, Patricia A and Bennett, David A},
  journal={Archives of neurology},
  volume={68},
  number={3},
  pages={351--356},
  year={2011},
  publisher={American Medical Association}
}

@article{biessels2018cognitive,
  title={Cognitive decline and dementia in diabetes mellitus: mechanisms and clinical implications},
  author={Biessels, Geert Jan and Despa, Florin},
  journal={Nature Reviews Endocrinology},
  volume={14},
  number={10},
  pages={591--604},
  year={2018},
  publisher={Nature Publishing Group UK London}
}

@article{chen2026emad,
  title={EMAD: Evidence-Centric Grounded Multimodal Diagnosis for Alzheimer's Disease},
  author={Chen, Qiuhui and Yao, Xuancheng and Zhou, Zhenglei and Hu, Xinyue and Hong, Yi},
  journal={arXiv preprint arXiv:2602.19178},
  year={2026}
}

@article{chen2023automatic,
  title={Automatic Detection of Alzheimer's Disease with Multi-Modal Fusion of Clinical MRI Scans},
  author={Chen, Long and Chen, Liben and Xu, Binfeng and Zhang, Wenxin and Razavian, Narges},
  journal={arXiv preprint arXiv:2311.18245},
  year={2023}
}

@article{leming2025differential,
  title={Differential dementia detection from multimodal brain images in a real-world dataset},
  author={Leming, Matthew and Im, Hyungsoon},
  journal={Alzheimer's \& Dementia},
  volume={21},
  number={7},
  pages={e70362},
  year={2025},
  publisher={Wiley Online Library}
}

@inproceedings{prabhu2022multi,
  title={Multi-modal deep learning models for alzheimer's disease prediction using mri and ehr},
  author={Prabhu, Sathvik S and Berkebile, John A and Rajagopalan, Neha and Yao, Renjie and Shi, Wenqi and Giuste, Felipe and Zhong, Yishan and Sun, Jimin and Wang, May D},
  booktitle={2022 IEEE 22nd International Conference on Bioinformatics and Bioengineering (BIBE)},
  pages={168--173},
  year={2022},
  organization={IEEE}
}

@inproceedings{liu2020design,
  title={On the design of convolutional neural networks for automatic detection of Alzheimer’s disease},
  author={Liu, Sheng and Yadav, Chhavi and Fernandez-Granda, Carlos and Razavian, Narges},
  booktitle={Machine learning for health workshop},
  pages={184--201},
  year={2020},
  organization={PMLR}
}

@article{liu2022generalizable,
  title={Generalizable deep learning model for early Alzheimer’s disease detection from structural MRIs},
  author={Liu, Sheng and Masurkar, Arjun V and Rusinek, Henry and Chen, Jingyun and Zhang, Ben and Zhu, Weicheng and Fernandez-Granda, Carlos and Razavian, Narges},
  journal={Scientific reports},
  volume={12},
  number={1},
  pages={17106},
  year={2022},
  publisher={Nature Publishing Group UK London}
}

@article{liu2021development,
  title={Development of a Deep Learning Model for Early Alzheimer’s Disease Detection from Structural MRIs and External Validation on an Independent Cohort},
  author={Liu, Sheng and Masurkar, Arjun V and Rusinek, Henry and Chen, Jingyun and Zhang, Ben and Zhu, Weicheng and Fernandez-Granda, Carlos and Razavian, Narges and Alzheimer’s Disease Neuroimaging Initiative},
  journal={medRxiv},
  pages={2021--05},
  year={2021},
  publisher={Cold Spring Harbor Laboratory Press}
}

@article{fortea2024apoe4,
  title={APOE4 homozygosity represents a distinct genetic form of Alzheimer’s disease},
  author={Fortea, Juan and Pegueroles, Jordi and Alcolea, Daniel and Belbin, Olivia and Dols-Icardo, Oriol and Vaque-Alcazar, Lidia and Videla, Laura and Gispert, Juan Domingo and Suarez-Calvet, Marc and Johnson, Sterling C and others},
  journal={Nature medicine},
  volume={30},
  number={5},
  pages={1284--1291},
  year={2024},
  publisher={Nature Publishing Group US New York}
}

@article{jung2022transferability,
  title={Transferability of Alzheimer disease polygenic risk score across populations and its association with Alzheimer disease-related phenotypes},
  author={Jung, Sang-Hyuk and Kim, Hang-Rai and Chun, Min Young and Jang, Hyemin and Cho, Minyoung and Kim, Beomsu and Kim, Soyeon and Jeong, Jee Hyang and Yoon, Soo Jin and Park, Kyung Won and others},
  journal={JAMA network open},
  volume={5},
  number={12},
  pages={e2247162},
  year={2022}
}

@article{brum2023two,
  title={A two-step workflow based on plasma p-tau217 to screen for amyloid $\beta$ positivity with further confirmatory testing only in uncertain cases},
  author={Brum, Wagner S and Cullen, Nicholas C and Janelidze, Shorena and Ashton, Nicholas J and Zimmer, Eduardo R and Therriault, Joseph and Benedet, Andrea L and Rahmouni, Nesrine and Tissot, Cecile and Stevenson, Jenna and others},
  journal={Nature Aging},
  volume={3},
  number={9},
  pages={1079--1090},
  year={2023},
  publisher={Nature Publishing Group US New York}
}

@article{gauthier2006mild,
  title={Mild cognitive impairment},
  author={Gauthier, Serge and Reisberg, Barry and Zaudig, Michael and Petersen, Ronald C and Ritchie, Karen and Broich, Karl and Belleville, Sylvie and Brodaty, Henry and Bennett, David and Chertkow, Howard and others},
  journal={The lancet},
  volume={367},
  number={9518},
  pages={1262--1270},
  year={2006},
  publisher={Elsevier}
}

@article{albert2011diagnosis,
  title={The diagnosis of mild cognitive impairment due to Alzheimer's disease: recommendations from the National Institute on Aging-Alzheimer's Association workgroups on diagnostic guidelines for Alzheimer's disease},
  author={Albert, Marilyn S and DeKosky, Steven T and Dickson, Dennis and Dubois, Bruno and Feldman, Howard H and Fox, Nick C and Gamst, Anthony and Holtzman, David M and Jagust, William J and Petersen, Ronald C and others},
  journal={Alzheimer's \& dementia},
  volume={7},
  number={3},
  pages={270--279},
  year={2011},
  publisher={Wiley Online Library}
}

@article{sperling2011toward,
  title={Toward defining the preclinical stages of Alzheimer’s disease: Recommendations from the National Institute on Aging-Alzheimer's Association workgroups on diagnostic guidelines for Alzheimer's disease},
  author={Sperling, Reisa A and Aisen, Paul S and Beckett, Laurel A and Bennett, David A and Craft, Suzanne and Fagan, Anne M and Iwatsubo, Takeshi and Jack Jr, Clifford R and Kaye, Jeffrey and Montine, Thomas J and others},
  journal={Alzheimer's \& dementia},
  volume={7},
  number={3},
  pages={280--292},
  year={2011},
  publisher={Elsevier}
}

@article{sirkis2022dissecting,
  title={Dissecting the clinical heterogeneity of early-onset Alzheimer’s disease},
  author={Sirkis, Daniel W and Bonham, Luke W and Johnson, Taylor P and La Joie, Renaud and Yokoyama, Jennifer S},
  journal={Molecular psychiatry},
  volume={27},
  number={6},
  pages={2674--2688},
  year={2022},
  publisher={Nature Publishing Group UK London}
}

@article{loi2023young,
  title={Young-onset dementia diagnosis, management and care: a narrative review},
  author={Loi, Samantha M and Cations, Monica and Velakoulis, Dennis},
  journal={Medical Journal of Australia},
  volume={218},
  number={4},
  pages={182--189},
  year={2023},
  publisher={Wiley Online Library}
}

\clearpage
\section{Extended Methods}
\subsection{Super agent for orchestration and action trajectory planning}
\CEREBRA is built upon a large language model (LLM) backbone. In our experiments, we employ GPT-4o and GPT-OSS-120B\footnote{Due to limited HIPAA-compliant model offerings, experiment with UF and INPC cohort is conducted with GPT-OSS-120B model.}~\cite{agarwal2025gpt} as an example implementation for the results, through our code base supports other LLM APIs as well as open-sourced models through vLLM~\cite{kwon2023efficient}, LiteLLM~\cite{litellm_github}, and Ollama~\cite{ollama_github}. The super agent uses the LLM backbone model to analyze task instructions, which is provided by the user in natural language, decides a problem-solving path by referring to the downstream agents that \CEREBRA has access to, and calls the relevant agents with appropriate input parameters. To support determining the optimal trajectory, a wide range of metadata is provided to the super agent, such as information about available agents and their tools, with descriptions of the function and example tool calling commands. Depending on the need, SQL database descriptions for tables and columns are also provided for a high-level understanding of available data. Note that we are only giving the database schema to the agent without exposing any real data, further ensuring top-level \textbf{HIPAA-compliant data privacy} for real-world clinical scenarios. We discuss various available agents in the following sections, each of which is designed for specific purposes (e.g. data query, data processing, model training and inference for certain modality, and aggregated output summarization).

\subsection{Data agent}
To enable efficient, code-free access to structured clinical data, we implemented a data agent that supports natural-language database querying via a Text-to-SQL~\cite{hong2025next} interface. The agent is initialized with database metadata, including table schemas and column descriptions, which are used by a large language model (LLM) to translate user queries into executable SQL. Each calling of the data agent operates on a single data modality to ensure clarity and reproducibility. In our experiment, we employed GPT-4o as the backbone LLM for data agent. To optimize query generation and downstream processing, the data agent operates in three modes: training, inference, and exploration, each guided by task-specific prompts. Upon receiving a user query, the agent automatically selects the most appropriate mode based on the inferred intent.

\paragraph{Training mode.}
In training mode, the agent retrieves a complete cohort satisfying the criteria specified in the natural-language query, including dementia cases and corresponding labels. The retrieved data are then standardized for downstream model training and split at the patient level into training, validation, and test sets to prevent data leakage. Modality-specific representations are generated: EHR data are encoded as SciPy sparse matrices, clinical notes are organized as per-patient text sequences, and imaging data are referenced via paths to pre-processed image files.

\paragraph{Inference mode.}
In inference mode, the agent retrieves the longitudinal medical history of an individual patient, optionally constrained by temporal or clinical criteria specified in the query. The data are processed using the same modality-specific pipelines as in training mode, but without associated outcome labels.

\paragraph{Exploration mode.}
The exploration mode enables ad hoc database interrogation through the user dashboard. This mode supports tasks such as reviewing patient medical histories under specific constraints, identifying cohorts with similar clinical profiles, and inspecting database schemas and metadata, facilitating both hypothesis generation and data understanding.
\newline

For extra flexibility, the data agent also accepts pro-processed dataset, allowing users to use the agent without setting up the SQL database.

\subsection{Modality agents}
\label{sec:modality_agents_method}
We implemented agents for data modalities (EHR, note and image) since each of which requires specialized handling to train models and to extract relevant insights from the data. Each agent contains a number of available tools for training, inference or insight extractions at disposal of the agent, who will decide a tool calling strategy based on the task and the data.

\subsubsection{Image agent}
The image agent will process, train, inference and extract insights from input image datasets generated by the data agent. We implemented a series of tools for different types of images (brain MRI and eye imaging), which the image agent can decide on calling the most appropriate ones for the prompted task. Similar to prior studies with brain MRI data~\cite{popuri2020using, hwang2022disentangling} and retinal imaging data~\cite{wang2025retinal, ibrahim2023systematic}, we utilizes segmentation tools to extract volumes from medical images as features for the machine learning model.

\paragraph{Brain MRI training tool.}
The image agent supports end-to-end modeling of structural brain MRI by orchestrating a dedicated MRI training tool. During training, raw T1-weighted scans are processed with FreeSurfer SynthSeg~\cite{BILLOT2023102789} to obtain automated whole-brain segmentation and region-level volumetric measurements. These regional volumes are normalized and used as structured inputs to train a machine learning (ML) model for downstream prediction tasks. The image agent can choose from a variety of ML models, including gradient-boosted decision tree model (XGBoost~\cite{chen2016xgboost}), logistic regression~\cite{hosmer2013applied} and random forests regression~\cite{breiman2001random}. For binary prediction tasks, models are optimized using binary logistic loss, while for time-to-event analysis in Cox survival modeling, negative partial log-likelihood loss functions are employed to account for censoring and variable follow-up durations. The training tool also performs hyperparameter tuning with grid search to find best performing models. This design offers the option of using high-fidelity anatomical volume features with interpretable ML approaches, enabling robust learning on interpretable features for subsequent agentic workflows.

\paragraph{Brain MRI inference and insight extraction tool.}
At inference time, the same segmentation and feature extraction pipeline is applied to unseen scans to produce region-wise volumes, which are then passed to the trained model to generate risk estimates. To facilitate interpretability, the image inference tool, if the model permits, further performs insight extraction by mapping the model’s top-ranked feature importances back to their corresponding anatomical regions. These salient regions are overlaid as highlights on the original MRI, providing spatially grounded explanations that link model outputs to neuroanatomical substrates and enabling intuitive visual inspection by clinicians. The MRI scan with highlighted regions is presented as a montage across axial, sagittal and coronal axes, which presents rich insights suitable for diversified clinical needs.

\paragraph{Retinal OCT training and inference tool.}
Similar to the training tool for brain MRI, the image agent also supports retinal OCT modality by using normalized thickness features extracted from the retinal scans, as the details provided in Appendix.~\ref{appendix:sec_data_imaging}. The training processes for retinal scans are similar to that of brain MRI, where image agent can choose from XGBoost, logistic regression and random forests regression for risk predictions, performs relevant hyperparameter search, while also following resemble loss function selection. At inference time, the feature extraction pipeline is applied to unseen scans to produce all thickness measures, which are then passed to the trained model to generate risk estimates.

\subsubsection{Note agent}
We aim to extract traces that may connect to dementia from the patient’s longitudinal clinical and radiology notes. The notes contain more subjective descriptions and evaluations of the patient’s health condition, such as from the subject chief complaint sections where patients self-report relevant complaints that may be relevant to the disease or that establish the health baseline in general, or cognitive exam results, which may serve as complementary information to the more structured EHR. Similar to previous work~\cite{mao2023ad, perfalk2024predicting}, we use natural language processing (NLP) and machine learning models to extract relevant evidence that contributes to higher or lower risks of dementia.

\paragraph{Note training tool.} 
During training, we concatenated the full list of historical notes into an aggregated string and then encoded by the SentenceAttentionBERT model in transformers and PyTorch package~\cite{wolf2020transformers, paszke2019pytorch} to produce contextualized sentence-level representations that capture both local semantics and document-level relevance. An attention mechanism is then applied to weight clinically salient sentences, allowing the model to focus on content most informative for dementia risk prediction. The embeddings are then used to train a multi-layer perceptron with apporiate loss function. For prediction and diagnosis tasks, binary cross-entropy with logits is used as the loss function, while for time-to-event, PyCox with CoxPH loss function is used as the objective function for survival analysis~\cite{katzman2018deepsurv, kvamme2019time}.

\paragraph{Note inference tool and evidence extraction with risk factor categorization.} 
In addition to predicting risk scores for the longitudinal medical notes, we designed the inference tool to extract dementia-related evidence by finding the most attention-weighted sentences as top features. Since the attention of the model is unidirectional and only represents magnitude, we pass the selected sentences to an LLM (same selection as with super agent) to categorize them as positive, neutral or negative risk factors to the disease. The categorization, as well as the magnitude of attention for the sentence, are then used as extracted evidence from the notes to be utilized by downstream workflows.

\subsubsection{EHR agent}
Electronic health records (EHRs) reflect a fine-grained medical snapshot of a patient with detailed diagnoses, medications, laboratory and demographics. We follow the approach of previous work to process EHR data and to train a machine learning model~\cite{tang2024leveraging, li2023early}. Specifically, we processed the EHR into a daily, longitudinal representation such that the relevant medical concept class’s value is recorded. Similar to other modality agents, the EHR agent is designed not only to generate the risk score, but also to extract interpretable, clinically meaningful patterns that contribute to dementia risk.

\paragraph{EHR training tool.}
From the data agent, raw EHR data are mapped to standardized medical concept vocabularies (diagnosis, medication, laboratory and demographic code systems) and aggregated into a high-dimensional, temporally ordered sparse matrix representation. The processed EHR features are then used to train with the agent-selected ML model (XGBoost, logistic regression, random forest regression). Similarly, the aforementioned loss function choices for regression or survival analysis will be employed on demand, where the EHR agent will configure and call the tool with the most appropriate loss setup for the target task. Details on the number of EHR features, categories and processing pipelines are in Supplementary Section.~\ref{sec:sup_data_extraction_and_preparation_ehr}.

\paragraph{EHR inference tool and evidence extraction with risk factors.}
At inference time, the trained EHR model is applied to unseen patient trajectories to generate individualized dementia risk estimates. To support interpretability, we extract model-attributed importance scores for individual medical concepts and concept families, identifying diagnoses, medications, or laboratory abnormalities that most strongly influence the predicted risk. To translate model outputs into clinically relevant evidence, the top contributing EHR features are categorized into positive, neutral, or negative risk factors using LLM based on their directional association with dementia risk. The categorized risk factors, together with their relative importance and temporal context, form a structured evidence set that can be consumed by downstream agentic workflows, enabling transparent reasoning over EHR-derived signals for downstream cross-modality evidence integration.

\subsection{Summary agent}
\label{method:summary_agent}
To better utilize risk evaluations from diversified modalities, we designed the summary agent to synchronize outputs from modality agents. Data being processed by respective modality agents provide modality-specific risk scores and clinically grounded evidence for designated tasks while showcasing heterogeneous and complementary insights, as shown in Figure.~\ref{fig:modality}d, where each modality possesses diversified unique risk evaluations, incentivizing us to fuse them together for a more holistic perspective of the case.
\paragraph{Multi-agent discussion.}
The summary agent aggregates predictions from $n$ modality-specific agents and coordinates a structured deliberation process to produce a final patient-level assessment. Each modality agent first outputs a risk estimate and supporting evidence derived from its corresponding data source (e.g., clinical notes, imaging, or EHR features).

To prioritize potentially critical findings, the modality agent with the highest predicted risk is selected as the primary proposer, presenting its risk assessment and justification. The bottom $m $ modality agents with the lowest predicted scores serve as review agents, evaluating the proposal by comparing it with their modality-specific evidence and risk estimates. Review agents may provide supporting evidence, highlight inconsistencies, or identify missing context. The summary agent then synthesizes the discussion by integrating the proposed risk assessment, reviewer feedback, and cross-modal evidence consistency. The final output includes a consolidated risk prediction, confidence score, and structured rationale grounded in multimodal evidence.

\paragraph{Risk aggregation.}
For the risk prediction task, the summary agent does not generate a new risk estimate independently. Instead, it produces a consensus risk bounded by the minimum and maximum risk estimates proposed by the modality agents. The final risk is determined by synchronizing reviewer feedback and cross-modal evidence consistency while ensuring the prediction remains within this range.

\subsection{Study Cohorts}
In this section, we report datasets from each institution and the study cohort selection criteria. Detailed cohort and data statistics are reported in Supplementary Table.~\ref{tab:cohort_statistics}.

\paragraph{New York University Langone.} Data were obtained from all NYU Langone hospitals (NYU) and affiliated clinics within the health system, covering the inclusion period from January 1, 2013 through January 1, 2023. To define an appropriate cohort for dementia-related analyses, we restricted the study population to patients who were at least 65 years of age as of the end of the dataset cutoff date (January 1, 2023), corresponding to the end of the data collection period. Consequently, all included patients were at least 55 years old at the earliest time point in their available records. The inclusion criteria, modality-specific patient counts, and the final cohort size are summarized in Table.~\ref{tab:cohort_statistics}.

We additionally identified a Long Island cohort to assess the generalizability of \CEREBRA. NYU Langone Hospital – Long Island (LI) is a recently integrated hospital within the NYU Langone system that shares the same electronic health record infrastructure but previously operated independently, with distinct physician cohorts and modest differences in clinical equipment. Accordingly, we analyzed the Long Island cohort separately from other NYU hospitals as an ``external'' evaluation dataset, and we used the models trained on NYU dataset to perform evaluation directly on LI dataset without retraining. The resulting NYU cohort included 52,843 unique patients, comprising approximately 369 million EHR records, 1 million note records, and 139,000 brain MRI images, while the LI cohort included 508 unique patients, comprising approximately 3 million EHR records, 11,000 medical notes, and 3,349 images.

\paragraph{Indiana Network for Patient Care (INPC).} INPC, standardized to the OMOP Common Data Model (CDM), aggregates real-world data from 130+ health organizations, covering >19 million patients across Indiana\footnote{\url{https://www.regenstrief.org/rds/data/}}. Using INPC, we constructed a base cohort consisting of patients with at least one recorded diagnosis of mild cognitive impairment (MCI), identified using International Classification of Diseases, Ninth Revision, Clinical Modification (ICD-9-CM) and International Classification of Diseases, Tenth Revision, Clinical Modification (ICD-10-CM) diagnosis codes. We excluded patients with any diagnosis of Alzheimer’s disease and related dementias (AD/ADRD) occurring on or before the MCI index date. The full list of ICD-10-CM codes used to define MCI and ADRD is provided in Supplementary Table.~\ref{tab:ad_criteria}. The resulting cohort included 37{,}402 unique patients, comprising approximately 16 million condition diagnosis records, with EHR data spanning from January 1, 2015 to April 14, 2025. Similarly, we restricted the study population to patients who were at least 65 years of age as of April 14, 2025 and had structured EHR data and unstructured clinical notes available. This cohort represents a clinically well-recognized population at elevated risk for progression to ADRD and served as the starting population for AD/ADRD progression prediction task. 

\paragraph{University of Florida Health.}
Data were sourced from the University of Florida Health (UF) system through Integrated Data Repository Research Services\footnote{\url{https://idr.ufhealth.org/}}. We defined a base cohort comprising patients who were at least 65 years old as of July 3, 2024 (the end of the data collection period), which implies that all included individuals were at least 55 years old at the earliest time point in their available records. Eligible patients were additionally required to have at least one optical image and at least one clinical narrative or radiology note within the inclusion window to ensure multimodal data availability for each participant.

The final base cohort included 4,393 unique patients, comprising 28,759 retinal OCT images and 96,048 condition diagnosis records, with EHR data spanning December 2, 2011 to July 3, 2024. All included patients had structured EHR data, unstructured clinical notes, and retinal OCT imaging available. This cohort ensures a complete multimodal representation of each individual, laying the foundation for all downstream task-specific analysis.

\subsection{Benchmarking \CEREBRA on a diverse set of clinical tasks}
\label{sec:method_dataset}

To evaluate the effectiveness of the \CEREBRA agent system, we constructed a suite of benchmark tasks that reflect heterogeneous clinical decision-making scenarios. Specifically, we consider three categories of tasks commonly encountered in practice: disease risk prediction, diagnosis, and longitudinal disease progression modeling. To ensure that these tasks are clinically meaningful and methodologically rigorous, we designed a dataset construction pipeline that carefully defines temporal windows and prevents label leakage from noisy real-world electronic health records. The overall framework for dataset construction is described in this section.

Importantly, these tasks represent complementary clinical decision-making scenarios that arise throughout the dementia care pathway. Dementia forecasting evaluates early risk prediction before symptoms become clinically apparent, diagnosis focuses on identifying existing disease from accumulated medical history, disease progression modeling captures longitudinal risk dynamics after baseline assessment, and the MCI-to-ADRD conversion task evaluates risk stratification among patients already exhibiting cognitive impairment. These tasks span prospective prediction, diagnostic assessment, and longitudinal disease modeling, allowing us to evaluate whether \CEREBRA provides consistent benefits across diverse clinical objectives and temporal settings. Across tasks, we used an 80/10/10 split of the NYU, UF, and INPC cohorts into training, validation, and test sets, respectively. The LI cohort was reserved as a held-out dataset and used exclusively for evaluations of models trained on the NYU cohort. A visualization of dataset curation logics presented in Appendix.~\ref{sec:additional_dataset_curation_details}.

\subsubsection{Observation, prediction, and label windows with rolling operation}

We define three temporal windows to structure each prediction task: the observation window, the prediction window, and the label window. The observation window specifies the time period from which patient data are used as model inputs. The end of the observation window is referred to as the \emph{index date}, which represents the current state of the patient at inference time. The prediction window spans the interval between the index date and the start of the label window. The label window defines the time interval during which clinical events are used to determine the outcome label. Separating these windows prevents future clinical information from leaking into the model inputs, which is particularly important when working with longitudinal real-world clinical data.

\subsubsection{Obtaining labels from EHR via medical concept matching}

Outcome labels are obtained by matching medical concept tokens within the patient's electronic health record. Specifically, the onset of AD/ADRD is defined as the first visit containing AD/ADRD-related diagnosis codes or medications, as listed in Supplementary Table~\ref{tab:ad_criteria}. 

If any of these concept tokens occur within the label window, the patient is assigned a positive label for AD/ADRD. Otherwise, the patient is considered negative within that window.

\subsubsection{Dementia forecasting task}

The dementia forecasting task aims to predict the risk of a patient being diagnosed with AD/ADRD in a future time horizon.

For this task, the prediction window is fixed to 180 days across most analyses. The label window corresponds to the time horizon over which incident dementia is assessed. We evaluate prediction horizons of 1, 2, and 3 years.

The observation window is constructed using a rolling-window procedure. For each patient, we begin with a six-month observation window starting from the earliest available record. The window is then extended in six-month increments to generate additional samples. For each iteration, the observation window is followed by the prediction window and then the label window.

This rolling-window design allows multiple training samples to be generated per patient while preserving the correct temporal ordering of clinical information. The final rolling iteration is constrained so that the label window remains at least six months away from the end of the cohort observation period, ensuring sufficient follow-up time.

To focus the task on forecasting new disease onset, we exclude any samples where AD/ADRD-related concept tokens occur within the observation or prediction window. A visualization of the dataset construction process is provided in Supplementary Fig.~\ref{fig:dataset_logic}c.

\subsubsection{Disease diagnosis task}

The dementia diagnosis task evaluates whether the model can identify patients with dementia based on their complete medical history up to a specific time point.

We first identify visit dates for each patient, defined as dates on which any medical concept tokens appear in the EHR. For patients with dementia-related concept tokens, the index date is defined as the date of the first occurrence of such tokens. For patients without dementia diagnoses, a visit date is randomly selected as the index date. The first visit is excluded as a candidate index date to ensure that prior medical history exists for model input.

The observation window spans from the beginning of the available records up to the index date. This setup reflects a clinical diagnostic scenario in which physicians evaluate a patient's full historical record when assessing possible dementia. Dataset construction for this task is illustrated in Supplementary Fig.~\ref{fig:dataset_logic}a.

\subsubsection{Disease progression prediction task}

To model disease progression, we formulate a survival analysis task using a Cox proportional hazards framework. The index date is defined as the date of the patient’s first available imaging scan. The observation window includes all clinical data prior to this date.

For patients with dementia, the time-to-event label is defined as the interval between the index date and the first occurrence of dementia-related concept tokens. For patients without dementia, the time-to-event is defined as the interval between the index date and the end of the cohort observation period, with a censoring indicator used for survival modeling.

To prevent label leakage, any patients with dementia-related concept tokens occurring before the index date are excluded from the analysis. The dataset construction procedure is illustrated in Supplementary Fig.~\ref{fig:dataset_logic}b.

\subsubsection{MCI-to-ADRD conversion task}

To evaluate the generalizability of \CEREBRA, we additionally consider the clinically important task of predicting conversion from MCI to AD/ADRD. The prediction cohort consists of patients with at least one recorded MCI diagnosis and no AD/ADRD diagnosis occurring on or before the first MCI diagnosis.

The index date is defined as the date of the first recorded MCI diagnosis, representing the clinically relevant time point at which clinicians begin assessing the risk of progression to dementia. To simulate a realistic clinical setting and prevent temporal leakage, model inputs are constructed using only data occurring before the index date (i.e. period when the patient is still clinically normal cognition). The observation window hence spans from the start of available records (January 1, 2015) up to, but not including, the index date.

The outcome of interest is incident AD/ADRD occurring after the MCI diagnosis. We evaluate prediction performance across three time horizons: 1 year, 2 years, and 3 years following the index date. For each horizon $H \in \{1,2,3\}$ years, patients are labeled as converters ($y=1$) if their first AD/ADRD diagnosis occurs within

\[
(\text{index date},\ \text{index date} + H].
\]

Patients are labeled as non-converters ($y=0$) if no AD/ADRD diagnosis occurs within the corresponding time window and if they have sufficient follow-up duration to cover the entire prediction horizon. Patients without sufficient follow-up are excluded to avoid label uncertainty.

The dataset construction process for the conversion task follows the general framework described above, with the additional requirement that patients must have an MCI diagnosis within the observation window.

\subsection{Ablation study on incomplete and heterogeneous inputs}
One advantage of \CEREBRA is its ability to flexibly incorporate heterogeneous modalities while remaining robust to missing or redundant information sources. To quantify the contribution of each modality and evaluate the effectiveness of the agentic fusion strategy, we conducted a series of ablation studies.

We first compared the full multimodal system against models trained on individual modalities alone, including EHR, clinical notes, and imaging, under identical training and evaluation protocols. This analysis isolates the marginal predictive value of each modality and highlights modality-specific strengths across prediction horizons.

To evaluate whether performance gains arise from the agentic fusion mechanism rather than simple score aggregation, we further compared against non-learned fusion baselines that combine modality-specific risk predictions using \textbf{minimum}, \textbf{maximum}, and \textbf{arithmetic mean} operators. These baselines control for improvements attributable solely to ensembling or aggregation effects.

In addition, to assess robustness to modality availability, we simulated missing-modality settings by systematically removing one modality at inference time, and extra-modality settings by introducing additional modality agents whose predictions may be weakly informative or noisy. These experiments evaluate whether the summary agent can appropriately balance modality-specific evidence without being dominated by spurious signals.

Across tasks, the proposed agentic fusion consistently outperformed both single-modality models and heuristic fusion baselines, indicating that performance improvements arise from structured cross-modal deliberation rather than dominance of a single modality or trivial ensembling effects.

\subsection{Interactive Dashboard}

The interactive dashboard provides the primary interface through which clinicians interact with the \CEREBRA agentic system. Rather than serving as a passive visualization layer, the dashboard is designed as a bidirectional control and oversight channel that supports querying, feedback, and decision support within clinical workflows.

\begin{figure}[H]
    \centering
    \includegraphics[width=0.85\linewidth]{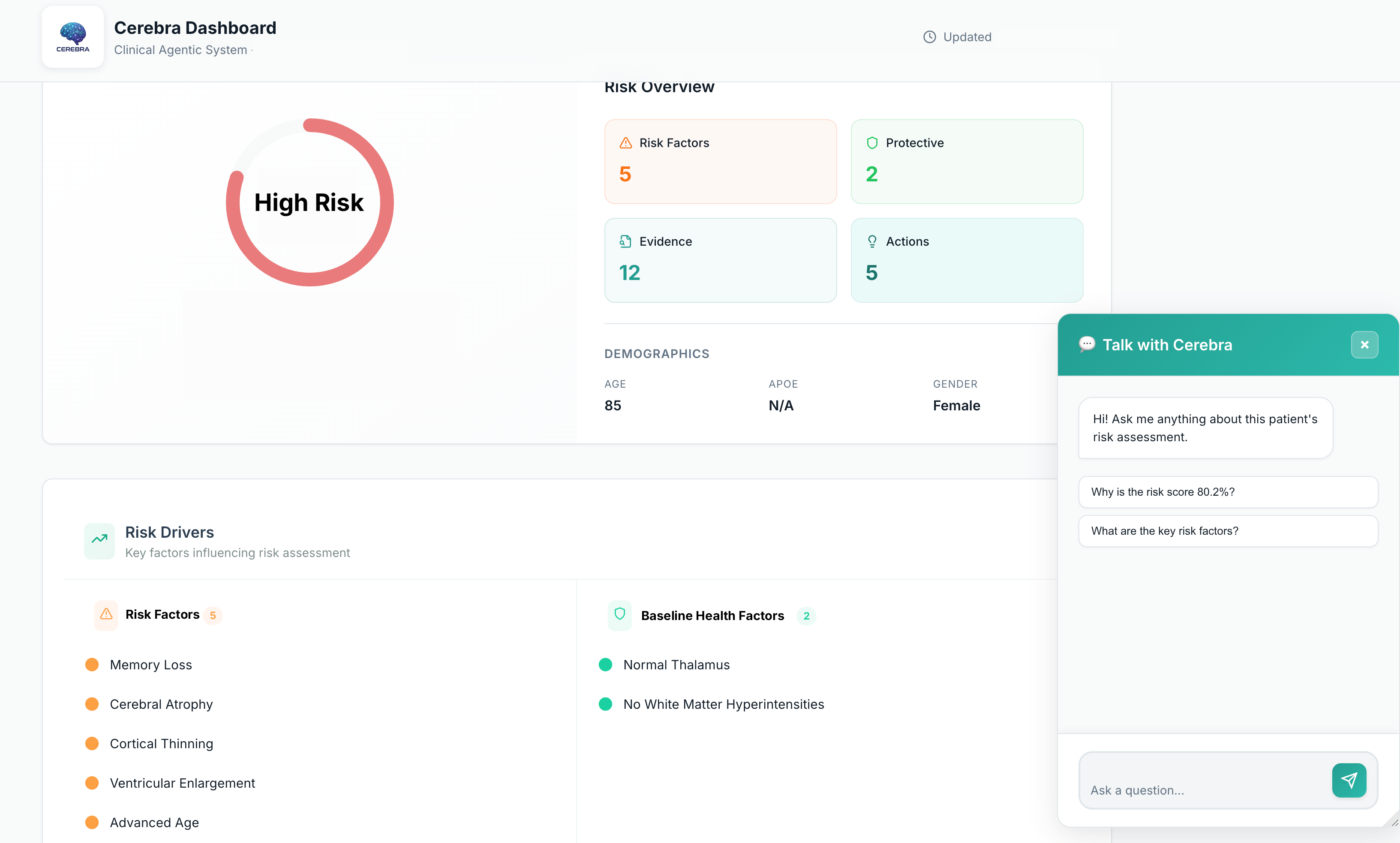}
    \caption{\textbf{Interactive interface of \CEREBRA }. Cerebra provides a conversational interface in the dashboard that enables clinicians to interact with \CEREBRA, exploring the reasoning behind the decision and providing feedback to guide the system.}
    \label{fig:dashboard_example}
\end{figure}

The dashboard (see example in Fig.~\ref{fig:dashboard_example}) includes a conversational interface with two core functions. First, it allows clinicians to issue natural-language queries and task requests, which are routed to the \CEREBRA super agent orchestrator and decomposed into agent-level actions across models, tools, and data sources. Second, it enables structured human feedback that can be used to steer system behavior, including triggering targeted model updates, revising agent policies, or updating a dynamic medical notebook that stores clinician-validated insights and experience for future reasoning. 

In addition to interaction and control, the dashboard presents a patient-centric summary view inspired by consumer-facing health reports (e.g., genetic testing dashboards), but adapted for clinical use. This view integrates calibrated risk estimates, salient risk factors, and supporting evidence grounded in raw clinical data such as electronic health records, clinical notes, and medical imaging. Importantly, all model outputs are accompanied by uncertainty estimates and evidence traces to support clinical interpretation and trust calibration.

Finally, the dashboard surfaces recommendations intended to support, rather than replace, clinician decision-making. These recommendations may include suggested follow-up tests, monitoring strategies, or referral considerations, and are explicitly framed as decision support with clear provenance and confidence indicators, allowing clinicians to exercise judgment and oversight at every step.

\subsection{\CEREBRA-augmented clinician assessments}
To evaluate the clinical utility of the interactive dashboard, we conducted a reader study with clinicians using a randomized cross-over design. Six clinicians with varying levels of clinical experience in dementia care (training level and years in practice collected via survey) participated in the study. All participants provided informed consent prior to participation.

We selected 40 patient cases (20 with high risk and 20 with low risk) from the held-out evaluation cohort and divided them into two matched sets (Set A and Set B; 20 cases each) with similar distributions of dementia outcomes and clinical complexity. Each clinician evaluated all 40 cases, including 20 cases without dashboard assistance and 20 cases with dashboard assistance. Participants were randomly assigned to one of two reading sequences to mitigate ordering effects: (1) Set A without dashboard followed by Set B with dashboard, or (2) Set A with dashboard followed by Set B without dashboard. Case order within each set was randomized independently for each participant. Survey questions and response options are reported in Appendix.~\ref{sec:reader_study_appendix}.

For each case, clinicians were provided with standardized clinical information derived from multimodal patient data, including demographics, relevant clinical history, and summarized findings from available modalities. In the dashboard condition, clinicians were additionally provided with the interactive system interface displaying predicted dementia risk scores, identified risk factors, supporting evidence extracted from clinical notes and imaging summaries, and recommended clinical considerations.

\paragraph{Participant Characteristics} Our evaluation involved a diverse cohort of six medical professionals, categorized into two primary groups: Neurologists ($n=3$) and Primary Care Physicians (PCPs, $n=3$). The neurology group represented a broad spectrum of expertise, including one senior attending with over 10 years of experience, one junior attending (under 5 years), and one resident/fellow. Within this group, clinical exposure to the target condition varied, with two specialists dedicating over 75\% of their practice to related cases. The PCP group consisted of three attendings, two seniors, and one junior, all of whom manage a moderate volume of relevant cases (10--25\% of their practice). This varied composition ensures that the \CEREBRA dashboard was evaluated across different levels of seniority and clinical specializations, reflecting its potential utility in both expert and generalist settings.

\paragraph{Prediction assessment.}
For both dashboard and non-dashboard conditions, clinicians completed three core assessment questions per case:
\begin{itemize}
    \item estimated dementia risk within the next 3 years (Low, Moderate, High, or Unable to determine),
    \item most likely diagnostic category within the next 3 years (multi-label selection among major dementia etiologies), 
    \item confidence in their assessment (Low, Moderate, or High).
\end{itemize}

These responses were used to evaluate prediction accuracy and clinician confidence across conditions.

\paragraph{Dashboard evaluation.}
For cases evaluated with dashboard assistance, clinicians additionally completed a structured questionnaire assessing the interpretability and clinical usefulness of the dashboard. The questionnaire evaluated:
the dashboard component that most influenced decision-making, the perceived accuracy and clinical relevance of displayed risk factors, the accuracy and alignment of supporting evidence, whether incorrect or missing evidence was identified, the appropriateness of recommendations, and perceived usefulness for understanding patient risk and facilitating decision-making.

\paragraph{Outcome measures.}
The primary endpoint was clinician prediction performance for 3-year dementia risk with versus without dashboard assistance. Secondary endpoints included diagnostic category agreement, clinician confidence, and subjective evaluations of dashboard utility and reliability. We report performance metrics as mean $\pm$ one standard deviation (SD) derived from 1,000 bootstrap iterations. 

\subsection{Dynamic medical notebook}
\label{sec:method_dynamic_medical_notebook}

General-purpose language models are not inherently optimized for clinical diagnosis or risk prediction and may exhibit inconsistent reasoning when applied to complex, longitudinal medical data. To address this limitation, we introduce a Dynamic Medical Notebook that distills recurring clinical patterns, risk factors, and diagnostic cues from prior cases into a structured, continuously updatable knowledge representation. Rather than relying solely on the knowledge of LLMs, the notebook serves as an explicit external reference that captures domain-relevant insights derived from historical patient trajectories. This design enables the model to condition its reasoning on accumulated clinical experience, promoting more consistent, clinically grounded interpretation and reducing reliance on ad hoc inference during evaluation. Feedback can be incorporated by including physician-authored comments, corrections, and summary judgments in human language into the memory module. To do that, \CEREBRA collects feedback from users (clinicians) for good practices and for corrective reasons on patient reports that they have second opinions on. The feedback will then be processed into a set of textual ``memory'' that is maintained by LLM (details in Appendix.~\ref{appendix:dynamic_medical_notebook}) for reference in future \CEREBRA workflows. The notebook remains continuously extensible and reflective of evolving clinical understanding for iterative refinement of as a reasoning copilot.

\paragraph{Model-generated memories for evaluation. }
To evaluate the scalability of the dynamic medical notebook, we simulated physician feedback at scale by distilling knowledge from mispredicted cases. GPT-4o served as an expert auditor, reviewing cases in which \CEREBRA underestimated risk despite a positive outcome or overestimated risk despite a negative outcome, and converting these discrepancies into a generalizable set of dementia-related memories. Following related approaches~\cite{suzgun2025dynamic, yuksekgonul2024textgrad}, we framed this process as a problem-solving task in which the LLM was given extracted evidence together with the ground-truth label and asked to identify risk factors that could improve subsequent predictions. This process produced a series of evidence-grounded memory updates, each paired with a judgment indicating whether the predicted risk should move higher or lower within the range defined by the lowest and highest modality-specific risk scores from the summary agent. We then incorporated these updates into the medical notebook only when they contributed novel, non-conflicting findings relative to the existing memory store. The updated notebooks were subsequently evaluated to assess the effectiveness and scalability of continuously refined, medically grounded memory for dementia forecasting.

\subsection{Validation of reasoning capability}
\paragraph{Setup.}
To evaluate the efficacy and accuracy of \CEREBRA's reasoning capability, we employed ICARE~\cite{dua2025clinically}, an automatic, clinically grounded, agent-based evaluation framework. ICARE assesses report accuracy by symmetrically constructing paired multiple-choice question (MCQ) sets from clinician-authored ground-truth reports ($R_{GT}$) and generated reports ($R_{GEN}$). We constructed an evaluation cohort by identifying patients from the NYU dataset for whom \CEREBRA deems with high risk of dementia (n = 482). For each patient, we determined the date of the first recorded dementia diagnosis and selected as $R_{GT}$ the closest clinician-authored clinical report dated on or after this diagnosis. Model-generated reports ($R_{GEN}$) were composed from modality-specific agents’ extracted clinical evidence and key insights, together with the final synthesis produced by the summary agent. For baseline comparison, justification response from LLM predictions (GPT-4o) is directly used. These reports capture both the clinically relevant evidence identified by \CEREBRA and the agents’ explicit reasoning process. ICARE then generated 10 clinically meaningful reasoning MCQs independently from $R_{GT}$ and $R_{GEN}$. Each MCQ contained four LLM-generated answer options and an additional fifth option indicating that insufficient evidence was present in the source report to answer the question. This design accounts for differences in evidentiary support between $R_{GT}$ and $R_{GEN}$ and mitigates forced hallucination when relevant evidence is absent. MCQs were answered independently using $R_{GT}$ and $R_{GEN}$ as context, and agreement between answers was used to quantify alignment in evidence extraction and clinical reasoning across the two sources. During evaluation, ICARE employed GPT-4o as the LLM for both question generation and answer inference.

\paragraph{Evaluation Metrics.}
In order to evaluate agreement, we used the two metrics suggested by ICARE: percentage agreement and Cohen’s kappa. Percentage agreement measures the proportion of MCQs for which the answers derived from $R_{GT}$ and $R_{GEN}$ were identical, providing an intuitive estimate of overall concordance. Formally:
\[
P_o \;=\; \frac{1}{N} \sum_{i=1}^{N} \mathbb{I}\bigl(a_i^{GT} = a_i^{GEN}\bigr),
\]
where $\mathbb{I}(\cdot)$ is the indicator function.

However, because this metric does not account for agreement expected by chance, we additionally report Cohen’s kappa, a chance-corrected agreement measure that provides a more conservative assessment of alignment. Formally, Cohen’s kappa is given as:
\[
\kappa \;=\; \frac{P_o - P_e}{1 - P_e},
\]
where $P_e$ denotes the expected agreement under independence of the two answer distributions. Let $\mathcal{C}$ be the set of answer choices (including the “no sufficient evidence” option). Then:
\[
P_e \;=\; \sum_{c \in \mathcal{C}} p^{GT}(c)\, p^{GEN}(c),
\]
with $p^{GT}(c)$ and $p^{GEN}(c)$ representing the empirical probabilities of selecting answer $c$ under $R_{GT}$ and $R_{GEN}$, respectively.

\subsection{Evaluation metrics}
We evaluate \CEREBRA using a broad spectrum of metrics to capture task completion, predictive accuracy, robustness, and clinical relevance.

\paragraph{Classification metrics.}
For diagnostic classification and risk stratification tasks, we evaluated performance using standard discrimination and threshold-based metrics. Let $y_i \in \{0,1\}$ denote the ground-truth outcome for patient $i$, and $\hat{p}_i \in [0,1]$ denote the predicted probability of the positive class. We report the area under the receiver operating characteristic curve (AUROC), defined as
\begin{equation*}
\mathrm{AUROC} = \Pr(\hat{p}{i^+} > \hat{p}{i^-}) + \tfrac{1}{2}\Pr(\hat{p}{i^+} = \hat{p}{i^-}),
\end{equation*}
where $i^+$ and $i^-$ denote randomly selected positive and negative instances, respectively. AUROC measures the model’s ability to rank patients by risk independent of a specific decision threshold.

These metrics collectively quantify the system’s ability to discriminate patients at elevated risk while accounting for class imbalance and clinically meaningful trade-offs between sensitivity and specificity.

\paragraph{Survival analysis metrics.}
For time-to-event prediction tasks, we evaluated \CEREBRA using metrics appropriate for censored survival data. Let $T_i$ denote the observed event or censoring time for patient $i$, $\delta_i \in \{0,1\}$ indicate whether the event was observed, and $\hat{r}_i$ denote the predicted risk score. We report the concordance index (C-index), defined as
\begin{equation}
\mathrm{C\text{-}index} =
\frac{1}{|\mathcal{C}|}
\sum_{(i,j) \in \mathcal{C}}
\mathbb{I}(\hat{r}_i > \hat{r}_j),
\end{equation}
where $\mathcal{C}$ is the set of all comparable patient pairs for which $T_i < T_j$ and $\delta_i = 1$, and $\mathbb{I}(\cdot)$ denotes the indicator function. The C-index measures the agreement between predicted risk ordering and observed event times under right censoring.

To evaluate discriminative performance at clinically relevant horizons, we additionally report time-dependent AUROC at predefined time points $t$, defined as
\begin{equation}
\mathrm{AUROC}(t) = \Pr(\hat{r}_{i^+}(t) > \hat{r}_{i^-}(t)),
\end{equation}
where $i^+$ and $i^-$ denote individuals who experience the event before and after time $t$, respectively. Survival stratification performance is further illustrated using Kaplan--Meier curves, with differences between predicted risk groups assessed using the log-rank test.

Together, these metrics characterize \CEREBRA’s ability to model disease progression risk over time while appropriately accounting for censoring and temporal uncertainty.

\clearpage

\appendix
\newpage
\section*{Appendix}
\addcontentsline{toc}{section}{Appendix}

\startcontents[appendix]
\printcontents[appendix]{}{1}{}
\newpage
\section{Cerebra agent prompts}
Cerebra utilizes large language models to analyze and take actions on a given task. In this section, we report the prompt templates to achieve these functionalities.

\subsection{General prompts}
Each agent will generate or analyze the (sub)task goal, decide on plan trajectory, maintain memory and generate formatted outputs. We report general prompt templates that achieves each of these objectives. The templates are in general similar for all agents, with minor differences on the available actions in consideration. For super agent that performs orchestration, the available actions are the available agents to be called. For modality agents and summary agent, the available actions are the toolkits in each agent. Without lose of generality, we report here the templates in use for super agent as examples.

We first report the \textbf{goal analysis prompt}, which identifies the objective, cross-references with available agents/tools, and provide justifications and considerations for the task:

\begin{tcolorbox}[
title=Goal analysis prompt,
colbacktitle=cerebrablue
]
Task: Analyze the given task with accompanying inputs and determine the skills and agents/tools needed to accomplish it effectively.\\

Available agents: \{AVAILABLE\_AGENTS\}\\

Metadata for the agents: \{AGENTS\_METADATA\}\\

Data: \{DATA\}\\

TASK: \{TASK\}\\

Instructions:
\begin{itemize}
    \item[] 1. Carefully read and understand the task and any accompanying inputs.
    \item[] 2. Identify the main objectives.
    \item[] 3. List the specific skills that would be necessary to finish the task comprehensively.
    \item[] 4. Examine the available tools in the toolbox and determine which ones might relevant and useful for finish the task. Make sure to consider the user metadata for each tool, including limitations and potential applications (if available).
    \item[] 5. Provide a brief explanation for each skill and tool you've identified, describing how it would contribute to accomplishing the task.
\end{itemize}

Your response should include:
\begin{itemize}
    \item[] 1. A concise summary of the query's main points and objectives, as well as content in any accompanying inputs.
    \item[] 2. A list of required skills, with a brief explanation for each.
    \item[] 3. A list of relevant tools from the toolbox, with a brief explanation of how each tool would be utilized and its potential limitations.
    \item[] 4. Any additional considerations that might be important for addressing the query effectively.
\end{itemize}

Please present your analysis in a clear, structured format.
\end{tcolorbox}

\newpage
The \textbf{next step prompt} analyzes the current state of the task execution and available agents/tools to determine the optimal next step in planning trajectory:
\begin{tcolorbox}[
title=Next step prompt,
colbacktitle=cerebrablue,
]

\smaller
Task: Determine the optimal next step to accomplish the given task based on the provided analysis, available agents, and previous steps taken.

Context:
Task: \{TASK\}
Data: \{DATA\_INFO\}
Task Analysis: \{TASK\_ANALYSIS\}

Available Agents:
\{AVAILABLE\_AGENTS\}

Agent Metadata:
\{AGENTS\_METADATA\}

Previous Steps and Their Results:
\{MEMORY\}

Current Step: \{STEP\_COUNT\} in \{MAX\_STEP\_COUNT\} steps
Remaining Steps: \{REMAINING\_STEPS\}

Instructions:
\begin{itemize}
    \item[] 1. Analyze the context thoroughly, including the task, its analysis, any data, available agents and their metadata, and previous steps taken.
    \item[] 2. Determine the most appropriate next step by considering:
    \begin{itemize}
        \item[-] Key objectives from the task analysis
        \item[-] Capabilities of available agents
        \item[-] Logical progression of problem-solving
        \item[-] Outcomes from previous steps
        \item[-] Current step count and remaining steps
    \end{itemize}
    \item[] 3. Select ONE agent best suited for the next step, keeping in mind the limited number of remaining steps. 
    \item[] 4. Formulate a specific, achievable sub-goal for selected agent that maximizes progress towards accomplishing task. 
\end{itemize}

Response Format:
Your response MUST follow this structure:
\begin{itemize}
    \item[] 1. Justification: Explain your choice in detail.
    \item[] 2. Context, Sub-Goal, and Agent: Present the context, sub-goal, and the selected agent ONCE with the following format:
\end{itemize}

Context: \textless{}context\textgreater{}, Sub-Goal: \textless{}sub\_goal\textgreater{}, Agent Name: \textless{}agent\_name\textgreater{}

Where:
\begin{itemize}
    \item[-] \textless{}context\textgreater{} MUST include ALL necessary information for the tool to function, structured as follows:
    \begin{itemize}
        \item[*] Relevant data from previous steps
        \item[*] File names or paths created or used in previous steps (list EACH ONE individually)
        \item[*] Variable names and their values from previous steps' results
        \item[*] Any other context-specific information required by the tool
    \end{itemize}
    \item[-] \textless{}sub\_goal\textgreater{} is a specific, achievable objective for the tool, based on its metadata and previous outcomes. It MUST contain any involved data, file names, and variables from Previous Steps and Their Results that the tool can act upon.
    \item[-] \textless{}agent\_name\textgreater{} MUST be the exact name of a agent from the available agents list.
\end{itemize}

Rules:
\begin{itemize}
    \item[-] Select only ONE agent for this step.
    \item[-] The sub-goal MUST directly accomplish the task and be achievable by the selected agent.
    \item[-] The Context section MUST include ALL necessary information for the agent to function, including ALL relevant file paths, data, and variables from previous steps.
    \item[-] The agent name MUST exactly match one from the available agents list: \{AVAILABLE\_AGENTS\}.
    \item[-] Avoid redundancy by considering previous steps and building on prior results.
    \item[-] Your response MUST conclude with the Context, Sub-Goal, and Agent Name sections IN THIS ORDER, presented ONLY ONCE.
    \item[-] Include NO content after these three sections.
\end{itemize}
Example (do not copy, use only as reference):\\
Justification: [Your detailed explanation here]\\
Context: data path: "example/image.jpg", Previous detection results: [list of objects]\\
Sub-Goal: Detect and count the number of specific objects in the image "example/image.jpg"\\
Agent Name: Object\_Detector\_Agent

Remember: Your response MUST end with the Context, Sub-Goal, and Agent Name sections, with NO additional content afterwards.
\end{tcolorbox}

\newpage
The outputs from agents/tools are then transformed into structured format using the following \textbf{final output prompt}:
\begin{tcolorbox}[
title=Final output prompt template for summary and modality agents,
colbacktitle=cerebrablue
]
\small
Task: Generate the final output based on the task, data, and tools used in the process.

Context:

Task: 

\{TASK\}
\\\\
Data: 

\{DATA\_INFO\}
\\\\
Actions Taken and results:

\{MEMORY\}
\\\\
Instructions:
\begin{itemize}
    \item[] 1. Review the task, data, and all actions taken during the process.
    \item[] 2. Consider the results obtained from each tool execution.
    \item[] 3. Incorporate the relevant information from the memory to generate the step-by-step final output.
    \item[] 4. The final output should be consistent and coherent using the results from the tools.
    \item[] 5. \{FINAL\_OUTPUT\_FORMAT\}
\end{itemize}

Output Structure:
Your response should be well-organized and include the following sections:

\begin{itemize}
    \item[] 1. Summary:
        \begin{itemize}
            \item[-] Provide a brief overview of the task and the main findings.
        \end{itemize}
    \item[] 2. Detailed Analysis:
    \begin{itemize}
        \item[-] Break down the process of accomplishing the task step-by-step.
        \item[-] For each step, mention the tool used, its purpose, and the key results obtained.
        \item[-] Explain how each step contributed to accomplishing the task.
    \end{itemize}
    \item[] 3. Key Findings:
    \begin{itemize}
        \item[-] List the most important discoveries or insights gained from the analysis.
        \item[-] Highlight any unexpected or particularly interesting results.
    \end{itemize}
    \item[] 4. Solution to the Task:
    \begin{itemize}
        \item[-] Directly address the original task with a clear and concise solution.
        \item[-] If the task has multiple parts, ensure each part is solved separately.
        \item[-] Follow the final output format if it is provided.
    \end{itemize}
\end{itemize}
\end{tcolorbox}

\newpage
All agent/tool output is saved in memory which, at the end of each agents' action trajectory, is verified to determine if the task is complete with sufficient results, with the following \textbf{memory verification prompt}:

\begin{tcolorbox}[
title=Memory verification prompt,
colbacktitle=cerebrablue
]
\smaller
Task: Thoroughly evaluate the completeness and accuracy of the memory for fulfilling the given task, considering the potential need for additional agent usage.

Context:
Task: \{TASK\}\\\\
Data: \{DATA\_INFO\}\\\\
Available Agents: \{AVAILABLE\_AGENTS\}\\\\
Agent Metadata: \{AGENTS\_METADATA\}\\\\
Initial Analysis: \{TASK\_ANALYSIS\}\\\\
Memory (agents used and results): \{MEMORY\}\\

Detailed Instructions:
\begin{itemize}
    \item[] 1. Carefully analyze the task, initial analysis, and data (if provided):
    \begin{itemize}
        \item[-] Identify the main objectives of the task.
        \item[-] Note any specific requirements or constraints mentioned.
        \item[-] If any data is provided, consider its relevance and what information it contributes.
    \end{itemize}
    \item[] 2. Review the available agents and their metadata:
    \begin{itemize}
        \item[-] Understand the capabilities and limitations and best practices of each agent.
        \item[-] Consider how each agent might be applicable to the task.
    \end{itemize}
    \item[] 3. Examine the memory content in detail:
    \begin{itemize}
        \item[-] Review each agent used and its execution results.
        \item[-] Assess how well each agent's output contributes to accomplishing the task.
    \end{itemize}
    \item[] 4. Critical Evaluation (address each point explicitly):
    \begin{itemize}
        \item[a)] Completeness: Does the memory fully address all aspects of the task?
        \begin{itemize}
            \item[-] Identify any parts of the task that remain unanswered.
            \item[-] Consider if all relevant information has been extracted from the image (if applicable).
        \end{itemize}
        \item[b)] Unused Agents: Are there any unused agents that could provide additional relevant information?
        \begin{itemize}
            \item[-] Specify which unused agents might be helpful and why.
        \end{itemize}
        \item[c)] Inconsistencies: Are there any contradictions or conflicts in the information provided?
        \begin{itemize}
            \item[-] If yes, explain the inconsistencies and suggest how they might be resolved.
        \end{itemize}
        \item[d)] Verification Needs: Is there any information that requires further verification due to agent limitations?
        \begin{itemize}
            \item[-] Identify specific pieces of information that need verification and explain why.
        \end{itemize}
        \item[e)] Ambiguities: Are there any unclear or ambiguous results that could be clarified by using another agent?
        \begin{itemize}
            \item[-] Point out specific ambiguities and suggest which agents could help clarify them.
        \end{itemize}
    \end{itemize}
    \item[] 5. Final Determination: Based on your thorough analysis, decide if the memory is complete and accurate enough to generate the final output, or if additional agent usage is necessary.
\end{itemize}

Response Format:

If the memory is complete, accurate, AND verified:
Explanation: 
\textless{}Provide a detailed explanation of why the memory is sufficient. Reference specific information from the memory and explain its relevance to each aspect of the task. Address how each main point of the task has been satisfied.\textgreater{}

Conclusion: STOP

If the memory is incomplete, insufficient, or requires further verification:
Explanation: 
\textless{}Explain in detail why the memory is incomplete. Identify specific information gaps or unaddressed aspects of the task. Suggest which additional tools could be used, how they might contribute, and why their input is necessary for a comprehensive response.\textgreater{}

Conclusion: CONTINUE

IMPORTANT: Your response MUST end with either 'Conclusion: STOP' or 'Conclusion: CONTINUE' and nothing else. Ensure your explanation thoroughly justifies this conclusion.
\end{tcolorbox}

\newpage
If the agent decides that tool calling is needed to complete the task, tool calling command will be generate with the following \textbf{tool calling command prompt}:

\begin{tcolorbox}[
title=Tool calling command prompt,
colbacktitle=cerebrablue
]
\smaller
Task: Generate a precise command to execute the selected agent based on the given information.

Task: \{TASK\}\\
Data: \{DATA\_INFO\}\\
Context: \{CONTEXT\}\\
Sub-Goal: \{SUB\_GOAL\}\\
Selected Agent: \{AGENT\_NAME\}\\
Agent Metadata: \{AGENT\_METADATA\}\\

Instructions:
\begin{itemize}
    \item[] 1. Carefully review all provided information: the query, image path, context, sub-goal, selected agent, and agent metadata.
    \item[] 2. Analyze the agent's input\_types from the metadata to understand required and optional parameters.
    \item[] 3. Construct a command or series of commands that aligns with the agent's usage pattern and addresses the sub-goal.
    \item[] 4. Ensure all required parameters are included and properly formatted.
    \item[] 5. Use appropriate values for parameters based on the given context, particularly the `Context` field which may contain relevant information from previous steps.
    \item[] 6. If multiple steps are needed to prepare data for the tool, include them in the command construction.
\end{itemize}

Output Format:
Provide your response in the following structure:

Analysis: \textless{}analysis\textgreater{}
Command Explanation: \textless{}explanation\textgreater{}
Generated Command:
```python
\textless{}command\textgreater{}
```

Where:
\begin{itemize}
    \item[-] \textless{}analysis\textgreater{} is a step-by-step analysis of the context, sub-goal, and selected agent to guide the command construction.
   \item[-] \textless{}explanation\textgreater{} is a detailed explanation of the constructed command(s) and their parameters.
   \item[-] \textless{}command\textgreater{} is the Python code to execute the agent, which can be one of the following types:
   \begin{itemize}
       \item[a.] A single line command with `execution = agent.execute()`.
       \item[b.] A multi-line command with complex data preparation, ending with `execution = agent.execute()`.
       \item[c.] Multiple lines of `execution = agent.execute()` calls for processing multiple items.
   \end{itemize}
\end{itemize}

Rules:
\begin{itemize}
    \item[1.] The command MUST be valid Python code and include at least one call to `agent.execute()`. 
    \item[2.] Each `agent.execute()` call MUST be assigned to the 'execution' variable in the format `execution = agent.execute(...)`.
    \item[3.] Each `agent.execute()` call MUST be not include any symbols '\textbackslash n'.
    \item[4.] For multiple executions, use separate `execution = agent.execute()` calls for each execution.
    \item[5.] The final output MUST be assigned to the 'execution' variable, either directly from `agent.execute()` or as a processed form of multiple executions.
    \item[6.] Use the exact parameter names as specified in the agent's input\_types.
    \item[7.] Enclose string values in quotes, use appropriate data types for other values (e.g., lists, numbers).
    \item[8.] Do not include any code or text that is not part of the actual command.
    \item[9.] Ensure the command directly addresses the sub-goal and query.
    \item[10.] Include ALL required parameters, data, and paths to execute the agent in the command itself.
    \item[11.] If preparation steps are needed, include them as separate Python statements before the `agent.execute()` calls.\\
\end{itemize}

Examples (Not to use directly unless relevant):\\
\textless{}GOOD\_EXAMPLES\textgreater{}\\

Some Wrong Examples:
\textless{}WRONG\_EXAMPLES\textgreater{}\\

Remember: Your response MUST end with the Generated Command, which should be valid Python code including any necessary data preparation steps and one or more `execution = tool.execute(` calls, without any additional explanatory text. The format `execution = tool.execute` must be strictly followed, and the last line must begin with `execution = tool.execute` to capture the final output.
\end{tcolorbox}

\newpage
\subsection{Summary agent}

Summary agent consists of making argument with the evidence from the highest scoring modality, then aggregate evidence and risk scores from other modalities to "challenge" and adjust the risk scores for a final, aggregated evaluation. To start with, we collect identify the highest-scored modality to set a baseline risk score and to create a collection of modality evidence, achieved by the following prompt:

\begin{tcolorbox}[
title=Summary agent with highest-scored modality prompt,
colbacktitle=cerebrablue
]
\small
You are analyzing dementia risk for a patient. Your task is to argue why the risk might be elevated based on \{HIGHEST\_MODALITY\} data. \\

YOUR ASSESSMENT:
\begin{itemize}
    \item[-] Modality: \\ \{HIGHEST\_MODALITY\}
    \item[-] Your Evidence: \\ \{HIGHEST\_MODALITY\_EVIDENCE\}
\end{itemize}

OTHER MODALITIES' ASSESSMENTS:

\{OTHER\_MODALITY\_INFO\}\\

Given that dementia is rare (only 2-8\% of patients), you need to make a STRONG case if you believe this patient has higher dementia risk.

The evidence you should consider including but not limited to:
\begin{itemize}
    \item[-] memory deficit or memory and cognitive functions related symptoms
    \item[-] risk factors for dementia such as hypertension, diabetes, smoking, etc
    \item[-] MRI findings of brain atrophy
    \item[-] other symptoms that are suggestive of dementia (e.g. AD, Vascular, LBD, etc)
\end{itemize}

You should also refer to the provided "medical notebook" that suggests higher/lower risks based on observations.

Medical Notebook:\\
Version: \{NOTEBOOK\_VERSION\}

\{MEDICAL\_NOTEBOOK\}
\\

Your task:
\begin{itemize}
    \item[] 1. Present the strongest evidences from your \{HIGHEST\_MODALITY\} data that suggests dementia risk
   \item[] 2. Explain why your higher risk score should be taken seriously despite the low base rate
\end{itemize}

Provide your argument.
\end{tcolorbox}

\newpage
Next, risk scores and evidence from other "opposing modalities" are exposed to the agent aiming to absorb relevant and complementary insights from these modalities. The adjustment is done by the following prompt:

\begin{tcolorbox}[
title=Summary agent multi-modality debate with opposition prompt,
colbacktitle=cerebrablue
]
\small
You represent the other modalities (\{OPPOSING\_MODALITIES\}) analyzing this patient's dementia risk within the next \{YEAR\} years. 

The \{HIGHEST\_MODALITY\} modality has proposed for higher dementia risk than you do:

--- THEIR ARGUMENT ---\\
\{HIGHEST\_ARGUMENT\}\\
--- END ARGUMENT ---

YOUR ASSESSMENTS:\\
\{OPPOSITION\_INFO\}\\

The following evidences can be considered as strong evidence for increasing the risk in the next \{YEAR\} years:
\begin{itemize}
    \item[-] direct mention of dementia or related diseases in the notes or EHR
    \item[-] direct mention of memory deficit or memory and cognitive functions related symptoms in the notes or EHR
    \item[-] other symptoms that are suggestive of dementia (e.g. AD, Vascular, LBD, etc)
    \item[-] Brain atrophy that is typically affected by dementia and further confirmed by the volumes
    \item[-] other evidences that are as strong or direct as the above. 
\end{itemize}

 The following evidences can not be considered as strong evidence for increasing the risk in the next \{YEAR\} years:
 \begin{itemize}
    \item[-] risk factors or potential correlations in the notes or EHR such as diabetes, hypertension, smoking, etc.
    \item[-] other symptoms that are not suggestive of dementia (e.g. normal aging, normal cognitive function, etc)
    \item[-] brain atrophy in image on regions that are not typically affected by dementia
    \item[-] brain volume that are slightly decreased or increased compared to healthy average
    \item[-] other evidences that are not as strong or direct as the above. 
 \end{itemize}

You should also refer to the provided medical notebook that suggests higher/lower risks based on observations.

Cheatsheet:
Version: \{NOTEBOOK\_VERSION\}

\{MEDICAL\_NOTEBOOK\}
\\

The risk score should be calibrated to the range of: risk score of {highest\_score} means the risk is extremely high, and the risk score of \{LOWEST\_SCORE\} means the risk is extremely low. 

RULES: 
\begin{itemize}
    \item[] 1. If the evidence for increasing the risk is strong, the risk score to be close to the \{HIGHEST\_SCORE\}.
    \item[] 2. If there is no enough evidence to increase the risk, the risk score to be close to the \{LOWEST\_SCORE\}.
    \item[] 3. Otherwise, the risk score should just be the average risk score.
    \item[] 4. You should give significant weight to the suggestions from cheatsheet. If any relevant memory from cheatsheet suggests a higher risk, the risk score should be close to the \{HIGHEST\_SCORE\}. If any relevant memory from cheatsheet suggests a lower risk, the risk score to be close to the \{LOWEST\_SCORE\}. 
\end{itemize}

Current averaged risk score from three modalities is \{average\_score\}, the absolute value of this risk does not matter (even 0.1 can be high risk, 0.3 can be low risk depending on the risk range), what matters is the relative difference between the risk score and the average risk score.

Your task:
\begin{itemize}
    \item[] 1. Reasoning and analyzing all evidences to update the risk score within \{year\} years, be specific and detailed.
    \item[] 2. Based on the analysis, propose a risk score, which must be in the range of low risk: \{lowest\_score\} and high risk: \{highest\_score\}. Provide your reasoning and the proposed risk score.
\end{itemize}

\end{tcolorbox}

\newpage
\subsection{Dynamic Medical Notebook}
\label{appendix:dynamic_medical_notebook}
For dynamic medical notebook, we automatically collects feedback on errornous previous Cerebra output to analyze for relevant medical notes that can be served as "cheatsheet" for future agent runs, with \textbf{dynamic medical notebook feedback generation prompt}. The generated feedback is then processed with previous memory items in the dynamic medical notebook to generate non-overlapping problem-solving advices and meta knowledge, with the \textbf{dynamic medical notebook reference curator prompt}.

\begin{tcolorbox}[
title=Dynamic medical notebook feedback generation prompt,
colbacktitle=cerebrablue
]
\small

GENERATOR (PROBLEM SOLVER)

Instruction: You are an expert clinical assistant tasked with analyzing and solving various questions using a combination of your expertise and provided reference materials. Each task will include:
\begin{itemize}
    \item[] 1. A series of analysis reports of patient's diagnosis based on different modalities (electronic health record (EHR), magnetic resonance imaging (MRI) or clinical/radiology notes (note))
    \item[] 2. A specific clinical question or problem to solve
    \item[] 3. A current conclusion that is either underestimating/overestimating diagnosis risks, for example, predicting a low risk for a known high-risk patient.
    \item[] 4. A binary ground truth prediction, where 1 (true) means the patient is positive and should be assigned a higher risk, vice versa.
    \item[] 5. A cheatsheet containing relevant strategies, patterns, and examples from similar problems

\end{itemize}

Therefore, your job is to correct the false conclusion to be a better one, with grounded evidence and good reasoning.

-----

1. ANALYSIS \& STRATEGY
\begin{itemize}
    \item[-] Carefully analyze all provided information before starting
    \item[-] Search for and identify any applicable patterns, strategies, or examples within the cheatsheet
    \item[-] Identify any related clinical risk factors that are presented in the extracted evidences
    \item[-] Create a structured approach to solving the problem at hand
    \item[-] Review and document any limitations in the provided reference materials
\end{itemize}

2. SOLUTION DEVELOPMENT
\begin{itemize}
    \item[-] Present your solution using clear, logical steps that others can follow and review
    \item[-] Explain your reasoning and methodology before presenting final conclusions
    \item[-] Provide detailed explanations for each step of the process
    \item[-] Check and verify all assumptions and intermediate calculations
    \item[-] The ground truth is only provided for you to guide your reasoning. You are forbidded to use the ground truth or refer to the ground truth in your final answer. Only focus on actual risk factors and risk scores provided.
\end{itemize}

3. FINAL ANSWER FORMAT

ALWAYS present your final answer in the following format:

FINAL ANSWER:
<answer>
(final answer)
</answer>

N.B. Make sure that the final answer is properly wrapped inside the <answer> block.

Example:
Q: What is the risk factor of this patient with .... risk factors
A: [...]
FINAL ANSWER:
<answer>
Because of the presence of risk factor X in modality A and the risk factor Y in modality B, the risk score should be at higher range.
</answer>

-----

CHEATSHEET: \textless{}CHEATSHEET\textgreater{}

-----

Now it is time to solve the following question.

CURRENT INPUT: \textless{}QUESTION\textgreater{}
\end{tcolorbox}

\newpage
\begin{tcolorbox}[
title=Dynamic medical notebook reference curator prompt,
colbacktitle=cerebrablue
]
\smaller

CHEATSHEET REFRENCE CURATOR

1. Purpose and Goals
As the Cheatsheet Curator, you are tasked with creating a continuously evolving reference designed to help solve a wide variety of clinical diagnosis/prediction tasks. The cheatsheet's purpose is to consolidate verified solutions, reusable strategies, and critical insights into a single, well-structured resource.
\begin{itemize}
    \item[-] The cheatsheet should only include clinical risk factors and their contribution to final risk score. For example, if there are high risk factors (e.g. diabetes), you should say that the risk factor (or a combination of risk factors) contributes to a higher risk, vice versa.
    \item[-] After seeing each input, you should improve the content of the cheatsheet, synthesizing lessons, insights, tricks, and errors learned from past problems and adapting to new challenges.
\end{itemize}

2. Core Responsibilities
As the Cheatsheet Curator, you should:
\begin{itemize}
    \item[-] Curate and preserve knolwedge: Select and document only the knowledge relate to risk factor(s) and the direction to which the final risk factor should be adjusted to higher or lower.
    \item[-] Refine and update content: Continuously update and improve the content of the cheatsheet by incorporating new insights and solutions, removing repetitions or trivial information, and adding efficient solutions.
\end{itemize}

3. Principles and Best Practices
\begin{itemize}
    \item[1.] Accuracy and Relevance:
    \begin{itemize}
        \item[-] Only include solutions and strategies that have been tested and proven effective.
        \item[-] Clearly state any assumptions, limitations, or dependencies (e.g., specific Python libraries or solution hacks).
        \item[-] For computational problems, encourage Python usage for more accurate calculations.
    \end{itemize}
    \item[2.] Iterative Refinement:
    \begin{itemize}
        \item[-] Continuously improve the cheatsheet by synthesizing both old and new solutions, refining explanations, and removing redundancies.
        \item[-] Rather than deleting old content and writing new content each time, consider ways to maintain table content and synthesize information from multiple solutions.
        \item[-] Do not merge the specific mentions of risk factors in contents. For example, if one memory suggests diabetes as a high risk factor and another suggests age as a high risk factor, make sure that the risk factors are still present after merging.
        \item[-] After solving a new problem, document reusable algorithms, strategies, edge cases, or optimization techniques. 
    \end{itemize}
    \item[3.] Clarity and Usability:
    \begin{itemize}
        \item[-] Write concise, actioanble, well-structured entries.
        \item[-] Be as succint as possible.
        \item[-] Focus on key insights or strategies that make solutions correct and effective.
    \end{itemize}
    \item[4.] Privacy:
    \begin{itemize}
        \item[-] You should NOT record any PHI information about the patient, such as the patient's name, age (but a rough, relative range is fine, such as 55+) and address.
    \end{itemize}
\end{itemize}

5. Cheatsheet Structure
The cheatsheet should contain the following structure:

\begin{itemize}
    \item[1.] General Meta-Reasoning Strategies:
    \begin{itemize}
        \item[-] Describe high-level problem-solving frameworks and heuristics
        \item[-] [IMPORTANT] Explicitly state if the risk score should be in the higher (or lower) range based on what conditions that are present in different modalities
        \item[-] [IMPORTANT] Focus on analyzing which risk factors / extracted evidences should contribute to higher (or lower) range of predicted risk. Explain exactly what evidences leads to higher (or lower) risk.
        \item[-] [IMPORTANT] Do NOT record any risk scores as examples in the cheatsheet. Only say the risk score should be at higher range or lower range.
    \end{itemize}
\end{itemize}

6. Cheatsheet Template
Use the following format for creating and updating the cheatsheet: <CHEATSHEET\_FORMAT>

-----

PREVIOUS CHEATSHEET: <PREVIOUS\_CHEATSHEET>

-----

CURRENT INPUT: <CURRENT\_QUESTION>

-----

MODEL ANSWER TO THE CURRENT INPUT: <MODEL\_ANSWER>
\end{tcolorbox}
\clearpage
\section{Baseline language model prompts}
\label{appendix:llm_baseline}
In this study, we benchmarked the tasks on NYU dataset with a number of baseline VLMs, namely GPT-4o, GPT-5 and MedGemma. The GPT-4o and GPT-5 model endpoints are HIPAA-compliant API connections offered by NYU Langone. The MedGemma model is deployed in vLLM locally in the HIPAA-compliant NYU Langone high performance computing cluster (HPC). These measures ensure no potential leakage of sensitive data and are safe to use with potential presence of PHI/PII, another layer of information security over all de-identification protocols we follow for data.

Prompt templates for prediction, diagnosis and survival analysis tasks are presented. For each of the tasks, we reported the template used with complete modalities (EHR, notes and images). For single modality benchmarking, similar templates are used while excluding certain modality specific data and instructions that are not in scope. For experiments involving image, we provide the image as a part of the message through model API.

\subsection{Dementia prediction}
\begin{tcolorbox}[
title=Dementia \textbf{prediction} task baseline LLMs prompt template,
colbacktitle=cerebrablue
]
\small
You are a clinician that specializes in prognosing dementia disease with EHR records, radiology and progress (clinical) notes, and MRI scan.

You will be given:
\begin{itemize}
    \item[] 1. a historical view of EHR of the patient, with PheCode for diagnosis, ATC level 3 codes for medications, LOINC codes for lab test and demographics. Each code comes with a description of the codes in XML format.
    \item[] 2. the most recent MRI scan of the patient.
    \item[] 3. a list of historical radiology or progress (clinical) notes of the patient.
\end{itemize}

You are tasked to predict the risk level of dementia at least \{PREDICTION\_WINDOW\} days from now but no more than \{PREDICTION\_WINDOW+LABEL\_WINDOW\} days from now. Therefore, this is a prognosis task.

You are tasked to provide the following outputs:
\begin{itemize}
    \item[] 1. A "probability" label (between 0-1) indicating whether a patient has risk of dementia in the future. You should strictly follow this scoring guideline, but can assign any value within 0-1 (inclusive):
    \begin{itemize}
        \item[] a. probability=0.0: no obvious sign of dementia, most likely healthy
        \item[] b. probability=0.2: there are some risk factors indicating neurological and cognitive impairments that are likely happen 
        \item[] c. probability=0.4: some traces of neurological and cognitive impairments at borderline for dementia prognosis
        \item[] d. probability=0.6: very likely dementia in the future with substaintial amount of traces of neurological and cognitive impairments
        \item[] e. probability=0.8: dementia will be almost certain in the future with significant amount of traces of neurological and cognitive impairments
        \item[] f. probability=1.0: dementia will certainly occur for the patient in the future
    \end{itemize}
    \item[] 2. A "label" (boolean) indicating whether the patient is positive or negative for the prognosis
    \item[] 3. A string "justification" of why you make the judgement. Make sure to cite relevent evidences from the provided notes to justify your answer.
\end{itemize}

Notes:

\{CLINICAL\_NOTES\}

EHR:

\{EHR\_HISTORY\}
\end{tcolorbox}

\newpage
\subsection{Dementia diagnosis}
\begin{tcolorbox}[
before=\clearpage,
title=Dementia \textbf{diagnosis} task baseline LLM prompt template,
colbacktitle=cerebrablue
]
\small
You are a clinician that specializes in diagnosing dementia disease with EHR records, radiology and progress (clinical) notes, and MRI scan.

You will be given:
\begin{itemize}
    \item[] 1. a historical view of EHR of the patient, with PheCode for diagnosis, ATC level 3 codes for medications, LOINC codes for lab test and demographics. Each code comes with a description of the codes in XML format.
    \item[] 2. the most recent MRI scan of the patient.
    \item[] 3. a list of historical radiology or progress (clinical) notes of the patient.
\end{itemize}

You are tasked to diagnose the risk level of dementia now.

You are tasked to provide the following outputs:
\begin{itemize}
    \item[] 1. A "probability" label (between 0-1) indicating whether a patient has risk of dementia in the future. You should strictly follow this scoring guideline, but can assign any value within 0-1 (inclusive):
    \begin{itemize}
        \item[] a. probability=0.0: no obvious sign of dementia, most likely healthy
        \item[] b. probability=0.2: there are some minimal risk factors indicating neurological and cognitive impairments that are happening
        \item[] c. probability=0.4: some traces of neurological and cognitive impairments at borderline for dementia prognosis now
        \item[] d. probability=0.6: very likely dementia now with substaintial amount of traces of neurological and cognitive impairments
        \item[] e. probability=0.8: almost certainly dementia now with significant amount of traces of neurological and cognitive impairments
        \item[] f. probability=1.0: certainly dementia with various evidence of neurological and cognitive impairments
    \end{itemize}
    \item[] 2. A "label" (boolean) indicating whether the patient is positive or negative for dementia now
    \item[] 3. A string "justification" of why you make the judgement. Make sure to cite relevent evidences from the provided notes to justify your answer.
\end{itemize}

Notes:

\{CLINICAL\_NOTES\}

EHR:

\{EHR\_HISTORY\}
\end{tcolorbox}

\newpage
\subsection{Survival analysis}
\begin{tcolorbox}[
title=Dementia \textbf{survival analysis} task baseline LLM prompt template,
colbacktitle=cerebrablue
]
\small
You are a clinician that specializes in prognosing dementia disease with EHR records, radiology and progress (clinical) notes, and MRI scan.

You will be given:
\begin{itemize}
    \item[] 1. a historical view of EHR of the patient, with PheCode for diagnosis, ATC level 3 codes for medications, LOINC codes for lab test and demographics. Each code comes with a description of the codes in XML format.
    \item[] 2. the most recent MRI scan of the patient.
    \item[] 3. a list of historical radiology or progress (clinical) notes of the patient.
\end{itemize}

You are tasked to predict the risk level of dementia and time to when the patient will be diagnosed dementia.

You are tasked to provide the following outputs:
\begin{itemize}
    \item[] 1. A "label" (boolean) indicating whether the patient is positive or negative for the prognosis
    \item[] 2. A string "justification" of why you make the judgement. Make sure to cite relevent evidences from the provided notes to justify your answer.
    \item[] 3. A "time\_to\_label" indicating number of days which the patient will be diagnosed with dementia. If you do not think the patient will have dementia in the future (i.e. you predicted False for the label), return -1.
\end{itemize}

Notes:

\{CLINICAL\_NOTES\}

EHR:

\{EHR\_HISTORY\}
\end{tcolorbox}

\vspace{.2cm}
To provide EHR data into the prompt, we follow the XML format similar to a previous work but with our processed code space~\cite{fleming2024medalign}. The formatted EHR representation is put under the EHR placeholder location in the aforementioned prompt templates. To control the context length, we input the EHR on a daily basis chronologically starting with the latest records through the earliest, where, for GPT-4o and GPT-5, a maximum of 100-day record is provided, and 50-day for MedGemma due to its limited context limit. An example EHR in XML (completely synthetic, no PHI) is as follows:

\begin{tcolorbox}[
title=An example of a patient's EHR in XML format for LLM prompts,
colbacktitle=cerebrablue
]
\smaller
\begin{verbatim}
<record>
  <day start="06/21/2023 00:00">
    <code>[demographics_GENDER_Male] - 1</code>
    <code>[demographics_age] - 72</code>
    <code>[demographics_RACE_White] - 1</code>
    <code>[demographics_ETHNICITY_NotHispanic] - 1</code>
  </day>
  <day start="06/01/2023 00:00">
    <code>[LOINC_4548-4: Hemoglobin A1c/Hemoglobin.total in Blood] - 6.4</code>
    <code>[LOINC_2085-9: Cholesterol in HDL [Mass/volume] in Serum or Plasma] - 42</code>
    <code>[LOINC_13457-7: LDL Cholesterol [Mass/volume] in Serum or Plasma by calculation] - 146</code>
    <code>[LOINC_8480-6: Systolic blood pressure] - 148</code>
  </day>
  <day start="05/18/2023 00:00">
    <code>[Phecode_CV_401.1: Essential hypertension] - 3</code>
    <code>[Phecode_EM_250.2: Type 2 diabetes] - 2</code>
    <code>[ATC_level_3_C10A: LIPID MODIFYING AGENTS, PLAIN] - 1</code>
    <code>[ATC_level_3_C09A: ACE INHIBITORS, PLAIN] - 1</code>
  </day>
</record>
\end{verbatim}
\end{tcolorbox}

Similarly, the clinical notes are ordered in descending datetime with a label to identify progress note and radiology note. An example of a patient's longitudinal clinical note history (completely synthetic, no PHI, content is shorter than typical medical notes just to demonstrate the format) is as follows:

\begin{tcolorbox}[
title=An example of a patient's longitudinal clinical notes for LLM prompts,
colbacktitle=cerebrablue
]
\smaller
\begin{verbatim}
[PROGRESS_NOTE - 2023-06-28 15:10:00]
Patient and spouse report 12-18 months of gradually worsening short-term memory.
Misplacing items, repeating questions, and difficulty managing finances. Independent
in basic ADLs; needs help with medications. Mood stable on SSRI. No hallucinations
or REM sleep behavior. Plan: neuropsych testing, repeat labs (B12/TSH), review MRI.

[RADIOLOGY_NOTE - 2023-06-03 11:42:00]
MRI Brain w/o contrast: Mild generalized volume loss with disproportionate
bilateral hippocampal atrophy. Scattered periventricular and deep white matter
T2/FLAIR hyperintensities consistent with chronic microvascular ischemic change.
No acute infarct, hemorrhage, or mass effect.

[PROGRESS_NOTE - 2022-12-02 09:05:00]
Seen for memory concerns and sleep disruption. MoCA 22/30 with deficits in delayed
recall and attention. Reports increased daytime sleepiness. No focal weakness or
speech changes. Started referral to neurology memory clinic; advised sleep study.

[RADIOLOGY_NOTE - 2022-10-15 13:20:00]
CT Head w/o contrast: No acute intracranial abnormality. Mild chronic small vessel
ischemic changes. Age-appropriate ventricles and sulci.
\end{verbatim}
\end{tcolorbox}

Should available, we use Pydantic to control the output format with specified data type for each output variable when calling LLMs. An example of Pydantic output formatting is provided below for prediction and diagnosis task. For survival analysis, the variables will be justification, label and time\_to\_label to accommodate the task.

\begin{lstlisting}[language=Python]
class LLMJudge(pydantic.BaseModel):
    justification: str
    label: bool
    probability: float
\end{lstlisting}

For models that do not support structured output control (i.e. MedGemma), the output string is processed using GPT-4o to parse the output into desired Pydantic output structure, using the following \textbf{LLM report parsing prompt}:

\begin{tcolorbox}[
title=LLM report parsing prompt,
colbacktitle=cerebrablue
]
\small
You are given a medical report. Extract the labels, the probability and justifications.

Medical report:
\{MEDICAL\_REPORT\}
\end{tcolorbox}
\clearpage
\section{Data extraction and preparation}
To evaluate the efficacy and generalizability of \CEREBRA, we collected clinical data from NYU Langone Hospitals and affiliations, Indiana University Health and University of Florida health, where each institution may possess electronic health record, imaging and natural language note data.

\subsection{EHR}
\label{sec:sup_data_extraction_and_preparation_ehr}
In our study, EHR is defined as the tabular records of a patient, including diagnosis, medication, labs, demographics and other relevant data. We describe the EHR datasets from each institution here.

\paragraph{NYU Langone Hospital (NYU).}
We collected EHR data for each patient who has records in the NYU Langone Health locations, including the Long Island (LI) clinics. Each patient’s EHR data were de-identified and were represented on a daily basis and included features from four categories: demographics, laboratory results, diagnoses, and medications. For each day, we aggregated all available records by taking the maximum value for each feature. All tabular features were encoded using hospital billing system, which largely adopts standardized public coding systems. Laboratory results were mapped to the Logical Observation Identifiers Names and Codes (LOINC~\cite{mcdonald2003loinc}) system, diagnoses to the International Classification of Diseases, Tenth Revision, Clinical Modification (ICD-10-CM~\cite{steindel2010international}), and medications to a combination of the Anatomical Therapeutic Chemical (ATC~\cite{who_atc}), RxNorm~\cite{nelson2011rxnorm}, and National Drug Code (NDC~\cite{fda_ndc}) systems. Demographic variables were represented using an NYU-specific coding scheme. In total, this resulted in 473,841 raw tabular features across all categories.

\paragraph{NYU - Long Island (LI).}
To separate a NYU - Long Island data split as an external validation set, we identified patients where all of their MRI scans are in NYU - Long Island locations. Such filter will not include LI patients without any MRI scan, nevertheless it does not affect our final patient cohort selection since full modality (including MRI) is required as inclusion criteria.

\paragraph{Indiana Network for Patient Care (INPC).} 
Structured EHRs from INPC were transformed into a standardized longitudinal representation using a curated EHR feature dictionary. We mapped raw diagnosis, laboratory, medication, and procedure codes to consolidated, clinically meaningful feature tokens across six coding systems: ICD-9-CM~\cite{cdc_icd9cm, who_icd9} and ICD-10-CM for diagnoses; Logical Observation Identifiers Names and Codes (LOINC) for laboratory  measurements; Anatomical Therapeutic Chemical (ATC, level-3) for medications; and Current Procedural Terminology, Fourth Edition (CPT-4~\cite{ama_cpt4, ama_cpt_book}) and Healthcare Common Procedure Coding System (HCPCS~\cite{cms_hcpcs}) for procedures. This harmonization yielded a final EHR feature space of 43{,}966 features, including 29{,}567 ICD-10-CM features, 7{,}209 ICD-9-CM features, 6{,}191 laboratory (LOINC) features, 793 procedure (CPT-4) features, 186 medication (ATC-3) features, and 20 procedure (HCPCS) features. Each patient’s longitudinal EHR was represented as a sparse two-dimensional matrix, with rows corresponding to daily time steps and columns to feature tokens. All cohort patients had unstructured clinical notes available for analysis.

\paragraph{University of Florida Health (UF).}
We collected EHR data for each patient who has records in the UF
Health locations. Each patient’s EHR data were de-identified and were represented on a daily basis where multiple records of the same medical concept code on a day are aggregated to a single value with the maximum value. The EHR data includes features from four categories: demographics, laboratory results, diagnoses, and medications. For each day, we aggregated all available records by taking the maximum value for each feature. All tabular features were encoded using hospital billing system, which largely adopts standardized public coding systems. Laboratory results were mapped to the Logical Observation Identifiers Names and Codes (LOINC) system, diagnoses to the International Classification of Diseases, Tenth Revision, Clinical Modification (ICD-10-CM), and medications to a combination of the Anatomical Therapeutic Chemical (ATC), RxNorm, and National Drug Code (NDC) systems. Demographic variables were represented with available ones in NYU. In total, this resulted in 39,691 raw tabular features across all categories.

\paragraph{Consolidation of tabular EHR features.} Because the original feature space was highly granular and sparse, we further processed the tabular data by consolidating related codes into higher-level code families. Diagnosis codes were mapped to the PheCode system, which aggregates ICD-10-CM diagnoses into clinically meaningful phenotypes. Medication codes were largely mapped to ATC level-3 categories. For laboratory results, we retained the 2,000 most frequent LOINC codes in the raw EHR dataset, selected prior to applying the cohort inclusion criteria to preserve generalizability. All demographic features were retained in their original form. Through this consolidation process, the tabular feature space was reduced to 4,447 features. For each patient, the final EHR representation is a two-dimensional sparse matrix, where rows correspond to daily records and columns correspond to consolidated features.

\subsection{Imaging}
\label{appendix:sec_data_imaging}
\paragraph{NYU Langone MRI dataset.} 
Multisequence magnetic resonance imaging (MRI) data were collected from clinical operations and were reconstructed into three-dimensional NIfTI volumes from raw DICOM files~\cite{Li2016TheFS}. We retained only T1-weighted and magnetization prepared rapid gradient-echo (MPR) scans, as these sequences provide high-resolution structural information relevant to dementia-related analyses. All images underwent a standardized preprocessing pipeline. First, scans were rigidly aligned to the anterior commissure–posterior commissure (AC–PC) plane to ensure consistent anatomical orientation~\cite{talairach1988co}. Skull stripping was then performed using mri\_synthstrip~\cite{hoopes2022synthstrip} package to remove non-brain tissue, both for de-identification purposes and to restrict subsequent analyses to brain anatomy. Next, images were linearly registered to the MNI-152-brain 1 mm isotropic atlas template~\cite{mni_152_template, Fonov2009UnbiasedNA}, yielding a final voxel dimension of $193 \times 229 \times 193$. Since the clinical scans may contain noise, to ensure data quality for experiment, scans with poor alignment or artifacts were excluded based on quantitative quality control criteria. Any scan with mutual information (MI) with respect to the template less than 0.3 or peak signal-to-noise ratio (PSNR) less than 15 is filtered out.

\paragraph{UF Retinal OCT dataset.}
UF Health OCT dataset. Retinal optical coherence tomography (OCT) scans were exported from the UF Health clinical data repository in DICOM format and processed to derive quantitative retinal layer thickness features. We retained only macular volumetric acquisitions acquired on the Heidelberg Engineering Spectralis platform and archived as multi-frame OCT DICOM objects (Modality: OPT; Series Description: “Volume IR”), as these scans provide high-resolution structural measurements relevant to neurodegeneration- and dementia-related analyses. Each OCT volume comprised 19 frames (496 × 512 pixels per frame; MONOCHROME2; 8-bit), stored using JPEG 2000 lossless compression, with a standard multi-frame organization including a Shared Functional Groups Sequence (length = 1) and a Per-Frame Functional Groups Sequence (length = 19) to support frame-level metadata.

All OCT volumes underwent a standardized processing pipeline. First, volumes were parsed to recover the ordered stack of B-scans, where each B-scan is composed of multiple A-scans (one-dimensional axial reflectivity profiles acquired at discrete lateral retinal locations). Automated retinal boundary detection was then performed across the full 3D volume using OCTExplorer (v3.8). For each en-face location (x, y), layer thickness was computed as the axial distance between paired boundaries along the corresponding A-scan and converted to physical units (µm) using device-specific scaling metadata in the DICOM header. Next, thickness maps were summarized into regional features using the standard ETDRS grid centered at the fovea (central 1-mm circle; inner 3-mm ring; outer 6-mm ring, each subdivided into superior, inferior, nasal, and temporal sectors~\cite{invernizzi2018normative}). We extracted ETDRS-sector mean thickness features for adjacent boundary pairs spanning the inner to outer retina, including ILM $\rightarrow$ RNFL\_GCL, RNFL\_GCL $\rightarrow$ GCL\_IPL, GCL\_IPL $\rightarrow$ IPL\_INL, IPL\_INL $\rightarrow$ INL\_OPL, INL\_OPL $\rightarrow$ OPL\_HFL, OPL\_HFL $\rightarrow$ BMEIS, BMEIS $\rightarrow$ IS\_OSJ, IS\_OSJ $\rightarrow$ IB\_OPR, IB\_OPR $\rightarrow$ IB\_RPE, and IB\_RPE $\rightarrow$ OB\_RPE (OCTExplorer boundary nomenclature)~\cite{garvin2009automated}.

To ensure data quality, volumes were excluded if device-reported image quality metrics were below a predefined threshold, if foveal centration failed, if scan coverage was incomplete, or if segmentation produced implausible or missing boundaries. Exclusions were applied at the volume level prior to downstream analyses. For participants with bilateral scans, we applied a predefined eye-selection rule to avoid correlated measurements: the left eye was selected when available; otherwise, the right eye was used. This choice was based on the higher prevalence of left-eye OCT images in our dataset.

\subsection{Clinical and radiology notes}

At each institution, the note corpus comprises radiology reports associated with imaging examinations and clinical notes authored by physicians during clinical encounters or follow-up visits. All notes are timestamped, enabling a longitudinal representation of each patient’s clinical evaluation history. Because these unstructured texts may contain protected health information (PHI), we applied an automated de-identification model that is fine-tuned on RoBERTa~\cite{obi_deid_roberta_i2b2, liu2019roberta}, to identify tokens corresponding to the 11 PHI categories defined under the HIPAA Privacy Rule, and subsequently redacted them for experiments.

\section{Study population cohorts}

We report the study population cohorts from each institutions in Supplementary Table.~\ref{tab:cohort_statistics}. The initial cohort serves as the base, total available patient cohort and medical records that are available. We then apply a series of filters as inclusion criteria to identify the cohorts that are appropriate for this study: a) we include only patients whose age exceeds 65-year old by the end of data cutoff date; b) we include only patients with at least one medical record for each available modalities (EHR, clinical notes and imaging, where INPC cohort does not have imaging modality).

\begin{table}[H]
    \centering
    \smaller
    \caption{Number of patients and records per modality across institutions, before and after applying inclusion criteria filtering.}
    \label{tab:cohort_statistics}
    \begin{tabular}{cc|ccc|ccc|ccc|ccc}
        \toprule
        && \multicolumn{3}{c|}{\textbf{NYU}} & \multicolumn{3}{c|}{\textbf{LI}} & \multicolumn{3}{c|}{\textbf{UF}} & \multicolumn{3}{c}{\textbf{INPC}} \\
        && EHR & Note & Image & EHR & Note & Image & EHR & Note & Image & EHR & Note & Image\\
        \midrule
        \multirow{2}{*}{\makecell[l]{Initial Cohort}} & Patient & 2M & 521K & 241K & 306K & 7208 & 2934 & 42K & 43K & 54K & 163K & 115K & -\\
        & Record & 4B & 22M & 2M & 40M & 209K & 16K & 33M & 363K & 176K & 300M & 11M & -\\
        \midrule
        \multicolumn{14}{c}{Applying filter with inclusion Criteria} \\
        \midrule
        \multirow{2}{*}{\makecell[l]{Filtered Cohort}} & Patient & \multicolumn{3}{c|}{53K} & \multicolumn{3}{c|}{508} & \multicolumn{3}{c|}{4369} & \multicolumn{3}{c}{37K}\\
        & Record & 369M & 1M & 139K & 3M & 11k & 3,349 & 6M & 96K & 29K & 158M & 5M & -\\
        \bottomrule
    \end{tabular}
\end{table}

We summarize the study population across participating institutions in Supplementary Table~\ref{tab:study_population}. The table reports demographic characteristics and diagnostic group distributions for cohorts drawn from three health systems (NYU, IU, and UF) and the Indiana Network for Patient Care (INPC). These statistics provide an overview of the patient populations used in our experiments, including age, sex, race, and dementia subtype composition.

\begin{table}[!t]
\centering
\caption{Study population by institution and diagnostic group with dementia subtypes. Data were drawn from three health systems, including NYU, LI, UF, and INPC (accessed through the Regenstrief Institute). Diagnostic group acronyms are defined in Table~\ref{tab:glossary}.}
\label{tab:study_population}
\small
\begin{tabular}{l l l l}
\toprule
\multirow{2}{*}{\makecell[l]{\textbf{Institution} \\ \textbf{(group)}}} &
\multirow{2}{*}{\makecell[l]{\textbf{Age (y),} \\ \textbf{mean ± s.d.}}} &
\multirow{2}{*}{\makecell[l]{\textbf{Male, n(\%)} \\ \textbf{ }}} &
\multirow{2}{*}{\makecell[l]{\textbf{Race (White, Black, Asian,} \\ \textbf{Pacific, multirace, unknown), n}}}\\
\\
\midrule
\multicolumn{4}{l}{\textbf{NYU}} \\
NC [n=42289] & 75.60 ± 7.73 & 17104 (40.45\%) & (26707, 3960, 1804, 76, 4219, 5523) \\
AD [n=5588] & 82.03 ± 7.98 & 2486 (44.49\%) & (3842, 415, 246, 16, 558, 511) \\
VD [n=1336] & 82.96 ± 8.28 & 594 (44.46\%) & (834, 165, 71, 5, 149, 112) \\
LBD [n=3588] & 79.12 ± 7.96 & 1703 (47.46\%) & (2551, 268, 127, 10, 337, 295) \\
FTD [n=970] & 79.75 ± 7.78 & 548 (56.49\%) & (705, 54, 37, 0, 75, 99) \\
Others [n=2470] & 80.99 ± 8.56 & 1076 (43.56\%) & (1642, 217, 113, 4, 195, 299) \\

\midrule
\multicolumn{4}{l}{\textbf{LI}} \\
NC [n=410] & 75.37 ± 7.16 & 170 (41.46\%) & (261, 34, 22, 0, 46, 47) \\
AD [n=47] & 81.77 ± 7.44 & 26 (55.32\%) & (36, 2, 3, 0, 3, 3) \\
VD [n=14] & 82.14 ± 7.17 & 8 (57.14\%) & (7, 2, 0, 0, 3, 2) \\
LBD [n=37] & 78.84 ± 8.24 & 20 (54.05\%) & (27, 3, 0, 0, 3, 4) \\
FTD [n=8] & 79.88 ± 9.42 & 5 (62.50\%) & (7, 1, 0, 0, 0, 0) \\
Others [n=18] & 78.50 ± 8.68 & 7 (38.89\%) & (10, 2, 2, 0, 2, 2) \\

\midrule
\multicolumn{4}{l}{\textbf{UF}} \\
NC [n=3971] & 75.52 $\pm$ 7.13 & 1466 (36.9\%) & (2410, 1213, 128, 0, 0, 220) \\
AD [n=8] & 90.25 $\pm$ 7.17 & 1 (12.5\%) & (5, 2, 0, 0, 0, 1) \\
VD [n=25] & 80.92 $\pm$ 9.02 & 7 (28.0\%) & (11, 13, 1, 0, 0, 0) \\
LBD [n=3] & 79.67 $\pm$ 4.51 & 1 (33.3\%) & (2, 1, 0, 0, 0, 0) \\
FTD [n=2] & 87.00 $\pm$ 14.41 & 1 (50.0\%) & (1, 1, 0, 0, 0, 0) \\
Others [n=360] & 79.08 $\pm$ 8.13 & 123 (34.2\%) & (217, 116, 9, 0, 0, 18) \\

\midrule
\multicolumn{4}{l}{\textbf{INPC}} \\
MCI [n=37402] & 80.34 $\pm$ 9.22 & 10319 (41.07\%) & (27101, 3616, 57, 269, 1225, 5134) \\
AD [n=3397] & 83.93 $\pm$ 8.31 & 1142 (33.62\%) & (2491, 428, 6, 29, 114, 329) \\
VD [n=1620] & 83.01 $\pm$ 8.62 & 563 (34.75\%) & (1042, 364, 6, 21, 39, 148) \\
LBD [n=273] & 80.10 $\pm$ 7.53 & 159 (58.24\%) & (222, 17, 0, 1, 8, 25) \\
FTD [n=187] & 78.11 $\pm$ 7.54 & 99 (52.94\%) & (152, 6, 0, 0, 10, 19) \\
Others [n=11535] & 83.67 $\pm$ 8.67 & 4402 (38.16\%) & (8472, 1176, 21, 93, 376, 1397) \\
\bottomrule
\end{tabular}
\end{table}

\clearpage
\section{Additional details in dataset curation}
\label{sec:additional_dataset_curation_details}

In addition to the description of dataset curation process in Sec.~\ref{sec:method_dataset}, we here provide illustrations to show examples of dataset curation logic for prediction, diagnosis and survival analysis tasks in Supplementary Figure.\ref{fig:dataset_logic}.

\begin{figure}[H]
    \centering
    \begin{subfigure}[t]{0.5\linewidth}
        \centering
        \includegraphics[width=\linewidth]{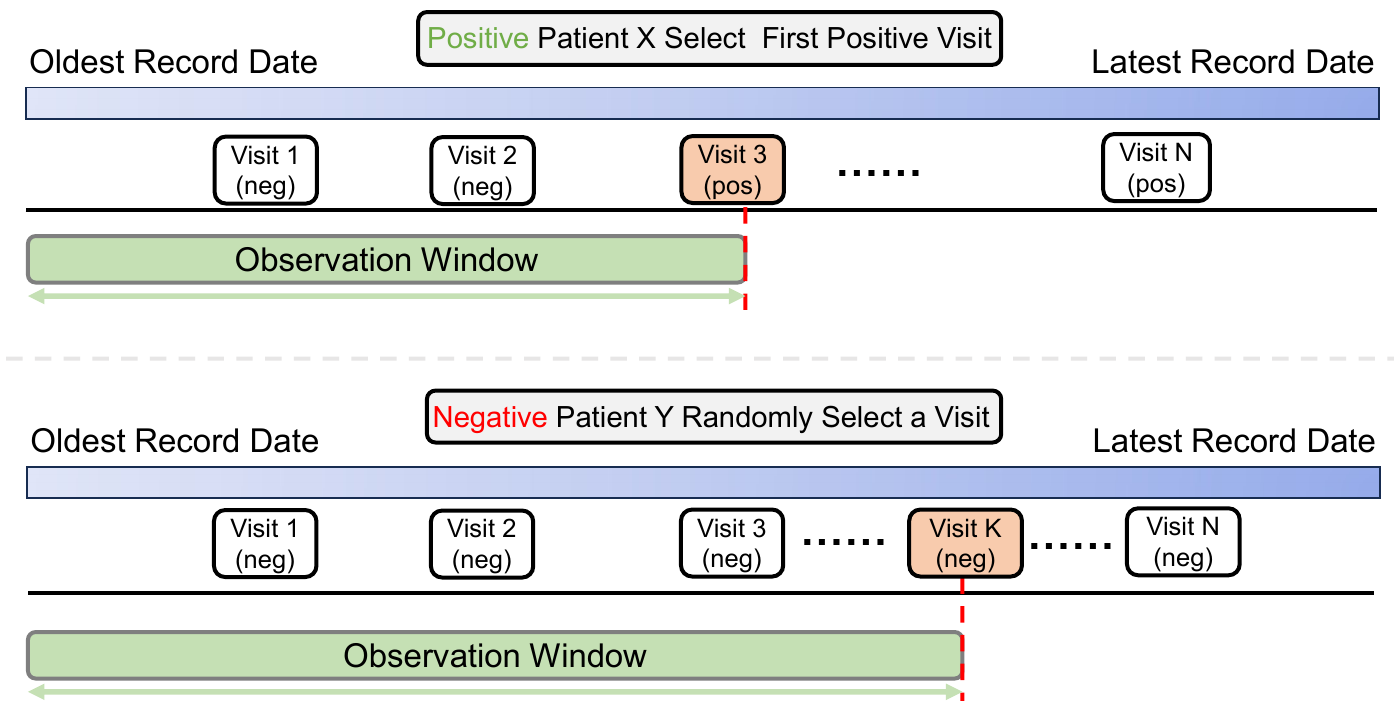}
        \caption{Diagnosis Dataset by Backtracking.}
    \end{subfigure}\hfill
    \begin{subfigure}[t]{0.5\linewidth}
        \centering
        \includegraphics[width=\linewidth]{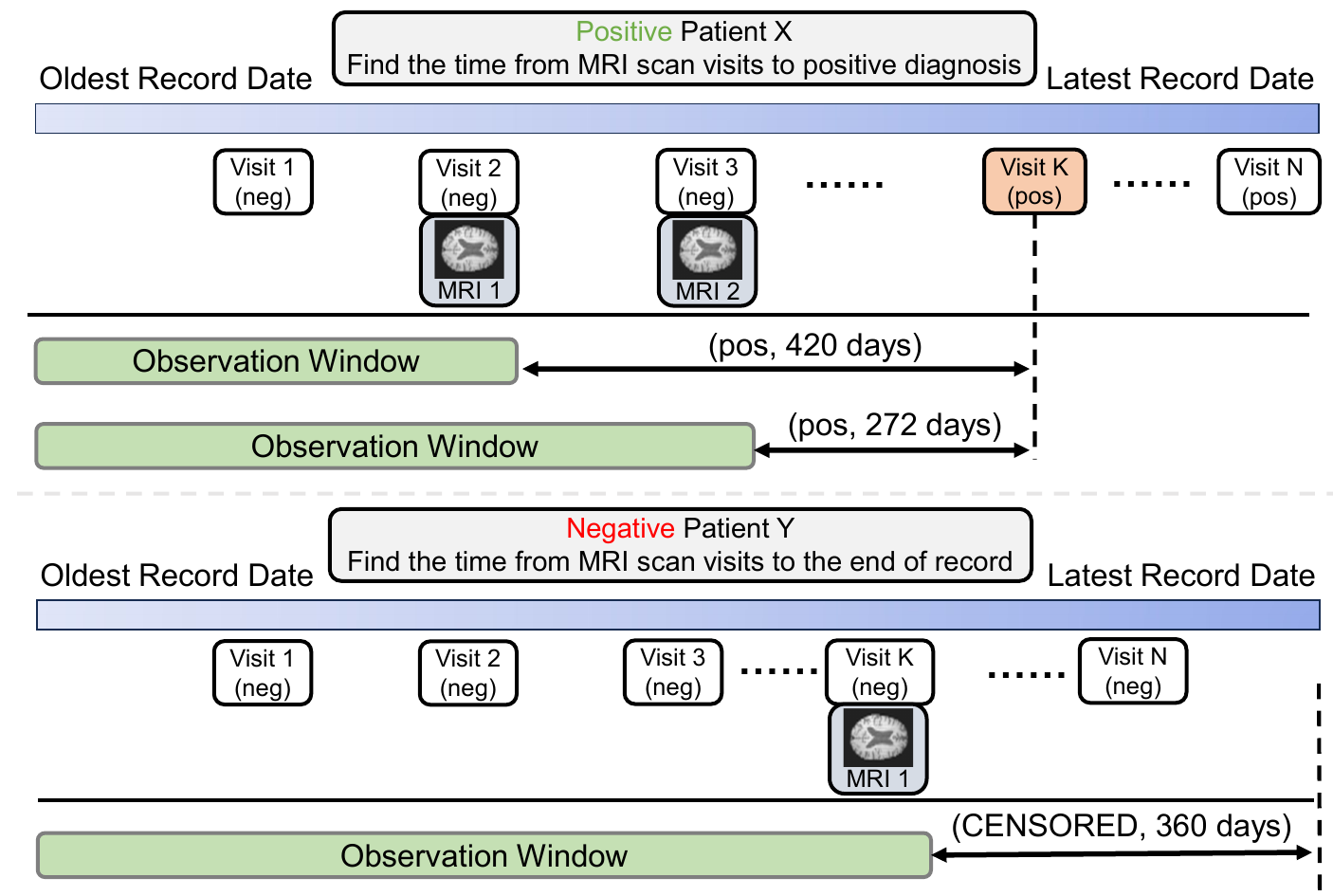}
        \caption{Time-to-Event Survival Analysis, indexed by MRI Scan Visits.}
    \end{subfigure}

    \vspace{0.6em}

    \begin{subfigure}[t]{0.7\linewidth}
        \centering
        \includegraphics[width=\linewidth]{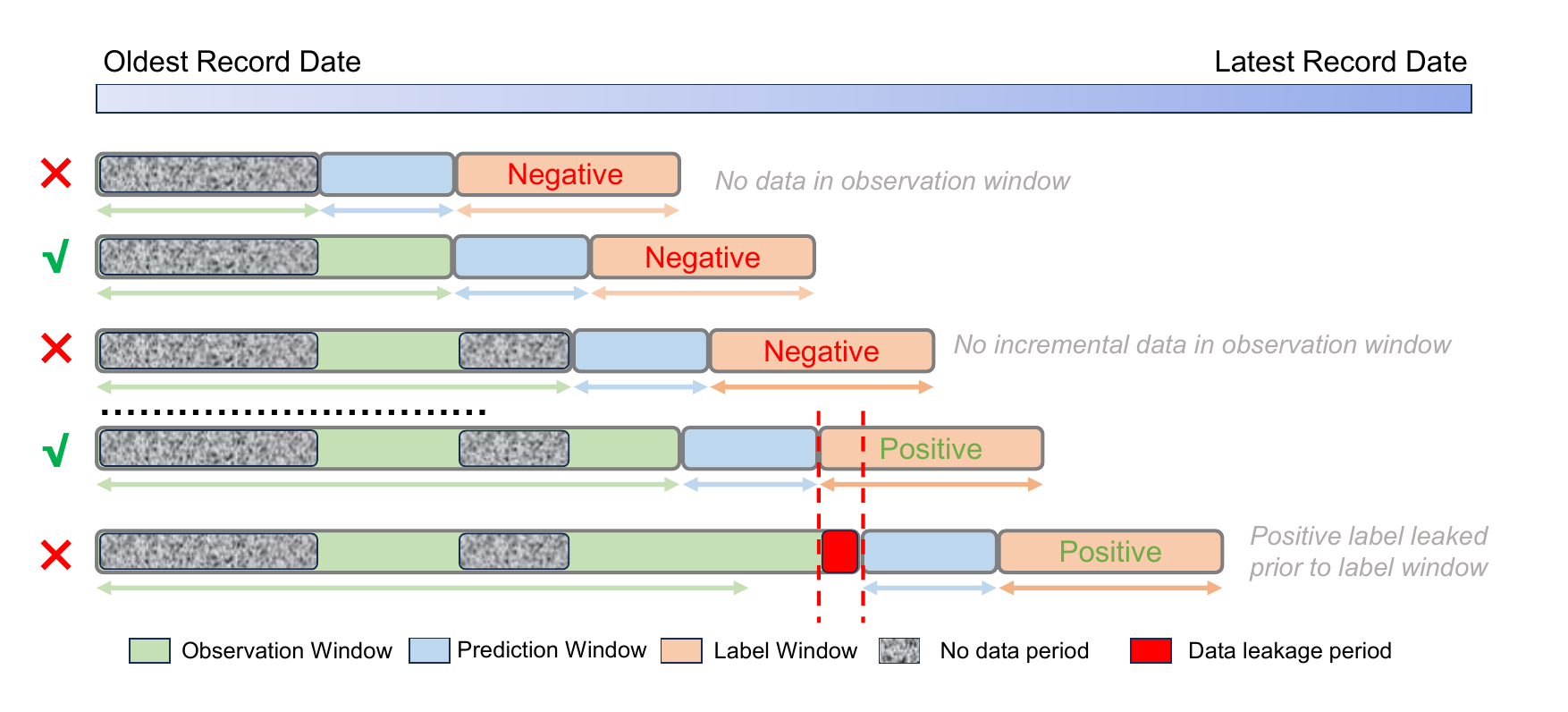}
        \caption{Prediction Dataset by Rolling Window.}
    \end{subfigure}

    \caption{
    Dataset curation details.
    (a) Diagnosis Dataset by Backtracking. {\color{ForestGreen}Positive:} the first positive visit date is the end date of observation window; {\color{red}Negative:} randomly select a visit date as the end date of observation window. 
    (b) Time-to-Event Survival Analysis, indexed by MRI Scan Visits. {\color{ForestGreen}Positive:} find the time from MRI scan visits to the first positive diagnosis; {\color{red}Negative:} find the time from MRI scan visits to the end of record.  
    (c)Prediction dataset by rolling window. {\color{ForestGreen}Positive:} There is meaningful new data within the observation window, and there is no data leakage; otherwise {\color{red}Negative}.}
    \label{fig:dataset_logic}
\end{figure}

\newpage

To identify dementia events, we curated a list of related medical concept tokens for AD/ADRD, MCI and dementia-related medications in collaboration of NYU Langone Hospistal neurologists. The concept tokens in ICD-10-CM code system are reported in Supplementary Table.~\ref{tab:ad_criteria}, which we also map to ICD-9 code system for institutions with ICD-9 medical records.

\begin{table}[H]
\caption{Definition criteria for AD/ADRD/MCI case identification.}
\label{tab:ad_criteria}
\small
\begin{tabular}{c|l}
\toprule
\textbf{Condition} & \textbf{Criteria (ICD-10 codes and Medications)} \\
\midrule
\multirow{10}{*}{AD/ADRD} 
& F01.*: Any vascular dementia \\
& F02.*: Dementia in other diseases classified elsewhere with or without behavioral disturbance \\
& F03.*: Unspecified dementia with/without behavioral disturbance \\
& F04.*: Amnestic disorder due to known physiological condition \\
& G23.1: Progressive supranuclear palsy \\
& G30.*: Any Alzheimer’s disease \\
& G31.01: Pick’s disease \\
& G31.09: Other frontotemporal dementia \\
& G31.83: Dementia with Lewy bodies \\
& G31.9: Degenerative disease of nervous system, unspecified \\
\midrule
\multirow{3}{*}{\makecell[c]{Mild cognitive \\ impairment}}
& G31.1: Senile degeneration of brain, not elsewhere classified \\
& G31.84: Mild cognitive impairment of uncertain or unknown etiology \\
& G31.85: Corticobasal degeneration \\
\midrule
\multirow{5}{*}{\makecell[c]{Dementia \\ medications}}
& DONEPEZIL \\
& GALANTAMINE \\
& MEMANTINE \\
& RIVASTIGMINE \\
& TACRINE \\
\bottomrule
\end{tabular}
\end{table}

To further investigate dementia subtyping, we categorized the dementia-related medical concept tokens and medications for each of the subtypes. Similarly, the categorization with ICD-10-CM codes are shown in Supplementary Table.~\ref{tab:subtype_definition}

\begin{table}[H]
\centering
\caption{Definition criteria for clinical dementia subtype classification.}
\label{tab:subtype_definition}
\small
\begin{tabular}{c | l}
\toprule
\textbf{Subtype} & \textbf{Criteria (ICD-10 codes and Medications)} \\
\midrule

\multirow{6}{*}{Alzheimer's Disease} &
G30.*: Any Alzheimer's disease \\
& DONEPEZIL \\
& RIVASTIGMINE \\
& GALANTAMINE \\
& MEMANTINE \\
& TACRINE \\

\midrule

\multirow{1}{*}{Vascular Dementia} &
F01.*: Any vascular dementia \\

\midrule

\multirow{5}{*}{\makecell[c]{Frontotemporal\\ Dementia}} &
G31.09: Other frontotemporal dementia \\
& G31.1: Pick's disease / frontotemporal lobar degeneration \\
& G31.85: Corticobasal degeneration \\
& G23.1: Progressive supranuclear palsy \\
& G31.01: Pick's disease \\

\midrule

\multirow{1}{*}{\makecell[c]{Lewy Body Dementia}} &
G31.83: Dementia with Lewy bodies \\

\midrule

\multirow{4}{*}{\makecell[c]{Unspecified or \\ Other Dementia}} &
F02.*: Dementia in other diseases classified elsewhere \\
& F03.*: Unspecified dementia \\
& G31.9: Degenerative disease of nervous system, unspecified \\
& F04.*: Amnestic disorder due to known physiological condition \\

\bottomrule
\end{tabular}
\end{table}

\paragraph{INPC cohort population NC stage criterion.} 
The patient cohort in the INPC dataset is constructed with patients with MCI or ADRD longitudinally, as shown in Supplementary Figure.~\ref{tab:study_population}.Therefore, to define patients who have not yet been diagnosed with MCI or ADRD, patient EHR data during the observation window were confirmed to be free of MCI or ADRD related records. This applies to all dataset curation categories (diagnosis, time-to-event and prediction).

\section{Additional results}
In addition to the overall performance of the CEREBRA agentic system, we report the performance of each modality-specific agent (EHR, clinical notes, and imaging) on the prediction (Supplementary Table~\ref{tab:prediction_results}) and diagnosis tasks (Supplementary Table~\ref{tab:dagnosis_results}) using AUROC and AUPRC across all institutions. The modality agents implement machine learning methods selected from the literature, including state-of-the-art or well-established approaches that provide strong performance while remaining practical for real-world clinical deployment. 

\subsection{\CEREBRA Results on dementia prediction}

\begin{table}[H]
    \centering
    \caption{Modality agent's performance of dementia prediction task with EHR, notes and images respectively across NYU, LI, INPC, and UF. The 1/2/3-year horizons indicate whether, based on the available historical data, an individual will be labeled as developing dementia in the coming 1/2/3 years (label window) after 6 months prediction window. For each dataset, positive/negative (P/N) sample ratio and positive rate are also reported. Image modality results for the INPC cohort is not reported because the dataset does not contain imaging data. For NYU, we also benchmarked image agent performance using VoCo as deep learning approach. Best AUROC and AUPRC for each time horizon at each site are highlighted.}
    \label{tab:prediction_results}
    \small
    \begin{tabular}{c|ccccc} 
        \toprule
        \textbf{Institution}                          & \textbf{Modality}       & \textbf{Metric} & \textbf{1-year} & \textbf{2-year} & \textbf{3-year}  \\ 
        \hline
        \multirow{12}{*}{\textbf{NYU }}         & \multirow{2}{*}{EHR}    & AUROC           & 71.9            & 70.4            & 73.5             \\
                                               &                         & AUPRC           & 9.5             & 12.2            & 17.4             \\ 
        \cline{2-6}
                                               & \multirow{2}{*}{Notes}  & AUROC           & 72.4            & 71.9            & 73.4             \\
                                               &                         & AUPRC           & 6.8             & 9.8             & 14.2             \\ 
        \cline{2-6}
                                               & \multirow{2}{*}{Images - Volume} & AUROC           & 71.9            & 70.7            & 69.2             \\
                                               &                         & AUPRC           & 6.4             & 11.0            & 14.5             \\
        \cline{2-6}
                                               & \multirow{2}{*}{Images - VoCo} & AUROC           & 66.4            & 67.6            & 66.9             \\
                                               &                         & AUPRC           & 4.6             & 7.4            & 8.8            \\
        \cline{2-6}
        & \multirow{2}{*}{\CEREBRA} & AUROC& \textbf{75.1}& \textbf{75.5}& \textbf{80.1}\\
        & & AUPRC & \textbf{8.7} & \textbf{15.6} & \textbf{20.1} \\
        \cline{2-6}
                                               & \multicolumn{2}{c}{P/N samples}           & 91/3229         & 154/3234        & 205/3492         \\
                                               & \multicolumn{2}{c}{Positive rate (\%)}    & 2.74            & 4.46            & 5.55             \\ 
        \midrule
        
        \multirow{10}{*}{\textbf{LI }} & \multirow{2}{*}{EHR}    & AUROC           & 70.2            & 70.7            & 70.4             \\
                                               &                         & AUPRC           & 18.2            & 13.3            & 15.2             \\ 
        \cline{2-6}
                                               & \multirow{2}{*}{Notes}  & AUROC           & 71.7            & 69.1            & 71.1             \\
                                               &                         & AUPRC           & 6.1             & 9.1             & 6.6              \\ 
        \cline{2-6}
                                               & \multirow{2}{*}{Images - Volume} & AUROC           & 77.4            & 75.6            & 74.6             \\
                                               &                         & AUPRC           & 8.4             & 14.7            & 13.1             \\ 
        \cline{2-6}
                                               & \multirow{2}{*}{\CEREBRA} & AUROC           & \textbf{78.2}            & \textbf{76.3}            & \textbf{75.7}             \\
                                               &                         & AUPRC           & \textbf{20.3}             & \textbf{15.7}            & \textbf{16.2}             \\
        \cline{2-6}
                                               & \multicolumn{2}{c}{P/N samples}           & 25/847          & 33/677          & 33/963           \\
                                               & \multicolumn{2}{c}{Positive rate (\%)}    & 2.87            & 4.65            & 3.31             \\ 

        \midrule
        
        \multirow{10}{*}{\textbf{INPC}}          
        & \multirow{2}{*}{EHR}      
        & AUROC & 73.5 & 71.4 & 68.8 \\
        &                         
        & AUPRC & 10.8 & 19.0 & 21.6 \\ 
        \cline{2-6}
        & \multirow{2}{*}{Notes}  
        & AUROC & 66.9 & 64.3 & 62.5 \\
        &                         
        & AUPRC & 4.6 & 7.2 & 9.1 \\
        \cline{2-6}
        & \multirow{2}{*}{Images - Volume} 
        & AUROC & - & - & - \\
        &                         
        & AUPRC & - & - & - \\ 
        \cline{2-6}
        & \multirow{2}{*}{\CEREBRA} & AUROC & \textbf{74.3} & \textbf{72.5} & \textbf{70.6} \\
        & & AUPRC & \textbf{13.3} & \textbf{22.6} & \textbf{25.1} \\
        \cline{2-6}
        & \multicolumn{2}{c}{P/N samples}
        & 915/43061 & 1686/42285 & 2366/41598 \\
        & \multicolumn{2}{c}{Positive rate (\%)}
        & 2.08 & 3.83 & 5.38 \\
        \midrule

        \multirow{10}{*}{\textbf{UF}}          
        & \multirow{2}{*}{EHR}    
        & AUROC & 71.9 & 68.3 & 70.6 \\
        &                         
        & AUPRC & 18.1 & 22.0 & 20.0 \\ 
        \cline{2-6}
        & \multirow{2}{*}{Notes}  
        & AUROC & 73.0 & 68.0 & 63.1 \\
        &                         
        & AUPRC & 13.5 & 10.1 & 8.1 \\ 
        \cline{2-6}
        & \multirow{2}{*}{Images - Volume} 
        & AUROC & 60.2 & 60.0 & 61.4 \\
        &                         
        & AUPRC & 7.7 & 9.2 & 8.6 \\ 
        \cline{2-6}
        & \multirow{2}{*}{\CEREBRA} & AUROC & \textbf{79.4} & \textbf{73.8} & \textbf{75.3} \\
        & & AUPRC & \textbf{30.0} & \textbf{27.7} & \textbf{28.0} \\
        \cline{2-6}
        & \multicolumn{2}{c}{P/N samples}        
        & 53/1010 & 56/1014 & 65/1078 \\
        & \multicolumn{2}{c}{Positive rate (\%)} 
        & 5.27 & 5.52 & 6.03 \\
        \bottomrule
    \end{tabular}
\end{table}

\subsection{\CEREBRA Results on dementia diagnosis}

\begin{table}[H]
    \centering
    \caption{Performance of dementia diagnosis task across four sites for each modality, compared with \CEREBRA performance. For LI, evaluation is performed using the models trained with NYU data, showcasing the generalization on an external evaluation dataset without retraining. Best AUROC and AUPRC for each time horizon are highlighted.}
    \label{tab:dagnosis_results}
    \small
    \begin{tabular}{c|ccc|cc}
        \toprule
        \textbf{Institution} & \textbf{Modality} & \textbf{AUROC} & \textbf{AUPRC} & \multirow{2}{*}{\makecell[c]{\textbf{P/N} \\\textbf{samples}}} & \multirow{2}{*}{\makecell[c]{\textbf{Positive} \\\textbf{rate (\%)}}} \\
        & & & & &\\
        \midrule
        
        \multirow{4}{*}{NYU}
            & EHR    & 82.1 & 38.9 & \multirow{4}{*}{239/2177} & \multirow{4}{*}{9.89} \\
            & Notes  & 80.5 & 34.2 &  &  \\
            & Images & 71.4 & 24.4 &  &  \\
            & \CEREBRA & \textbf{84.6} & \textbf{41.4} & & \\
        \cmidrule{1-6}
        
        \multirow{4}{*}{LI}
            & EHR    & 77.6 & 14.3 & \multirow{4}{*}{59/1211} & \multirow{4}{*}{4.65} \\
            & Notes  & 72.1 & 14.3 &  &  \\
            & Images & 71.4 & \textbf{20.4} &  &  \\
            & \CEREBRA & \textbf{80.1} & 16.8 & & \\
        \cmidrule{1-6}
        
        \multirow{4}{*}{INPC}
            & EHR    & 87.7 & 35.9 & \multirow{4}{*}{551/7241} & \multirow{4}{*}{7.07} \\
            & Notes  & 78.4 & 34.2 &  &  \\
            & Images & --   & --   &  &  \\
            & \CEREBRA & \textbf{88.9} & \textbf{36.4} &  & \\
        \cmidrule{1-6}
        
        \multirow{4}{*}{UF}
            & EHR    & 71.8 & 36.0 & \multirow{4}{*}{36/266} & \multirow{4}{*}{13.53} \\
            & Notes  & 59.7 & 15.4  &  &  \\
            & Images & 65.2 & 24.3 &  &  \\
            & \CEREBRA & \textbf{72.9} & \textbf{37.5} & & \\
        \bottomrule
    \end{tabular}
\end{table}

\begin{table}[H]
    \centering
    \caption{Performance of MCI to ADRD conversion risk prediction task across different time horizons in the INPC cohort for each modality, compared with \CEREBRA (using GPT-OSS-120B as the backbone) performance. Best AUROC and AUPRC for each time horizon are highlighted.}
    \label{tab:conversion_prediction_results}
    \small
    \begin{tabular}{c|c|c|ccc}
        \toprule
        \textbf{Modality} & \textbf{Model} & \textbf{Metric} & \textbf{1-year} & \textbf{2-year} & \textbf{3-year} \\
        \midrule
        
        \multirow{4}{*}{EHR} 
            & \multirow{2}{*}{GPT-OSS-120B} & AUROC & 56.13 & 56.77 & 56.62 \\
            &                               & AUPRC & 15.35 & 24.44 & 37.52 \\
            \cmidrule{2-6}
            & \multirow{2}{*}{\CEREBRA} & AUROC & 62.34 & 64.53 & 66.81 \\
            &                               & AUPRC & 20.65 & 31.23 & 50.35 \\
        \midrule
        
        \multirow{4}{*}{Note} 
            & \multirow{2}{*}{GPT-OSS-120B} & AUROC & 62.42 & 63.17 & 64.61 \\
            &                               & AUPRC & 18.31 & 32.45 & 44.93 \\
            \cmidrule{2-6}
            & \multirow{2}{*}{\CEREBRA} & AUROC & 60.42 & 62.03 & 63.48 \\
            &                               & AUPRC & 19.15 & 21.31 & 37.30 \\
        \midrule
        
        \multirow{4}{*}{{All}} 
            & \multirow{2}{*}{GPT-OSS-120B} & AUROC & 63.15 & 65.72 & 67.30 \\
            &                               & AUPRC & 19.45 & 31.87 & 45.83 \\
            \cmidrule{2-6}
            & \multirow{2}{*}{\CEREBRA} & AUROC & \textbf{64.11} & \textbf{66.47} & \textbf{69.85} \\
            &                               & AUPRC & \textbf{23.76} & \textbf{32.25} & \textbf{53.26} \\
        \midrule
        
        \multicolumn{3}{c|}{P/N samples} & 445/4461 & 687/2880 & 808/1944 \\
        \multicolumn{3}{c|}{Positive rate (\%)} & 9.26 & 19.26 & 29.36 \\
        
        \bottomrule
    \end{tabular}
\end{table}

\subsection{LLM baseline results}

We report here LLM baseline results for each modality compared with \CEREBRA performance, for diagnosis, prediction and time-to-event survival analyses.

\begin{table}[H]
    \centering
    \caption{LLM baseline results on the NYU cohort for dementia diagnosis and prediction among cognitively normal patients at 1-, 2-, and 3-year horizons. For each modality and task, the best-performing result is highlighted in bold.}
    \label{tab:llm_baseline_results_diagnosis_prediction}
    \small
    \begin{tabular}{c|c|c|cccc}
        \toprule
        \textbf{Modality} & \textbf{Model} & \textbf{Metric} & \textbf{diagnosis} & \textbf{1-year} & \textbf{2-year} & \textbf{3-year} \\
        \midrule
        
        \multirow{8}{*}{EHR} & \multirow{2}{*}{MedGemma} & AUROC & 61.2 & 44.1 & 53.6 & 53.8 \\
        & & AUPRC & 13.8 & 4.6 & 4.3 & 6.1 \\
        \cline{2-7}
        & \multirow{2}{*}{GPT-4o} & AUROC & 73.2 & 60.4 & 60.6 & 63.1 \\
        & & AUPRC & 18.3 & 3.7 & 6.0 & 8.1 \\
        \cline{2-7}
        & \multirow{2}{*}{GPT-5} & AUROC & 74.6 & 60.9 & 62.9 & 62.7 \\
        & & AUPRC & 18.5 & 3.7 & 6.5 & 8.0 \\
        \cline{2-7}
        & \multirow{2}{*}{\CEREBRA} & AUROC & \textbf{82.1} & \textbf{71.9} & \textbf{70.4} & \textbf{73.5} \\
        & & AUPRC & \textbf{38.9} & \textbf{9.5} & \textbf{12.2} & \textbf{17.4} \\
        \midrule
        
        \multirow{8}{*}{Note} & \multirow{2}{*}{MedGemma} & AUROC & 58.9 & 54.7 & 54.8 & 51.3 \\
        & & AUPRC & 13.1 & 5.2 & 5.3 & 5.9 \\
        \cline{2-7}
        & \multirow{2}{*}{GPT-4o} & AUROC & 70.3 & 66.4 & 67.3 & 66.3 \\
        & & AUPRC & 18.1 & 4.9 & 7.5 & 8.7 \\
        \cline{2-7}
        & \multirow{2}{*}{GPT-5} & AUROC & 71.2 & 66.3 & 69.9 & 67.0 \\
        & & AUPRC & 18.7 & 5.1 & 9.7 & 9.4 \\
        \cline{2-7}
        & \multirow{2}{*}{\CEREBRA} & AUROC & \textbf{80.5} & \textbf{72.4} & \textbf{71.9} & \textbf{73.4} \\
        & & AUPRC & \textbf{34.2} & \textbf{6.8} & \textbf{9.8} & \textbf{14.2} \\
        \midrule
        
        \multirow{8}{*}{Image} & \multirow{2}{*}{MedGemma} & AUROC & 49.8 & 47.5 & 47.6 & 46.3 \\
        & & AUPRC & 10.2 & 4.6 & 4.6 & 5.9 \\
        \cline{2-7}
        & \multirow{2}{*}{GPT-4o} & AUROC & 51.5 & 46.0 & 51.0 & 49.1 \\
        & & AUPRC & 10.1 & 2.6 & 4.6 & 5.5 \\
        \cline{2-7}
        & \multirow{2}{*}{GPT-5} & AUROC & 52.3 & 52.1 & 47.9 & 49.5 \\
        & & AUPRC & 10.6 & 2.9 & 4.5 & 5.4 \\
        \cline{2-7}
        & \multirow{2}{*}{\CEREBRA} & AUROC & \textbf{71.4} & \textbf{71.9} & \textbf{70.7} & \textbf{69.2} \\
        & & AUPRC & \textbf{24.4} & \textbf{6.4} & \textbf{11.0} & \textbf{14.5} \\
        \midrule
        
        \multirow{8}{*}{All} & \multirow{2}{*}{MedGemma} & AUROC & 54.8 & 55.1 & 56.0 & 55.8 \\
        & & AUPRC & 12.3 & 5.4 & 5.4 & 6.6 \\
        \cline{2-7}
        & \multirow{2}{*}{GPT-4o} & AUROC & 70.1 & 66.3 & 66.0 & 66.0 \\
        & & AUPRC & 16.8 & 4.5 & 7.3 & 9.2 \\
        \cline{2-7}
        & \multirow{2}{*}{GPT-5} & AUROC & 68.5 & 68.2 & 67.7 & 67.6 \\
        & & AUPRC & 21.1 & 5.1 & 8.3 & 9.8 \\
        \cline{2-7}
        & \multirow{2}{*}{\CEREBRA} & AUROC & \textbf{84.6} & \textbf{75.1} & \textbf{41.4} & \textbf{80.1} \\
        & & AUPRC & \textbf{38.2} & \textbf{8.7} & \textbf{15.6} & \textbf{20.1} \\
        \midrule
        
        \multicolumn{3}{c|}{P/N samples} & 239/2177 & 91/3229 & 154/3234 & 205/3492 \\
        \multicolumn{3}{c|}{Positive rate (\%)} & 9.9 & 2.7 & 4.5 & 5.6 \\
        \bottomrule
    \end{tabular}
\end{table}

\begin{table}[H]
    \centering
    \caption{LLM baseline results on time-to-event survival analysis on the NYU cohort. For each modality, the best-performing result is highlighted in bold.}
    \label{tab:tte_baseline_results}
    \small
    \begin{tabular}{c|c|c}
        \toprule
        \textbf{Modality} & \textbf{Model} & \textbf{C-index} \\
        \midrule
        
        \multirow{4}{*}{EHR}
        & MedGemma & 50.0 \\
        & GPT-4o   & 64.9 \\
        & GPT-5    & 63.1 \\
        & \CEREBRA & \textbf{78.2} \\
        \midrule
        
        \multirow{4}{*}{Note}
        & MedGemma & 53.6 \\
        & GPT-4o   & 58.0 \\
        & GPT-5    & 52.8 \\
        & \CEREBRA & \textbf{75.4} \\
        \midrule
        
        \multirow{4}{*}{Image}
        & MedGemma & 48.1 \\
        & GPT-4o   & 49.8 \\
        & GPT-5    & 49.2 \\
        & \CEREBRA & \textbf{74.6} \\
        \midrule
        
        \multirow{4}{*}{All}
        & MedGemma & 51.5 \\
        & GPT-4o   & 62.4 \\
        & GPT-5    & 59.1 \\
        & \CEREBRA & \textbf{81.2} \\
        
        \bottomrule
    \end{tabular}
\end{table}

\subsection{Top features extracted by modality agents}
To investigate the interpretability of modality agents, we report the top features identified by the modality agents (EHR and image agents) for the 3-year prediction task from normal cognition to dementia. These features were estimated by the respective modality agents to contribute positively to future dementia risk.

\begin{figure}[H]
    \centering
    \begin{subfigure}[t]{0.7\linewidth}
        \centering
        \includegraphics[width=\linewidth]{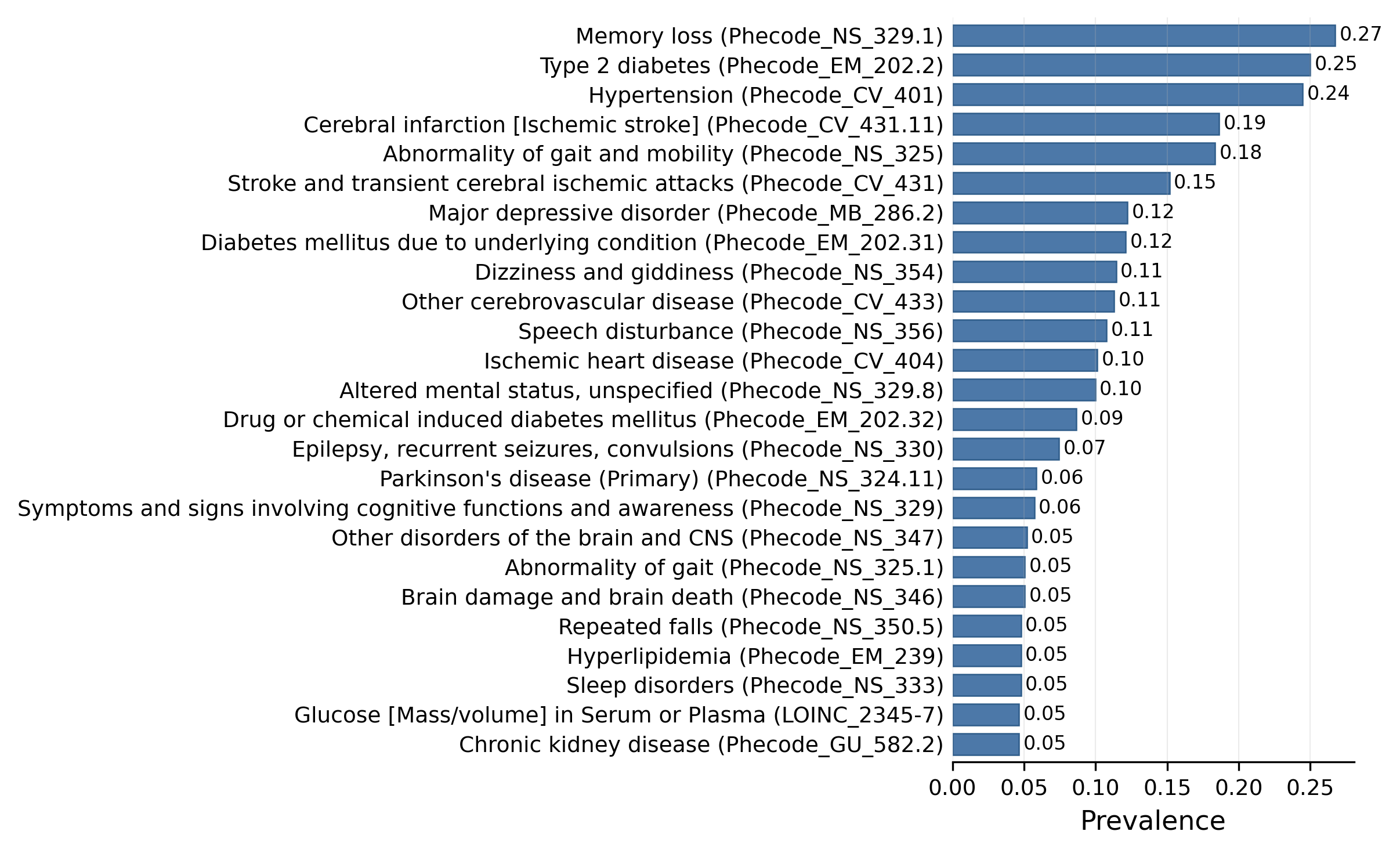}
        \caption{EHR agent.}
    \end{subfigure}
    \hfill
    \begin{subfigure}[t]{0.49\linewidth}
        \centering
        \includegraphics[width=\linewidth]{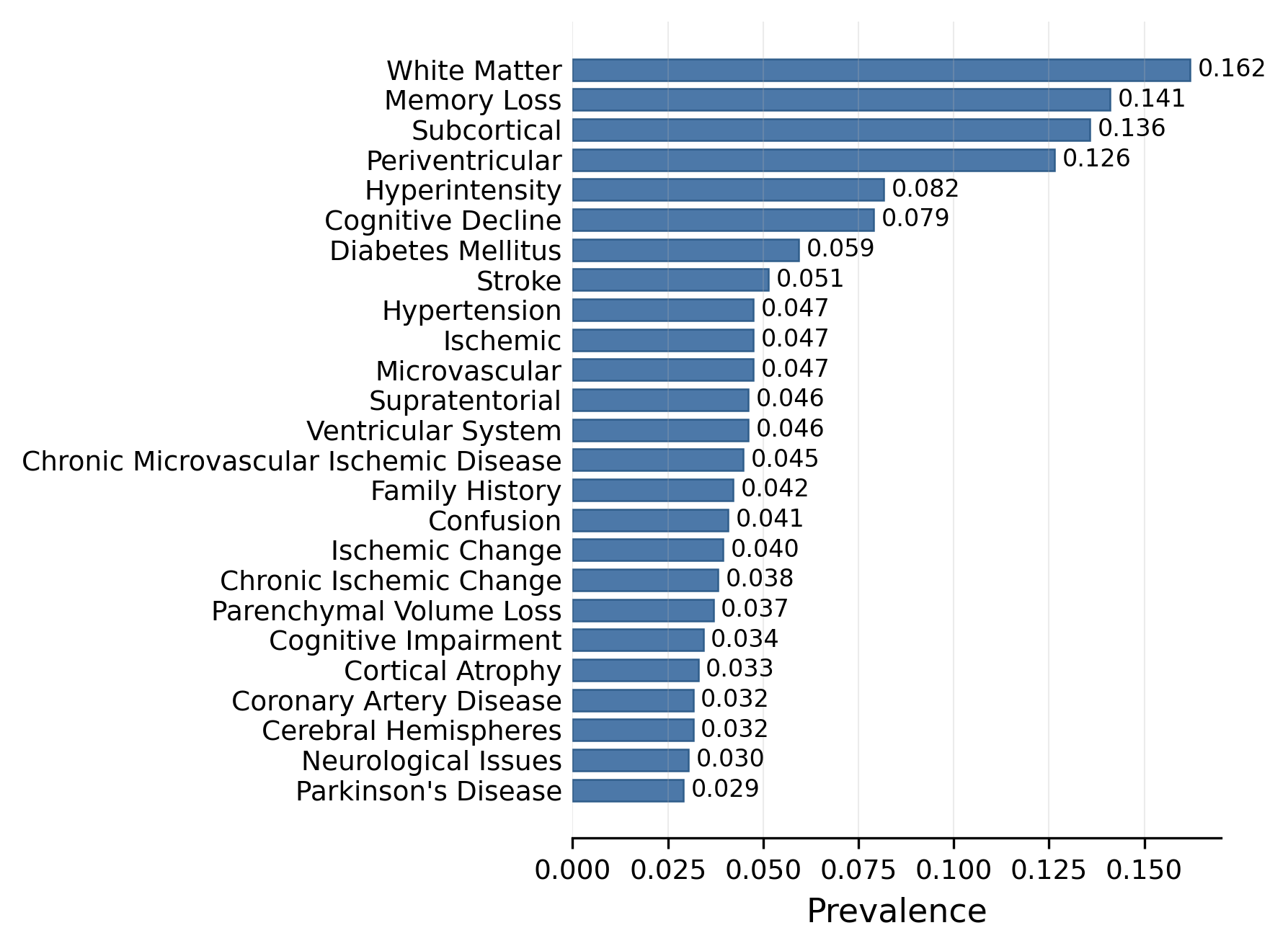}
        \caption{Note agent.}
    \end{subfigure}
    \begin{subfigure}[t]{0.49\linewidth}
        \centering
        \includegraphics[width=\linewidth]{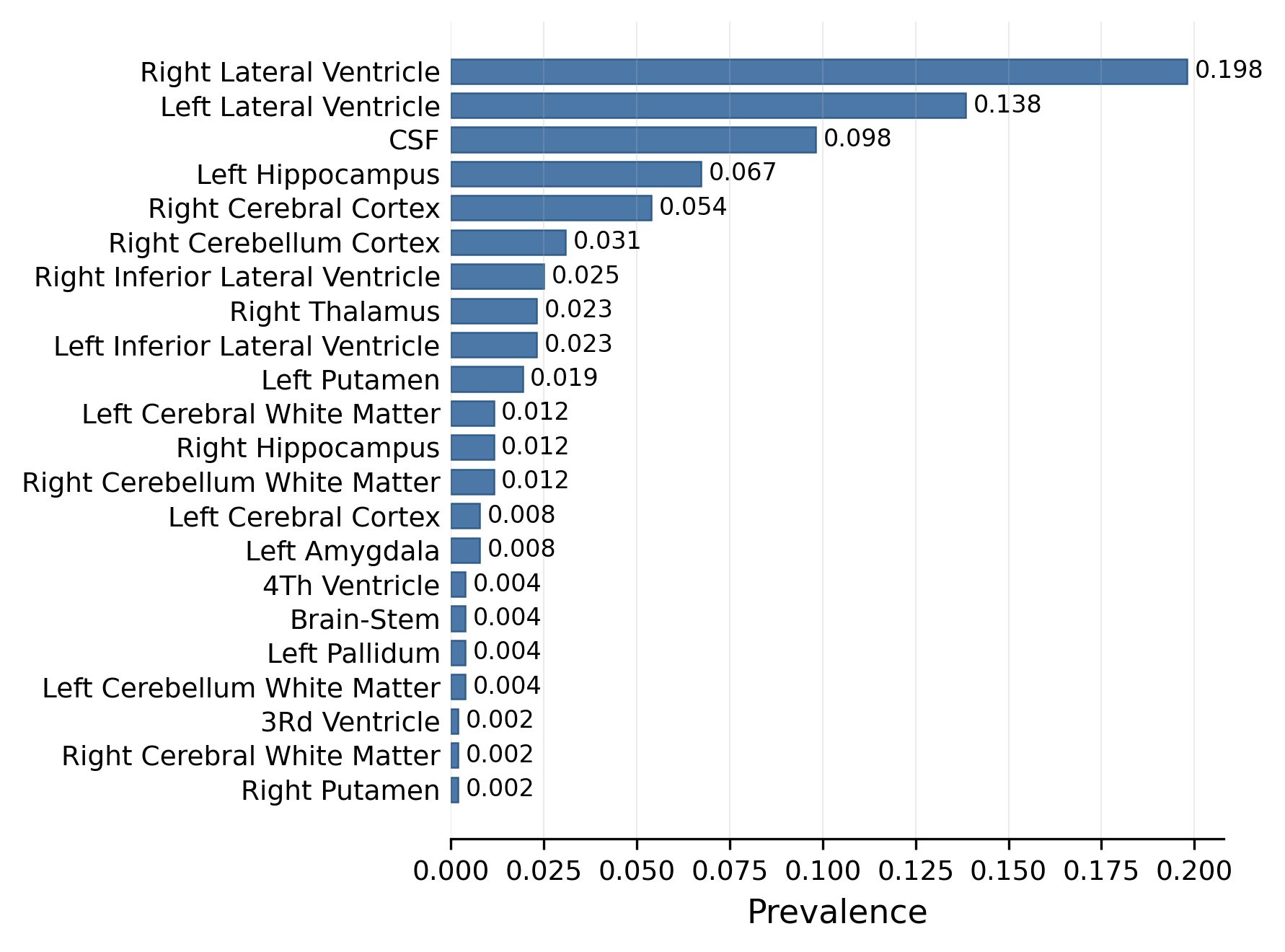}
        \caption{Image agent.}
    \end{subfigure}
    \caption{Top features extracted by modality agents for 3-year normal cognition to dementia prediction task. These features are determined by respective modality agents to have positively contributed to dementia.}
    \label{fig:modality_agent_evidences}
\end{figure}

\clearpage
\section{Clinician reader study questions}
\label{sec:reader_study_appendix}
We report here the clinician assessment questions we asked for clinicians with \CEREBRA dashboard and without, using original data. 

\begin{tcolorbox}[
title=\textbf{CEREBRA} Dashboard Questions,
colbacktitle=cerebrablue
]
\smaller

\textbf{Section 1 --- Clinician Evaluation}
\begin{itemize}[itemsep=0.2em, topsep=0.2em, parsep=0pt]
    \item[\textbf{Q1.}] Estimated dementia risk within the next 3 years
    \begin{itemize}[itemsep=0.2em, topsep=0.2em, parsep=0pt]
        \item Low
        \item Moderate
        \item High
        \item Unable to determine
    \end{itemize}

    \item[\textbf{Q2.}] Most likely diagnostic category (select all that apply)
    \begin{itemize}[itemsep=0.2em, topsep=0.2em, parsep=0pt]
        \item Normal cognition / No impairment
        \item Mild cognitive impairment (MCI)
        \item Alzheimer's disease
        \item Vascular cognitive impairment / Vascular dementia
        \item Frontotemporal dementia
        \item Lewy body dementia
        \item Not enough information / Unsure
    \end{itemize}

    \item[\textbf{Q3.}] Confidence in your assessment
    \begin{itemize}[itemsep=0.2em, topsep=0.2em, parsep=0pt]
        \item Low confidence
        \item Moderate confidence
        \item High confidence
    \end{itemize}
\end{itemize}

\vspace{0.2cm}
\textbf{Section 2 --- Risk Factors}
\begin{itemize}[itemsep=0.2em, topsep=0.2em, parsep=0pt]
    \item[\textbf{Q4.}] Accuracy of the risk factors shown
    \begin{itemize}[itemsep=0.2em, topsep=0.2em, parsep=0pt]
        \item Inaccurate
        \item Partially accurate
        \item Mostly accurate
    \end{itemize}

    \item[\textbf{Q5.}] Clinical relevance of the risk factors
    \begin{itemize}[itemsep=0.2em, topsep=0.2em, parsep=0pt]
        \item Not relevant
        \item Somewhat relevant
        \item Highly relevant
    \end{itemize}
\end{itemize}

\vspace{0.2cm}
\textbf{Section 3 --- Supporting Evidence}
\begin{itemize}[itemsep=0.2em, topsep=0.2em, parsep=0pt]
    \item[\textbf{Q6.}] Accuracy of the supporting evidence
    \begin{itemize}[itemsep=0.2em, topsep=0.2em, parsep=0pt]
        \item Inaccurate
        \item Partially accurate
        \item Accurate
    \end{itemize}

    \item[\textbf{Q7.}] Did the supporting evidence align with your own interpretation?
    \begin{itemize}[itemsep=0.2em, topsep=0.2em, parsep=0pt]
        \item No
        \item Partially
        \item Yes
    \end{itemize}
\end{itemize}

\vspace{0.2cm}
\textbf{Section 4 --- Trust \& Usefulness}
\begin{itemize}[itemsep=0.2em, topsep=0.2em, parsep=0pt]
    \item[\textbf{Q8.}] The dashboard helped me understand patient risk
    \begin{itemize}[itemsep=0.2em, topsep=0.2em, parsep=0pt]
        \item Disagree
        \item Neutral
        \item Agree
    \end{itemize}

    \item[\textbf{Q9.}] The dashboard made decision-making easier
    \begin{itemize}[itemsep=0.2em, topsep=0.2em, parsep=0pt]
        \item Disagree
        \item Neutral
        \item Agree
    \end{itemize}
\end{itemize}
\end{tcolorbox}

\begin{tcolorbox}[
title=Original Data Questions,
colbacktitle=cerebrablue
]
\smaller

\textbf{Section 1 --- Clinician Evaluation}
\begin{itemize}[itemsep=0.2em, topsep=0.2em, parsep=0pt]
    \item[\textbf{Q1.}] Estimated dementia risk within the next 3 years
    \begin{itemize}[itemsep=0.2em, topsep=0.2em, parsep=0pt]
        \item Low
        \item Moderate
        \item High
        \item Unable to determine
    \end{itemize}

    \item[\textbf{Q2.}] Most likely diagnostic category (select all that apply)
    \begin{itemize}[itemsep=0.2em, topsep=0.2em, parsep=0pt]
        \item Normal cognition / No impairment
        \item Mild cognitive impairment (MCI)
        \item Alzheimer's disease
        \item Vascular cognitive impairment / Vascular dementia
        \item Frontotemporal dementia
        \item Lewy body dementia
        \item Not enough information / Unsure
    \end{itemize}

    \item[\textbf{Q3.}] Confidence in your assessment
    \begin{itemize}[itemsep=0.2em, topsep=0.2em, parsep=0pt]
        \item Low confidence
        \item Moderate confidence
        \item High confidence
    \end{itemize}
\end{itemize}

\end{tcolorbox}
\clearpage
\section{\CEREBRA dashboard demonstration}
\label{sec:cerebra_dashboard_demo}

\begin{figure}[H]
    \centering
    \begin{subfigure}[t]{.85\linewidth}
        \centering
        \includegraphics[width=\linewidth]{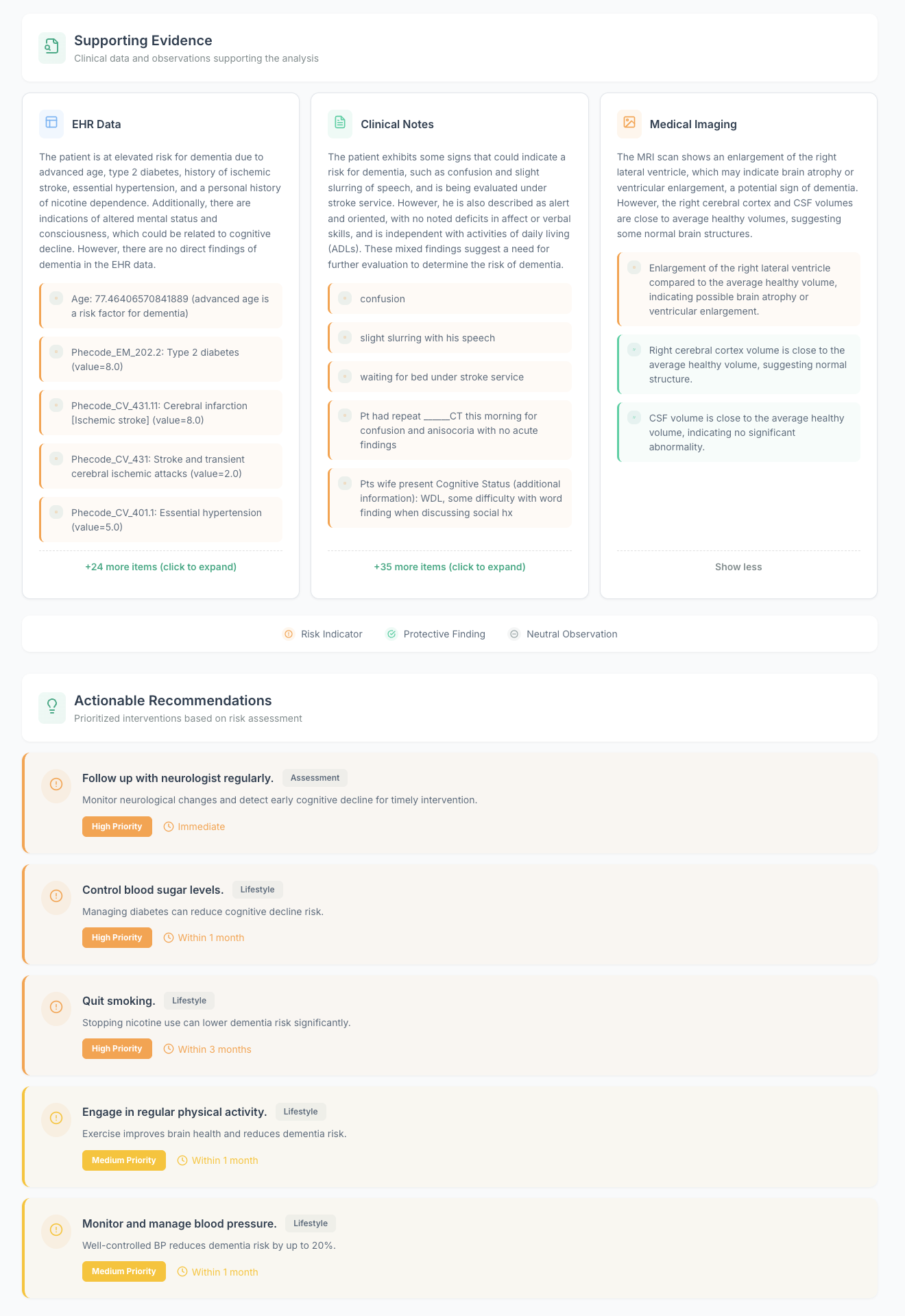}
    \end{subfigure}
    \caption{\CEREBRA dashboard, modality supporting evidence and recommendation sections.}
    \label{fig:cerebra_dashboard_2}
\end{figure}

\end{document}